\documentclass{article}

\usepackage{arxiv}
\usepackage{amsmath, amsthm, amsfonts, amssymb}
\usepackage[utf8]{inputenc} % allow utf-8 input
\usepackage[T1]{fontenc}    % use 8-bit T1 fonts
\usepackage{hyperref}       % hyperlinks
\usepackage{url}            % simple URL typesetting
\usepackage{booktabs}       % professional-quality tables
\usepackage{amsfonts}       % blackboard math symbols
\usepackage{nicefrac}       % compact symbols for 1/2, etc.
\usepackage{microtype}      % microtypography
\usepackage{cleveref}       % smart cross-referencing
\usepackage{lipsum}         % Can be removed after putting your text content
\usepackage{graphicx}
\usepackage[numbers]{natbib}
\usepackage{doi}
\usepackage{bm}

\usepackage{mathrsfs}
\usepackage[noend]{algpseudocode}
\usepackage{algorithmicx,algorithm}
\usepackage{graphicx}
\usepackage{wrapfig}
\usepackage{fancyhdr}
\usepackage{mathtools}
\usepackage{xcolor,color}
\usepackage{esint}
\usepackage{enumerate}
\usepackage{bbm}
\numberwithin{equation}{section}

\usepackage{mathrsfs}
\usepackage{appendix}
\usepackage{subcaption}
\usepackage{bm}
\usepackage{booktabs}
\usepackage{multirow}

\title{Solving time dependent Fokker-Planck equations via temporal normalizing flow}

\author{{Xiaodong Feng} \thanks{ LSEC, Institute of Computational Mathematics and Scientific/Engineering
		Computing, AMSS, Chinese Academy of Sciences, Beijing, China. Email: xdfeng@lsec.cc.ac.cn.}
	\\
	\And
	{Li Zeng} \thanks{ LSEC, Institute of Computational Mathematics and Scientific/Engineering
		Computing, AMSS, Chinese Academy of Sciences, Beijing, China. Email: zengli@lsec.cc.ac.cn.}
	\\
	 \And
	 {Tao Zhou} \thanks{ LSEC, Institute of Computational Mathematics and Scientific/Engineering
	 	Computing, AMSS, Chinese Academy of Sciences, Beijing, China. Email: tzhou@lsec.cc.ac.cn. This work was partially supported by the national natural science foundation of China (Nos. 11731006, 11831010 and 11871068), the national key R\&D program (No. 2020YFA0712000), and the national key basic research program (No. 2018YFA0703903).}\\
}

\begin{document}
	\maketitle
	
\begin{abstract}
In this work, we propose an adaptive learning approach based on temporal normalizing flows for solving time-dependent Fokker-Planck (TFP) equations. It is well known that solutions of such equations are probability density functions, and thus our approach relies on modelling the target solutions with the temporal normalizing flows. The temporal normalizing flow is then trained based on the TFP loss function, without requiring any labeled data. Being a machine learning scheme, the proposed approach is mesh-free and can be easily applied to high dimensional problems. We present a variety of test problems to show the effectiveness of the learning approach.
\end{abstract}


\section{Introduction}
The Fokker-Planck (FP) equations, which describe the time evolution of probability density functions (PDFs) of complex stochastic systems,  have been widely used in different fields such as physical and biological modelling\cite{risken1996fokker}\cite{sjoberg2009fokker}. Solving the FP equations numerically has been an important research topic in the past few decades. Generally, there are two main ways to obtain the PDFs of stochastic dynamics: solving the FP equations directly, or evaluating the transition probability density of the associated stochastic differential equations(SDEs). Traditional numerical methods for doing this include the finite element methods \cite{deng2009finite}, the finite difference methods \cite{kumar2006solution}, the path integral methods \cite{wehner1983numerical}, to name just a few. One of the biggest difficulties of these approaches is that either discretizition of  a high dimensional (unbounded) physical space is needed, or a large number of sample paths via Monte Carlo method \cite{hirvijoki2015monte} should be used. %Since the accuracy of solution is related to the number of sample paths, the computational cost of solving the joint probability density function may become prohibitive.
	
In recent years, machine learning techniques have been widely used to solve partial differential equations (PDEs),  see e.g. \cite{weinan2020machine,shin2020convergence,huang2021augmented} and references therein. Among others, we mention the deep Galerkin method \cite{sirignano2018dgm}, the deep Ritz method \cite{weinan2018deep}, and the so-called  physics-informed neural networks (PINNs) \cite{raissi2019physics}. These approaches have been widely applied to many realistic problems, such as fluid mechanics \cite{raissi2020hidden,brunton2020machine}, high dimensional PDEs (with applications in computational finance) \cite{han2018solving,zang2020weak}, uncertainty quantification \cite{yang2021b,qin2021deep,zhang2018deep,iten2020discovering,meng2020composite}, to name just a few.  Meanwhile, generative models such as generative adversarial networks \cite{goodfellow2014generative}, variational autoencoder \cite{kingma2013auto} and normalizing flow (NF) \cite{papamakarios2019normalizing,rezende2015variational}, have also been successfully applied to learn forward and inverse PDEs \cite{chen2021solving,zhu2019physics,yang2019adversarial,liu2020neural}. For instance, physics-informed generative adversarial model was proposed in \cite{yang2020physics} to tackle high dimensional stochastic differential equations. In \cite{guo2021normalizing}, normalizing field flow was developed to build surrogate models for uncertainty quantification problems.

As the solution of the Fokker-Planck equation is a probability density function, solving this problem can also be considered as a density estimation problem. This motivates us to propose in this work an adaptive learning scheme based on the normalizing flow. More precisely, our approach relies on modelling the target solutions of the FP equations. Consequently, the temporal normalizing flow is trained based on the TFP loss function (the physics informed residual), without requiring any labeled data. We list in the following the main features and related works of our approach:

\begin{itemize}
\item Our approach is an extension of the previous interesting work \cite{tang2021adaptive} where only \textit{steady state} FP equations are investigated. To address time dependent problems, we propose an adaptive density approximation scheme based on temporal normalizing flow.
\item Being a machine learning scheme, the proposed approach is mesh-free and can be easily applied to high dimensional problems.
\item Our approach is based on PDE-loss functions, and does not need sample paths generated from stochastic differential equations. This is different from previous works such as \cite{lu2021learning} where sample paths of the corresponding stochastic dynamics are used.
\item We present a variety of test problems, including FP equations with linear or nonlinear drift terms and high dimensional problems, to show the effectiveness of the learning approach.
\end{itemize}
	
	The remainder of this paper is structured as follows. In section \ref{section:2}, we provide with some preliminary results. Section \ref{section:3} provides our adaptive density approximation scheme based on the temporal normalizing flow. In Section \ref{section:4} we demonstrate the efficiency of our adaptive sampling approach with several numerical experiments. We then give some concluding remarks in Section \ref{section:5}.

\section{Problem setup}\label{section:2}
	The main aim of this work is to solve the time dependent FP equations. To this end, we first give a brief introduction to the FP equations.
	\subsection{Fokker-Planck equations}
	Consider the state variable $\bm{X}_t$ modeled by the following stochastic differential equation
	\begin{equation}
		\mathrm{d}\bm {X} _{t}={\boldsymbol {\mu }}(\bm {X} _{t},t)\,\mathrm{d}t+{\boldsymbol {\sigma }}(\bm {X} _{t},t)\,\mathrm{d}\bm {W} _{t},
		\label{sde}
	\end{equation}
	where $\bm {X} _{t}$  and $\bm{\mu}(\bm {X} _{t},t)$ are $d$-dimensional random vectors,  $\bm{\sigma}(\bm{X}_t, t)$ is a $d\times M$ matrix and $\bm{W}_t$ is an $M$-dimensional standard Wiener process. The probability density $p(\bm{x} ,t)$ for $\bm {X} _{t}$ satisfies the so called time dependent Fokker-Planck (TFP) equation:
	\begin{equation}
		{\displaystyle {\frac {\partial p(\bm{x} ,t)}{\partial t}}=-\sum _{i=1}^{d}{\frac {\partial }{\partial x_{i}}}\left[\mu_{i}(\bm{x} ,t)p(\bm{x} ,t)\right]+\sum _{i=1}^{d}\sum _{j=1}^{d}{\frac {\partial ^{2}}{\partial x_{i}\,\partial x_{j}}}\left[D_{ij}(\bm{x} ,t)p(\bm{x} ,t)\right],}
		\label{time_depe_fp_eqn}
	\end{equation}
	where $\bm{x}=(x_1,\cdots, x_d)$, $\bm{\mu}$ is the drift vector $\bm{\mu}=(\mu_1,\cdots, \mu_d)$ and $\bm{D}$ is the diffusion tensor $\bm{D}=\frac{1}{2}\bm{\sigma}\bm{\sigma^T},$ i.e.,
	\begin{equation*}
		{\displaystyle D_{ij}(\bm{x} ,t)={\frac {1}{2}}\sum _{k=1}^{M}\sigma _{ik}(\bm{x} ,t)\sigma _{jk}(\bm{x} ,t).}
	\end{equation*}
	Sometimes, one may focus on the stationary solution of (\ref{time_depe_fp_eqn}), i.e., the invariant measure independent of time,
	\begin{equation}
		{\sum _{i=1}^{d}{\frac {\partial }{\partial x_{i}}}\left[\mu_{i}(\bm{x})p(\bm{x})\right]+\sum _{i=1}^{d}\sum _{j=1}^{d}{\frac {\partial ^{2}}{\partial x_{i}\,\partial x_{j}}}\left[D_{ij}(\bm{x})p(\bm{x})\right]=0.}
		\label{fp_eqn}
	\end{equation}
	In general, the solutions of the above TFP equations are defined in the unbounded domain with the following boundary condition:
	\begin{equation}\label{BCs}
		p(\bm{x})\to 0 \quad \mbox{as} \quad \Vert \bm{x}\Vert_2 \to \infty.
	\end{equation}
Furthermore, the solution, being a density function,  should also satisfy the following extra constraint
	\begin{equation}\label{density_constraint}
		\int_{\mathbb{R}^d}p(\bm{x},t)\mathrm{d}\bm{x}\equiv 1, \quad \mbox{and} \quad p(\bm{x},t)\geq 0.
	\end{equation}

\subsection{Normalizing flow and RealNVP}

Notice that both the constraints (\ref{BCs}) and (\ref{density_constraint}) bring essential difficulties for mesh-dependent numerical schemes when solving the TFP equations. To this end, we introduce in this section the normalizing flow (NF), which serves as a potentially efficient tool for handling TFP equations.

Normalizing flow provides a way for constructing flexible probability distributions over continuous random variables. Specifically, suppose we want to approximate an unknown random variable $\bm{x}\in \mathbb{R}^d$ (we define $p_{\bm {X}}(\bm{x})$ as its density function). The NF model seeks to find an invertible mapping $f:\bm{x}\to \bm{z}$, between a simple reference variable $\bm{z}\in \mathbb{R}^d$ (with known probability distribution $p_{\bm{Z}},$ e.g., Gaussian) and the target variable $\bm{x}$, i.e., $\bm{x}=f^{-1}(\bm{z}).$ Then the distribution of $\bm{x}$ is given by
	\begin{equation}
		p_{\bm{X}}(\bm{x})=p_{\bm{Z}}(f(\bm{x}))\bigg|\det \nabla_{\bm{x}} f(\bm{x})\bigg|,
		\label{variable_formula}
	\end{equation}
where $\nabla_{\bm{x}} f(\bm{x})$ is the Jacobian. Given observations of $\bm{x},$ the unknown invertible mapping can be learned via the maximum likelihood estimations.

In the NF model, a complex invertible mapping $f(\cdot)$ can be constructed by stacking a sequence of simple bijections, each of which is a shallow neural network, thus the overall mapping is a deep neural network. Namely, the mapping $f(\cdot)$ can be written in a composite form:
	\begin{equation}
		\bm{z}=f(\bm{x})=f_{[L]}\circ f_{[L-1]}\circ \cdots \circ f_{[1]}(\bm{x}).
		\label{com_no_time}
	\end{equation}
Its inverse and Jacobian determinants are given by
\begin{eqnarray}
	\bm{x} = f^{-1}(\bm{z})=f_{[1]}^{-1}\circ \cdots \circ f_{[L-1]}^{-1}\circ f^{-1}_{[L]}(\bm{z}),\\
	\vert \det \nabla_{\bm{x}} f(\cdot)\vert = \prod_{i=1}^L\vert \det \nabla _{\bm{x}_{[i-1]}}f_{[i]}(\cdot)\vert,
\end{eqnarray}
where $\bm{x}_{[i-1]}$ indicates the immediate variables with $\bm{x}_{[0]}=\bm{x}, \bm{x}_{[L]}=\bm{z}.$
Since computing the Jacobian determinants of large matrices is generally computationally very expensive, the structure of function $f$ should be carefully designed. A successful example is RealNVP \cite{dinh2016density}. Let $\bm{x}_{[i]}=(\bm{x}_{[i],1}, \bm{x}_{[i],2})$ be a partition with $\bm{x}_{[i],1}\in \mathbb{R}^m$ and $\bm{x}_{[i],2}\in \mathbb{R}^{d-m}$. RealNVP proposes to use the following affine transformation:
\begin{equation}
	\begin{aligned}
		& \bm{x}_{[i], 1}=\bm{x}_{[i-1],1},\\
		& \bm{x}_{[i], 2}=\bm{x}_{[i-1],2}\odot \exp (\bm{s}_i(\bm{x}_{[i-1],1})) + \bm{b}_i(\bm{x}_{[i-1],1}),
	\end{aligned}
\label{realnvp_coupling}
\end{equation}
where $\bm{s}_i:\mathbb{R}^{m}\to \mathbb{R}^{d-m},$  $\bm{b}_i:\mathbb{R}^{m}\to \mathbb{R}^{d-m}$ are scaling and translation depending on $\bm{x}_{[i-1], 1}$, and $\odot$ is Hadamard product. By doing this, the Jacobian matrix $f(\bm{x})$ (for the vector case) is lower-triangular whose determinant can be evaluated efficiently. Furthermore, $\bm{s}_i$ and $\bm{b}_i$ can be modeled by complex neural networks to enhance the expression capacity of the invertible map.

The RealNVP model has been widely adopted recently \cite{kingma2016improved,kingma2018glow, prenger2019waveglow}. Nevertheless, we mention here the interesting work \cite{tang2020deep} where a so-called KR-net is proposed to enhance the expressive power and improve the numerical stability over RealNVP.

%\begin{equation}
%\begin{aligned}
%	& x_{[i], 1}=x_{[i-1],1},\\
%	& x_{[i], 2}=x_{[i-1],2}\odot \big(1+\alpha\tanh (s_i(x_{[i-1],1}))\big) + e^{\beta_i}\odot \tanh\big(b_i(x_{[i-1],1})\big),
%\end{aligned}
%\end{equation}
%where $s_i$ and $b_i$ are the same as formula (\ref{realnvp_coupling}), the parameters $\beta _ i \in \mathbb{R}^{d-m}$ are learnable variables and $\alpha \in(0,1]$ is a user-specified hyperparameter.

\section{Methodology} \label{section:3}	

In this section, we shall propose our learning scheme for the TFP equations.

\subsection{Temporal normalizing flow} \label{section:3.1}
Note that the NF model discussed in the above section is time-independent. To address time dependent FP equations, we first introduce the so-called temporal normalizing flow (TNF). The TNF was proposed by Both and Kusters \cite{both2019temporal} aiming at estimating time evolving distributions.
	
	Let $\widehat{\bm{x}}$ be a $d+1$ dimensional real vector including the temporal variable, namely, $\widehat{\bm{x}}=(\bm{x},t)$. In addition, $\widehat{\bm{z}}$ is a $d+1$ dimensional latent variable, $\widehat{\bm{z}}=(\bm{z}, t^*)$, obeying a simple distribution. Then, we consider the following transformation:
	\begin{equation}
		p_{\widehat{\bm{X}}}(\widehat{\bm{x}})=p_{\widehat{\bm{Z}}}(\widehat{\bm{z}})\vert \det J\vert ,\quad \widehat{\bm{z}}=f(\widehat{\bm{x}}),
		\label{variable_formula_multi_dim}
	\end{equation}
	where $J$ is Jacobian of $f(\bm{\widehat{x}})$.  As the conservation law for the temporal axis is not always correct, i.e. $\int p(\bm{x}, t)\mathrm{d}\bm{x}\mathrm{d}t\neq 1$, we cannot include it as an additional dimension in eq. (\ref{variable_formula_multi_dim}). Notice that the determinant of the Jacobian is
	\begin{equation}
		\det J = \left|
		\begin{split}
			&\frac{\partial \bm{z}}{\partial \bm{x}}  &\frac{\partial \bm{z}}{\partial t}\\
			&\frac{\partial t^*}{\partial \bm{x}} &\frac{\partial t^*}{\partial t}
		\end{split}
		\right|.
	\end{equation}
where $\bm{z}$ is the latent spatial coordinate and $t^*$ the latent temporal coordinate. By letting the latent time $t^*$ be exactly equal to the real time $t$, one obtains
	\begin{equation}
	\det J = \left|
		\begin{aligned}
		&\frac{\partial \bm{z}}{\partial \bm{x}}  &\frac{\partial \bm{z}}{\partial t}\\
		&\;\;0 &1\;\;
		\end{aligned}
	\right|=\left|\frac{\partial \bm{z}(\bm{x},t)}{\partial \bm{x}}\right|,
\end{equation}
Consequently, the temporal normalizing flow can be written as
\begin{equation}
	p_{\widehat{\bm{X}}}(\bm{x},t) = p_{\widehat{\bm{Z}}}(\bm{z},t) \left|\frac{\partial \bm{z}}{\partial \bm{x}}\right|,\quad \bm{z}=f(\bm{x},t).
\end{equation}

Similar to (\ref{com_no_time}), $f(\cdot,t)$ can be constructed by stacking a sequence of simple bijections. Namely, given time $t$, we have
\begin{equation}
	\bm{z} = f(\bm{x},t) = f_{[L]}\circ f_{[L-1]} \circ \cdots \circ f_{[1]}(\bm{x},t).
\end{equation}
The inverse of $f$ is then given by
\begin{equation}
	\bm{x} = f^{-1}(\bm{z},t) = f^{-1}_{[1]}\circ f^{-1}_{[2]}\circ \cdots \circ f^{-1}_{[L]}(\bm{z},t).
\end{equation}
Consequently, the corresponding Jacobian can be obtained by using the chain rule:
\begin{equation}
	\vert \det \nabla_{\bm{x}} f(\cdot,t)\vert = \prod_{i=1}^L\vert \det \nabla _{\bm{x}_{[i-1]}}f_{[i]}(\cdot, t)\vert,
\end{equation}
where $\bm{x}_{[i-1]}$ indicates the immediate variables with $\bm{x}_{[0]}=\bm{x},\bm{ x}_{[L]}=\bm{z}.$

\subsection{Architecture} \label{section:3.2}
\begin{figure}[!h]
	\centering
	\includegraphics[scale=0.25]{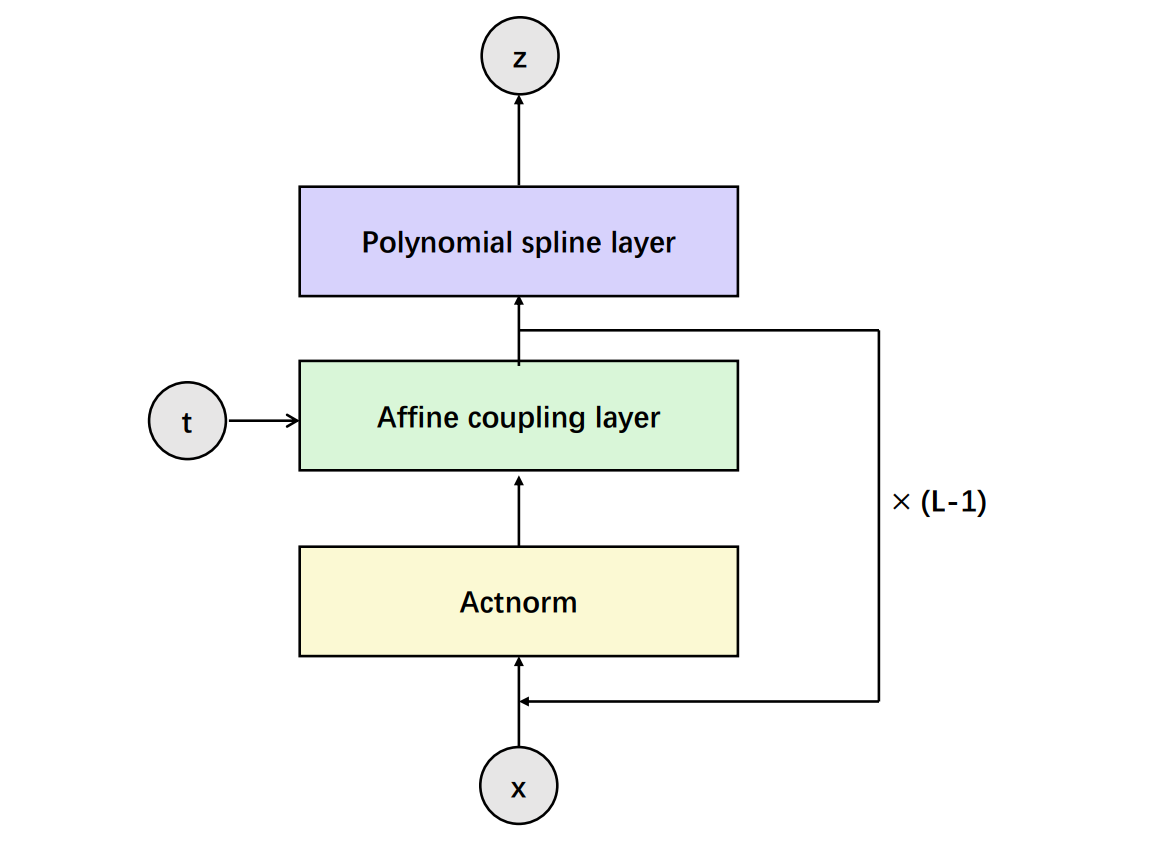}
	\caption{The architecture of our proposed model.}
	\label{fig:architecture}
\end{figure}

In this section, we present the main architecture of our TNF. Inspired by the construction of KR-net \cite{tang2020deep}, our TNF model can be regarded as a simplified extension of KR-net from spatial domain to temporal-spatial domain. Specifically, each $f_{[i]}$ in our TNF model consists of an Actnorm layer, followed by a modified time-dependent affine mapping \cite{dinh2016density}. For the last layer, we apply a polynomial spline transformation to increase the modelling power \cite{muller2019neural}. Figure \ref{fig:architecture} shows the flow chart of our proposed model, and we shall provide with more details in the following subsections.

\subsubsection{ Actnorm: scale and bias layer}
We adopt the Actnorm layer with data dependent initialization proposed by  Kingma and Dhariwal \cite{kingma2018glow}:
\begin{equation}
%	\widetilde{\bm{x}}_{[i]}=\bm{a}_i \odot \bm{x}_{[i]} + \bm{b}_i,
		\bm{y}_{[i]}=\bm{a}_i \odot \bm{x}_{[i]} + \bm{b}_i,
\end{equation}
where $\bm{a}_i$ and $\bm{b}_i$ are trainable parameters. The parameters can be initialized by the statistic mean and variance related to mini batch data. After initialization, the scale and bias are treated as regular trainable parameters that are
independent of the data.
The inverse can be easily obtained via
\begin{equation}
%	\bm{x}_{[i]} = (\widetilde{\bm{x}}_{[i]}-\bm{b}_i)/\bm{a}_i.
		\bm{x}_{[i]} = (\bm{y}_{[i]}-\bm{b}_i)/\bm{a}_i,
\end{equation}
where division here is operated on each corresponding component.

\subsubsection{Affine coupling layer}
 Let $\bm{x}_{[i]}=(\bm{x}_{[i],1}, \bm{x}_{[i],2})$ be a partition with $\bm{x}_{[i],1}\in \mathbb{R}^m$ and $\bm{x}_{[i],2}\in \mathbb{R}^{d-m}$. Then we consider a time-dependent affine coupling layer $f_{[i]}(\cdot,t)$ as follows:
\begin{equation}
	\begin{aligned}
		& \bm{x}_{[i], 1}=\bm{x}_{[i-1],1},\\
		& \bm{x}_{[i], 2}=\bm{x}_{[i-1],2}\odot \big(\bm{1}_{d-m}+\beta\tanh (\bm{s}_i(\bm{x}_{[i-1],1}, t))\big) + e^{\bm{\zeta}_i}\odot\tanh(\bm{q}_i(\bm{x}_{[i-1],1}, t)),
	\end{aligned}
\end{equation}
where $|\beta|< 1$ is a user-specified parameter (e.g. 0.6), $\bm{1}_{d-m}$ denotes a $d-m$ dimensional vector whose components are all 1, $\bm{s}_i:\mathbb{R}^{m+1}\to \mathbb{R}^{d-m},$  and $\bm{q}_i:\mathbb{R}^{m+1}\to \mathbb{R}^{d-m}$ are scaling and translation depending on $\bm{x}_{[i-1], 1}$ and $t.$ While $\bm{\zeta}_i\in \mathbb{R}^{d-m}$ is a trainable variable which depends on data directly. Notice that the inverse can be easily computed via:
\begin{equation}
	\begin{aligned}
		& \bm{x}_{[i-1], 1}=\bm{x}_{[i],1},\\
		& \bm{x}_{[i-1], 2}=(\bm{x}_{[i],2} - e^{\bm{\zeta}_i}\odot\tanh(\bm{q}_i(\bm{x}_{[i],1}))) \odot \big(\bm{1}_{d-m}+\beta\tanh (\bm{s}_i(\bm{x}_{[i],1}, t))\big)^{-1}.
	\end{aligned}
\end{equation}
The Jacobian of $\bm{x}_{[i]}(\cdot,t)$ is given by
\begin{equation}
	\nabla _{\bm{x}_{[i-1]}}\bm{x}_{[i]}(\cdot,t) = \left[\begin{array}{cc}
		\bm{I}&\bm{0}\\
		\nabla _{\bm{x}_{[i-1],1}} \bm{x}_{[i],2}& \mathrm{diag}(\bm{1}_{d-m}+\alpha\tanh (\bm{s}_i(\bm{x}_{[i-1],1}, t)))
	\end{array}\right].
\end{equation}
Furthermore, we can model $\bm{s}_{i},\bm{b}_{i}$ via neural networks
\begin{equation}
	(\bm{s}_{i},\bm{q}_{i}) =\mathrm{NN}_{[i]}(\bm{x}_{[i-1]}, t).
	\label{NN_i}
\end{equation}
%That is, $\bm{s}_i$ and $\bm{b}_i$ are the output of neural network whose inputs are $\bm{x}_{[i-1], 1}$ and $t$. 
Note that $f_{[i]}(\cdot,t)$ only changes $\bm{x}_{[i-1],2}$, hence we can exchange the positions of $\bm{x}_{[i],1}$ and $\bm{x}_{[i],2}$ to ensure that each component of $\bm{x}$ can be updated.
\subsubsection{Polynomial spline  layer}
For the last layer, we use the polynomial spline layer proposed by \cite{tang2021adaptive}. %Work \cite{muller2019neural} proposed a kind of polynomial spline layer for bounded domain. For unbounded domain, we can truncate the domain by $[-c, c]$ where $c$ is large enough. Then area $[-c, c]$ can be treated like work \cite{muller2019neural}. As for other area, we can map it to fixed value artifically. 
Without loss of generality, we present the formula below for the case of $d=1.$ While for higher dimensional cases one simply uses the tensor product arguments. The associated transformation is given by
\begin{equation}
	G(x) = \left\{
	\begin{array}{cc}
		\gamma (x+c)-c,& x\in(-\infty, -c)\\
		2c\widehat{G}(\frac{x+c}{2c})-c,& x\in(-c,c)\\
		\gamma (x-c) + c,& x\in(c,\infty)
	\end{array}
	\right.
	\label{non_eq}
\end{equation}
where $\gamma>0$ is a used-specified small parameter (e.g. $10^{-6}$), $c$ is a prior constant describing the range of nonlinear transformation, and $\widehat{G}$ is a continuous piecewise linear cumulative probability function. Specifically, let $0=l_0<l_1<\cdots<l_{m-1}<l_m=1$ be a given partition in the unit interval and $\{k_j\}_{j=0}^m$ be the corresponding weights satisfying $\sum_j k_j=1$. A piecewise linear polynomial can be defined as follows:
\begin{equation}
	p(x) =\frac{k_{j+1}-k_j}{l_{j+1}-l_j}(x-l_j) + k_j,\quad \forall x\in[l_{j},l_{j+1}],\;\forall t.
	\label{non_pdf_eq}
\end{equation}
Then the corresponding cumulative probability function $\widehat{G}$ admits the following formulation:
\begin{equation}
	\widehat{G}(x) = \frac{k_{j+1}-k_j}{2(l_{j+1}-l_j)}(x-l_j)^2+k_j(x-l_j)+\sum_{i=0}^{j-1}\frac{k_{i}+k_{i+1}}{2}(l_{i+1}-l_i),\quad \forall x\in[l_{j},l_{j+1}],\;\forall t.
	\label{non_cdf_eq}
\end{equation}

For the sake of the continuity of $p(x)$ at $x=0$ and $x=1$, one can set $k_0=k_{m}=\gamma$. Furthermore, we can model $\{k_j\}_{j=1}^{m-1}$ as:
	\begin{equation}
		k_j = \frac{\exp(\tilde{k}_j)}{\sum_{i=0}^m \exp(\tilde{k}_i)}, \quad j=1,\cdots,m-1,
	\end{equation}
where $\{\tilde{k}_j\} $ are trainable parameters. Notice that the polynomial spline layer (\ref{non_eq})-(\ref{non_cdf_eq}) yields explicit  monotonous expressions, and its inverse can be readily computed.
%{\color{red}\begin{remark}
%	 Analogous to the affine coupling layer, one can also build a time-dependent polynomial spline layer. The main difference is that the parameters $\{k_j\}$ rely on time $t$, i.e. $\tilde{k}_j(t) = \mathrm{NN}(l_j, t)$ for $j\geq 1$.
%\end{remark}}
%\textcolor{red}{adaptivity?}
%Now assume we have some sample paths of stochastic differential equation, namely, a given dataset $\mathcal{D}=\{D_j\}_{j=1}^m$,  where $D_j=\{x_i^j\}_{i=1}^n$ is sampled from $p_{\mathcal{X}}(x,t_j)$, hence one can induce corresponding negative log-likelihood:
%\begin{equation}
%	\begin{aligned}
%		\mathcal{L}&=-\sum_{j=1}^{m}\sum_{i=1}^{n}\log p_{\mathcal{X}}(x_i^j,t_j)\\
%		&=-\big[\sum_{j=1}^{m}\sum_{i=1}^{n}\log p_{\mathcal{Z}}(f(x_i^j,t_j)) +\log\vert \det J(x,t) \vert_{x=x_i^j,t=t_j}\big]
%	\end{aligned}
%\label{time_likelihood}
%\end{equation}
%If the above $t_j$ is a constant for all $j=1,\cdots,m$, the functional $\mathcal{L}$ \ref{time_likelihood}
\subsection{Solving time dependent FP equations via temporal normalizing flow} \label{section:3.3}
We are now ready to present our scheme for solving time-dependent FP equations (\ref{sde}-\ref{time_depe_fp_eqn}) with temporal normalizing flow. Specifically, consider the following TFP equations:
	\begin{equation}
		\left\{
		\begin{aligned}
			&p_t +\mathcal{N}_{\bm{x}}[p] = 0,\quad  \bm{x}\in \Omega, \,\, t\in(0,T],\\
			&p(\bm{x}, 0) = p_0(\bm{x}),  \quad \bm{x}\in\Omega,\\
            & p(\bm{x}, 0) \rightarrow 0 \,\,\, \textmd{as} \,\,\, \|\bm{x}\| \rightarrow 0,\\
			&\int_{\Omega} p(\bm{x},t)\mathrm{d}\bm{x}=1,\,\, p(\bm{x},t)\geq0, \,\,\,  \bm{x}\in \Omega, \,\, t\in[0,T],
		\end{aligned}
		\right.
	\end{equation}
where $\mathcal{N}_{\bm{x}}$ denotes a differential operator with respect to $\bm{x}$ and $\Omega=\mathbb{R}^d$. In addition, $p:{\Omega}\times{[0,T]}\to \mathbb{R}^+ \cup \{0\}$ denotes the unknown latent quantity of interest.

We proceed by approximating $p(\bm{x},t)$  via a temporal normalizing flow $p_{\bm{\theta}}(\bm{x},t)$, where $\bm{\theta}$ denotes all trainable parameters of the model, and the prior of $\bm{x}$ is assumed to be Gaussian. Then we adopt the physical law (i.e., the TFP equation) to yield the following residual
\begin{equation}
	\bm{r}(\bm{x},t)  \coloneqq \frac{\partial }{\partial t}p_{\bm{\theta}}(\bm{x},t) + \mathcal{N}_{\bm{x}}[p_{\bm{\theta}}(\bm{x},t)],
\end{equation}
where the partial derivatives with respect to time and space coordinates can be readily computed using automatic differentiation. Notice that, unlike traditional mesh-dependent approaches (such as the finite element/difference methods), the non-negativity and conservation constraints naturally hold for $p_{\bm{\theta}}(
\cdot, t).$ To this end, we propose the following loss function for training the parameter $\bm{\theta}$:
\begin{equation}
	\mathcal{L}_{\mathrm{pde}}(\bm{\theta}) = \lambda_r \mathcal{L}_r(\bm{\theta})+\lambda_{\mathrm{ic}}\mathcal{L}_{\mathrm{ic}}(\bm{\theta}),
	\label{loss_function}
\end{equation}
where $\mathcal{L}_r$ and $\mathcal{L}_{\mathrm{ic}}$ are the equation loss and the initial condition loss, respectively. More precisely, we have
\begin{equation}
	\begin{aligned}
		\mathcal{L}_r(\bm{\theta}) = \frac{1}{N_r}\sum_{i=1}^{N_r}\vert \bm{r_\theta} (\bm{x}_r^i,t_r^i)\vert ^2, \quad
		\mathcal{L}_{\mathrm{ic}}(\bm{\theta})= \frac{1}{N_{\mathrm{ic}}}\sum_{i=1}^{N_{\mathrm{ic}}}\vert p_{\bm{\theta}}(\bm{x}_{\mathrm{ic}}^i, 0)-p_0(\bm{x}_{\mathrm{ic}}^i)\vert ^2.
		\end{aligned}
\end{equation}
Here, $N_r, N_{\mathrm{ic}}$ denote the batch sizes of training data $\{(\bm{x}_r^i,t_r^i)\}_{i=1}^{N_r},\{\bm{x}_{\mathrm{ic}}^i, p_0(\bm{x}_{\mathrm{ic}}^i)\}_{i=1}^{N_{\mathrm{ic}}}$, respectively. $\{\lambda_r,\lambda_{\mathrm{ic}}\}$ are weight parameters.

Notice that for each $t,$ $p(\bm{x},t)$ is defined on the whole domain $\mathbb{R}^d.$ However, in practice $p(\bm{x},t)$ is generally concentrated in a small (yet unknown) region. Moreover, the regularity of solution $p(\bm{x},t)$ may also vary in the computation domain. Consequently, how to efficiently choose the training points becomes a core issue. Obviously, a uniform sampling approach in a possibly large domain is not practical. We thus need to develop adaptive sampling strategies. In the TNF framework, this becomes possible, and we propose the following strategy:
\begin{enumerate}
	\item[1.] Initialization: generate an initial training set with samples uniformly distributed in a given domain:
$$C_0=\{\bm{x}_i^r,t_i^r\}_{i=1}^{N_r}\subset D_0.$$
	\item[2.] Train the temporal normalizing flow by minimizing the loss function $(\ref{loss_function})$ with training data $C_0$ to obtain a new probability distribution $p_1(\bm{x},t;\bm{\theta})$.
	
	\item[3.] Generate samples with $p_1(\cdot, t_i^r;\bm{\theta})$ to get a new training set $C_1=\{\tilde{\bm{x}}_i^r,t_i^r\}_{i=1}^{N_r},$ and set $C_0=C_1$. Notice that for each $t_i^r$, $\tilde{\bm{x}}_i^r$ can be obtained by transforming the prior Gaussian samples via the inverse temporal normalizing flow.

\item[4.] Repeat steps 2-3 for $N_{\mathrm{adaptive}}$ times to get a convergent approximation.
\end{enumerate}
More details for the sampling procedure are given in Algorithm \ref{alg:1}.
	\begin{algorithm}
		\caption{Temporal normalizing flow for time-dependent FP equations}
		\label{alg:1}
		\begin{algorithmic}
			\State \textbf{Input:} maximum epoch number $N_e$, maximum iteration number $N_{\mathrm{adaptive}},$ hyper parameter $\alpha, \lambda_r,\lambda_{\mathrm{ic}}$, initial training data $C_{r}=\{(\bm{x}_r^i,t_r^i)\}_{i=1}^{Nr}, C_{\mathrm{ic}}=\{\bm{x}_{\mathrm{ic}}^i\}_{i=1}^{N_{\mathrm{ic}}}, C_T =\{t_r^i\}_{i=1}^{N_r}\cup \{0\}_{i=1}^{N_{\mathrm{ic}}}$, tolerance $\epsilon_1,\epsilon_2$;
			\State $L_{old} = 0$;
			\For{$k=1,\cdots,N_{\mathrm{adaptive}}$}
			\For{$j=1,\cdots,N_e$}
			\State Divide $C=\{C_r, C_{\mathrm{ic}}\}$ into $m$ batch $\{C^{ib}\}_{ib=1}^m$ randomly;
			\For{$ib = 1,\cdots,m$}
			\State Compute the loss function (\ref{loss_function}) ${L_{\mathrm{pde}}^{ib}}$ for mini-batch data $C^{ib}$;
			\State Update $\bm{\theta}$ by using Adam optimizer;
			\EndFor
			${L_{new}}=\frac{1}{m}\sum_{ib=1}^{m}{L_{\mathrm{pde}}^{ib}};$
			\If{${L_{new}}<\epsilon_1$ or $|{L_{old}-L_{new}}|<\epsilon_2$}
			 \State \textbf{Break};
			\Else
			\State{${L_{old} = L_{new}}$};
			\EndIf
			\EndFor
			\State $N_e = \alpha * N_e$;
			\State Sample from $p(\cdot,t;\bm{\theta})$ for $t\in C_T$ and update training set $C$;
			\EndFor
			\State \textbf{Output:} The predicted solution $p(\bm{x},t;\bm{\theta})$.
		\end{algorithmic}
	\end{algorithm}

	\section{Numerical results} \label{section:4}

In this section, we present a series of comprehensive numerical tests to demonstrate the effectiveness of the proposed algorithm. To quantitatively evaluate the accuracy of the numerical solution $p_{\bm{\theta}}$, we shall consider both the relative $L^2$ error $\Vert p^* - p_{\bm{\theta}}\Vert_2 / \Vert p^*\Vert_2$ and the relative KL divergence given by $$\frac{D_{\mathrm{KL}}(p^*(t)||p_{\bm{\theta}}(t))}{H(p^*(t))} = \frac{\mathbb{E}_{p^*(t)}\log(p^*(t)/p_{\bm{\theta}}(t))}{-\mathbb{E}_{p^*(t)}\log p^*(t)},$$ where $\mathbb{E}$ denotes the expectation. We approximate the above advocated relative KL divergence by Monte Carlo integration, namely,
	$$\frac{D_{\mathrm{KL}}(p^*(t)||p_{\bm{\theta}}(t))}{H(p^*(t))}\approx \frac{1}{N_v}\frac{\sum_{i=1}^{N_v}\big(\log(p^*(\bm{x}_i;t)-\log p_{\bm{\theta}}(\bm{x}_i;t))\big)}{-\sum_{i=1}^{N_v}\log p^*(\bm{x}_i;t)}.$$
Here $p^*$ is the reference/exact solution, and $\bm{x}_i(t)$ is drawn from the ground truth $p^*(\bm{x};t)$ and the amount of validation data is set to $N_v=10^6$ for each $t$.
	
%Throughout all our tests, we use the feed forward neural networks with six hidden layers and 200 neurons for the coupling flows $f(\cdot, t).$ 
We shall employ hyperbolic tangent  function (Tanh) as the activation function. For each $i$, $\mathrm{NN}_{[i]}$ (see \ref{NN_i}) is a feed forward neural network with two hidden layers and $32$ neurons. We use a half-half partition $\bm{x}_{[i]}=(\bm{x}_{[i], 1}, \bm{x}_{[i],2})$, $\bm{x}_{[i], 1}\in\mathbb{R}^{\lfloor d/2\rfloor}$, $\bm{x}_{[i],2}\in\mathbb{R}^{d-\lfloor d/2 \rfloor}$. We initialize all trainable parameters using Glorot initialization \cite{glorot2010understanding}. For the training procedure, we use the Adam optimizer \cite{kingma2014adam} with an initial learning rate 0.001. All training data have the shape $N_{\bm{x}}\times {N_t}$, where $N_{\bm{x}}$ is the number of spatial sample points and $N_t$ is the temporal sample points. All numerical tests are implemented with Pytorch.
	
	\subsection{A toy example}
	As our first example, we start with a  benchmark example to illustrate how the adaptive algorithm works. We consider a 2D TFP equation of the form
	\begin{equation}
		\begin{aligned}
			&\frac{\partial p}{\partial t}-\frac{1}{2}\Delta p=0, &  t\in(0,1],\,\,\, \bm{x}\in \mathbb{R}^2,\\
			&p(\bm{x},0)=\frac{1}{2\pi}\exp\bigg(-\frac{1}{2}\|\bm{x}-4\cdot\bm{1}_2\|^2\bigg),& \bm{x}\in \mathbb{R}^2,\\
		\end{aligned}
	\end{equation}
	and the exact solution yields
	\begin{equation}
		p(\bm{x},t)=\frac{1}{2\pi (t+1)}\exp\bigg(-\frac{\|\bm{x}-4\cdot\bm{1}_2\|^2}{2 (t+1)}\bigg).
	\end{equation}
The number of epochs is chosen as $N_e=20$, $\alpha=2$, batch size is set to 1000, and five adaptivity iterations are conducted for
this problem, i.e.,  $N_{\mathrm{adaptive}} = 5$. We take $L=6$ affine coupling layers and actnorm layers. In addition, we turn off the polynomial spline layer.
 The initial spatial training set is generated via the uniform distributed points in $[-3,3]^2$ (which only contains partial information of the exact solution), and the sample size is $N_{\bm{x}}=1000$ for each moment. The temporal training set is uniformly sampled in the unit interval $[0,1]$ with size $N_t=20$ which results in total $1000\times 20=20,000$ training points for each iteration step for $k=1,\dots, N_{\rm{adaptive}}$. 
	\begin{figure}[!h]
		\centering
		\begin{minipage}[t]{0.3\linewidth}
			\includegraphics[scale=0.31]{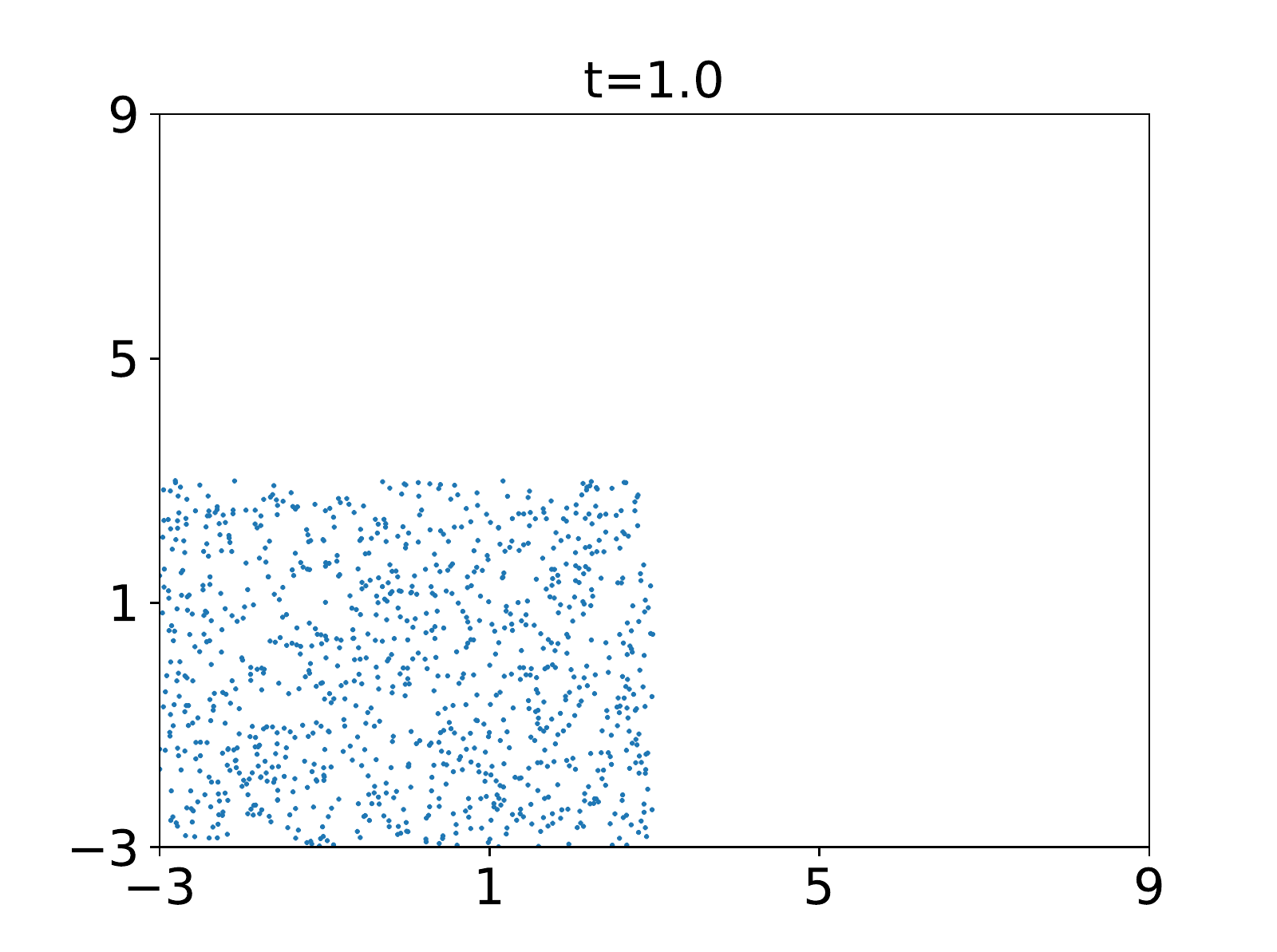}
			%		\subcaption{DGM-2}
		\end{minipage}
		\begin{minipage}[t]{0.3\linewidth}
				\includegraphics[scale=0.31]{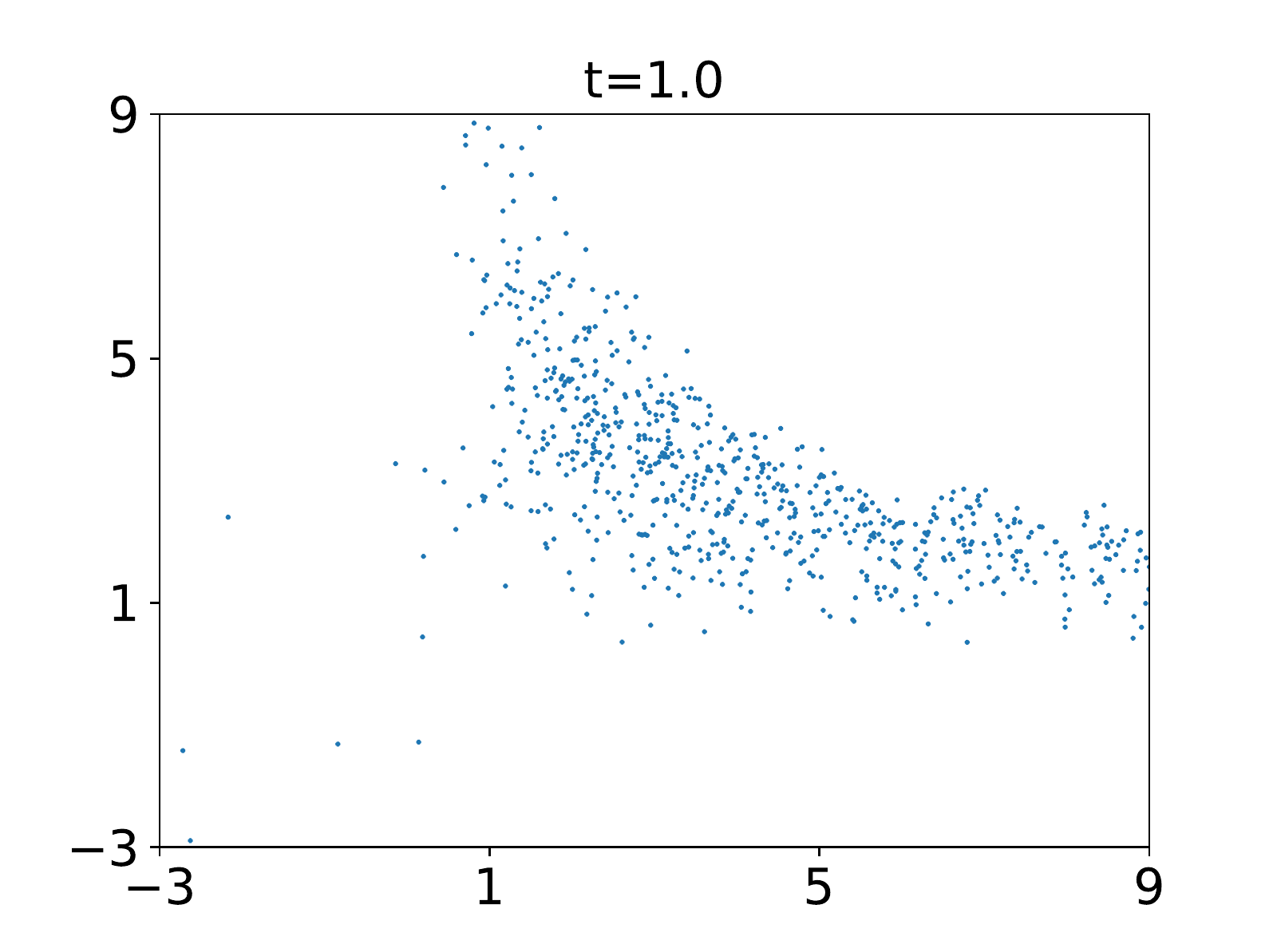}
			%		\subcaption{DGM-2-uncouple}
		\end{minipage}
		\begin{minipage}[t]{0.3\linewidth}
			\includegraphics[scale=0.31]{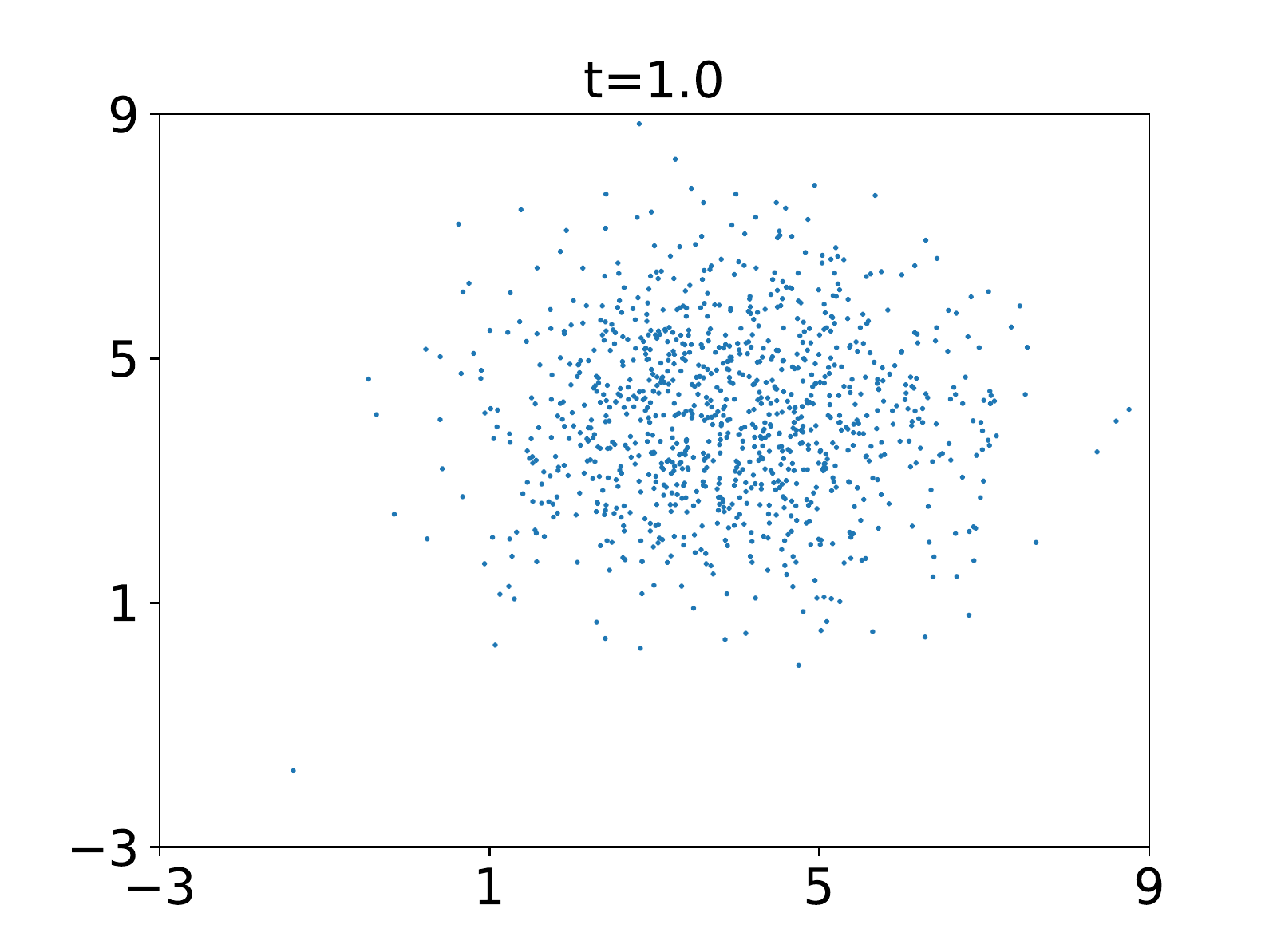}
			%		\subcaption{DGM-2-couple}
		\end{minipage}
		\caption{A toy example. Training samples at different adaptive iterations. Left panel: $N_{\mathrm{adaptive}}=1$. Middle panel: $N_{\mathrm{adaptive}}=2$. Right panel: $N_{\mathrm{adaptive}}=4$.}
		\label{ex0_sample}
	\end{figure}

\begin{figure}[!h]
	\centering
	\begin{minipage}[t]{0.3\linewidth}
		\includegraphics[scale=0.3]{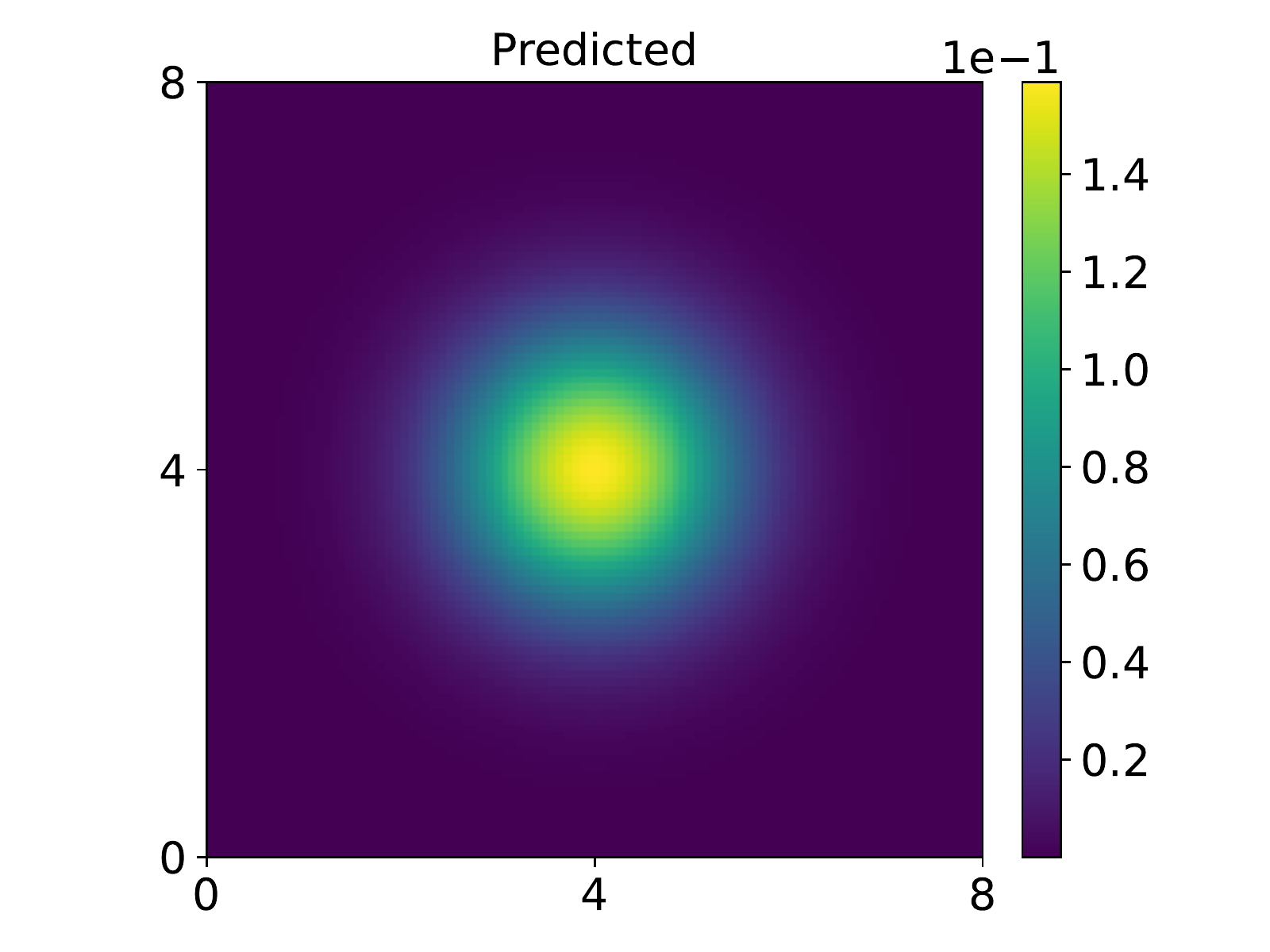}
		%		\subcaption{DGM-2}
	\end{minipage}
	\begin{minipage}[t]{0.3\linewidth}
		\includegraphics[scale=0.3]{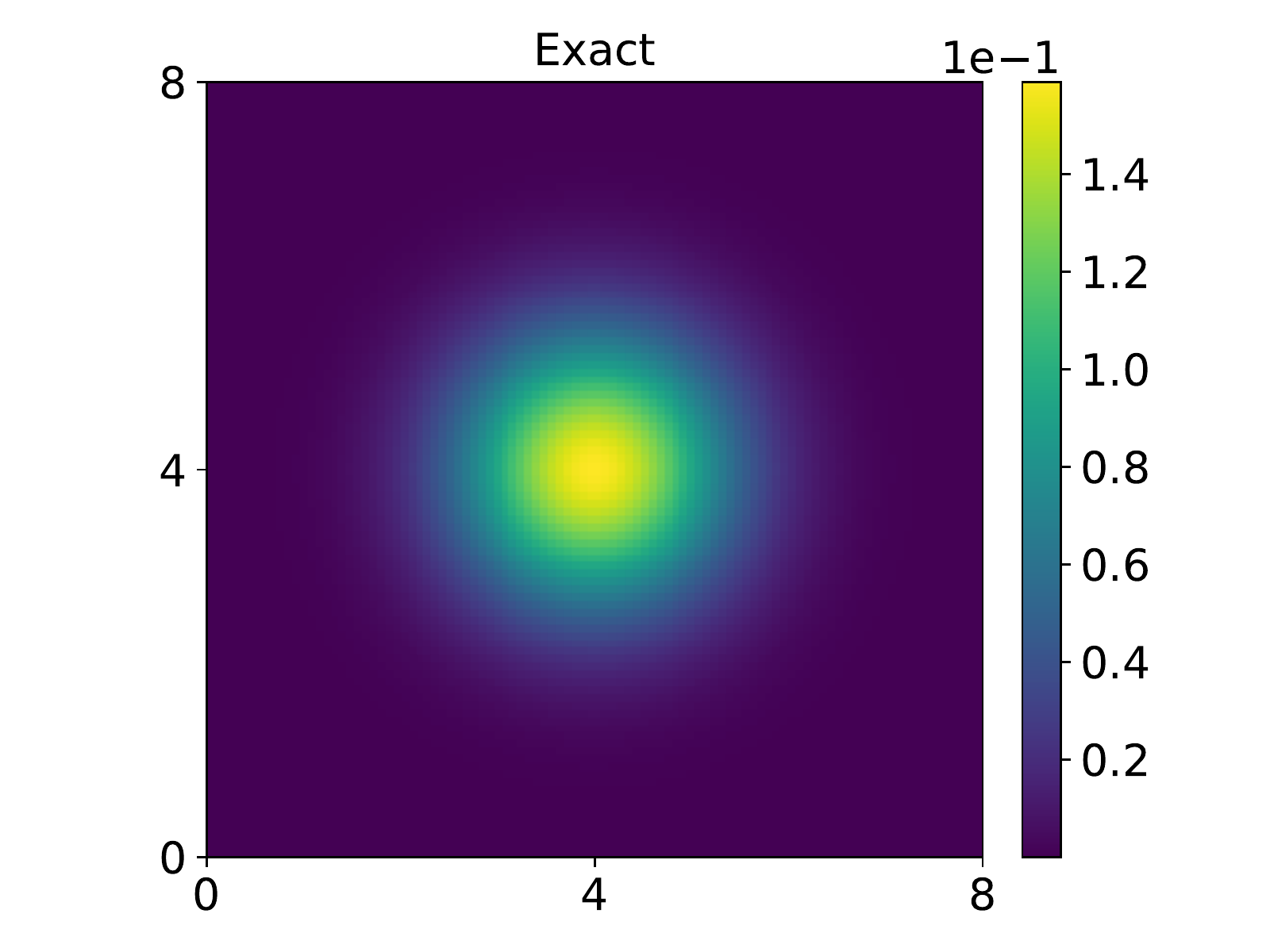}
		%		\subcaption{DGM-2-uncouple}
	\end{minipage}
	\begin{minipage}[t]{0.3\linewidth}
		\includegraphics[scale=0.3]{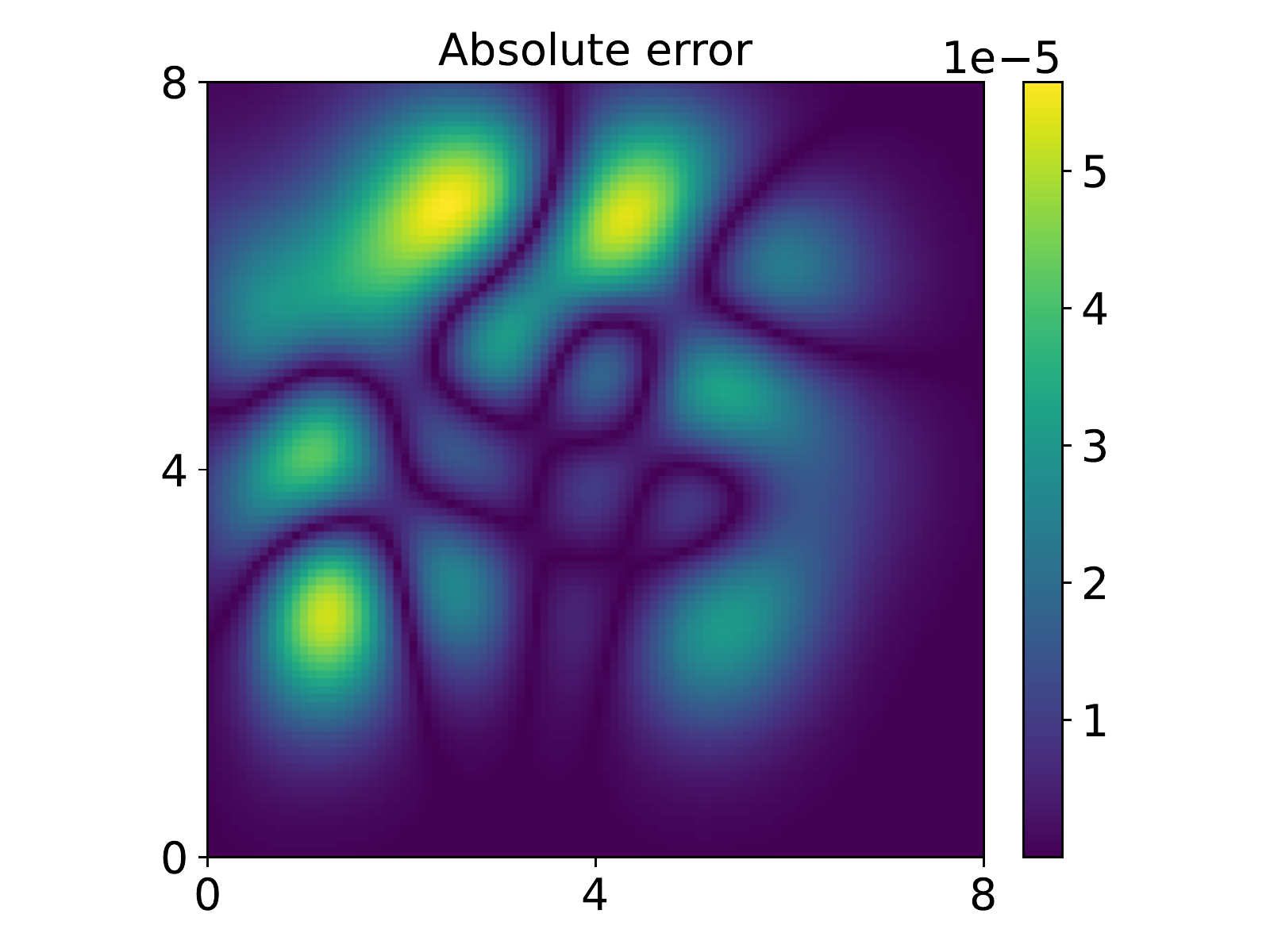}
		%		\subcaption{DGM-2-couple}
	\end{minipage}
	
	\begin{minipage}[t]{0.3\linewidth}
		\includegraphics[scale=0.3]{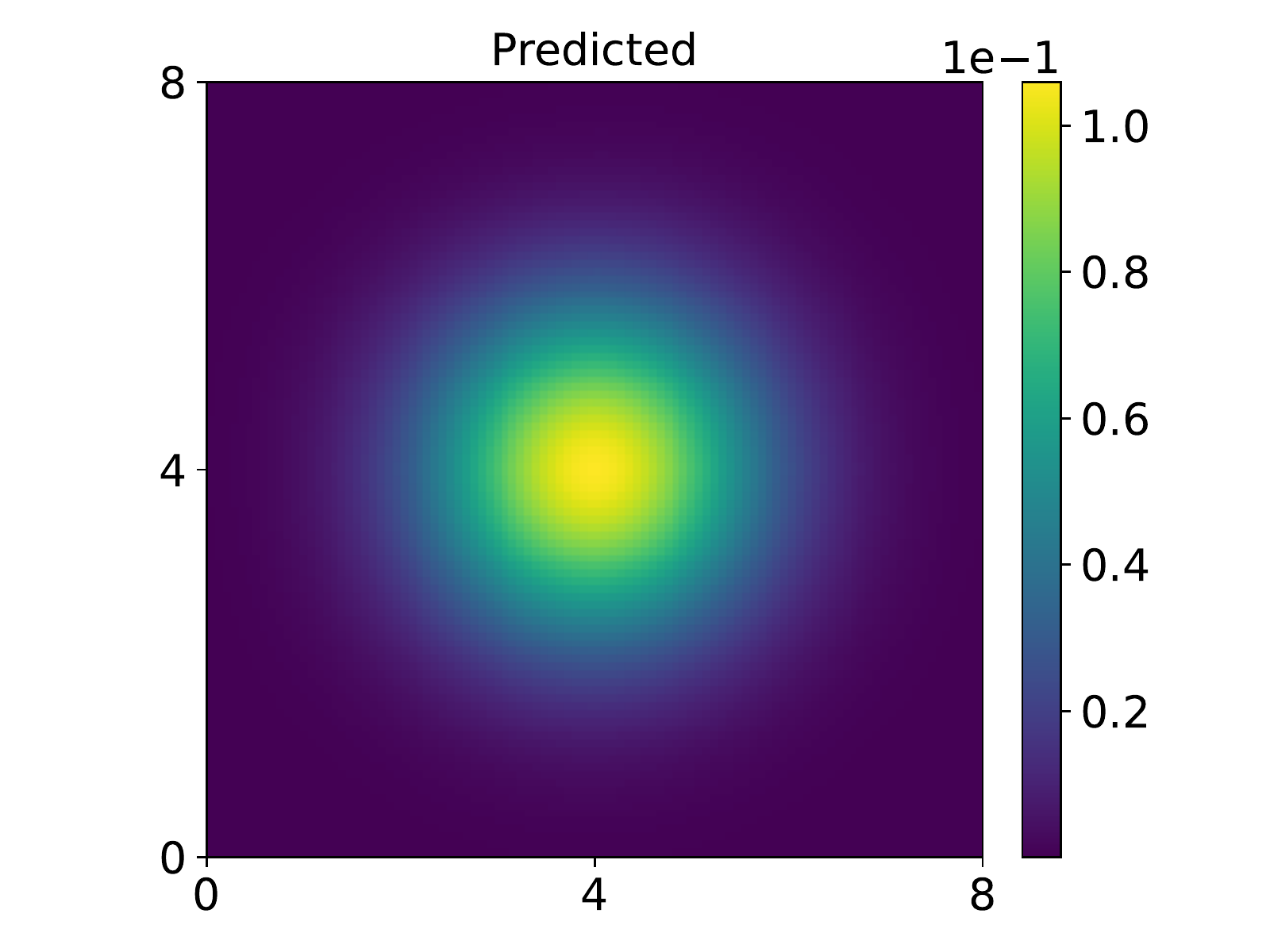}
		%		\subcaption{DGM-2}
	\end{minipage}
	\begin{minipage}[t]{0.3\linewidth}
		\includegraphics[scale=0.3]{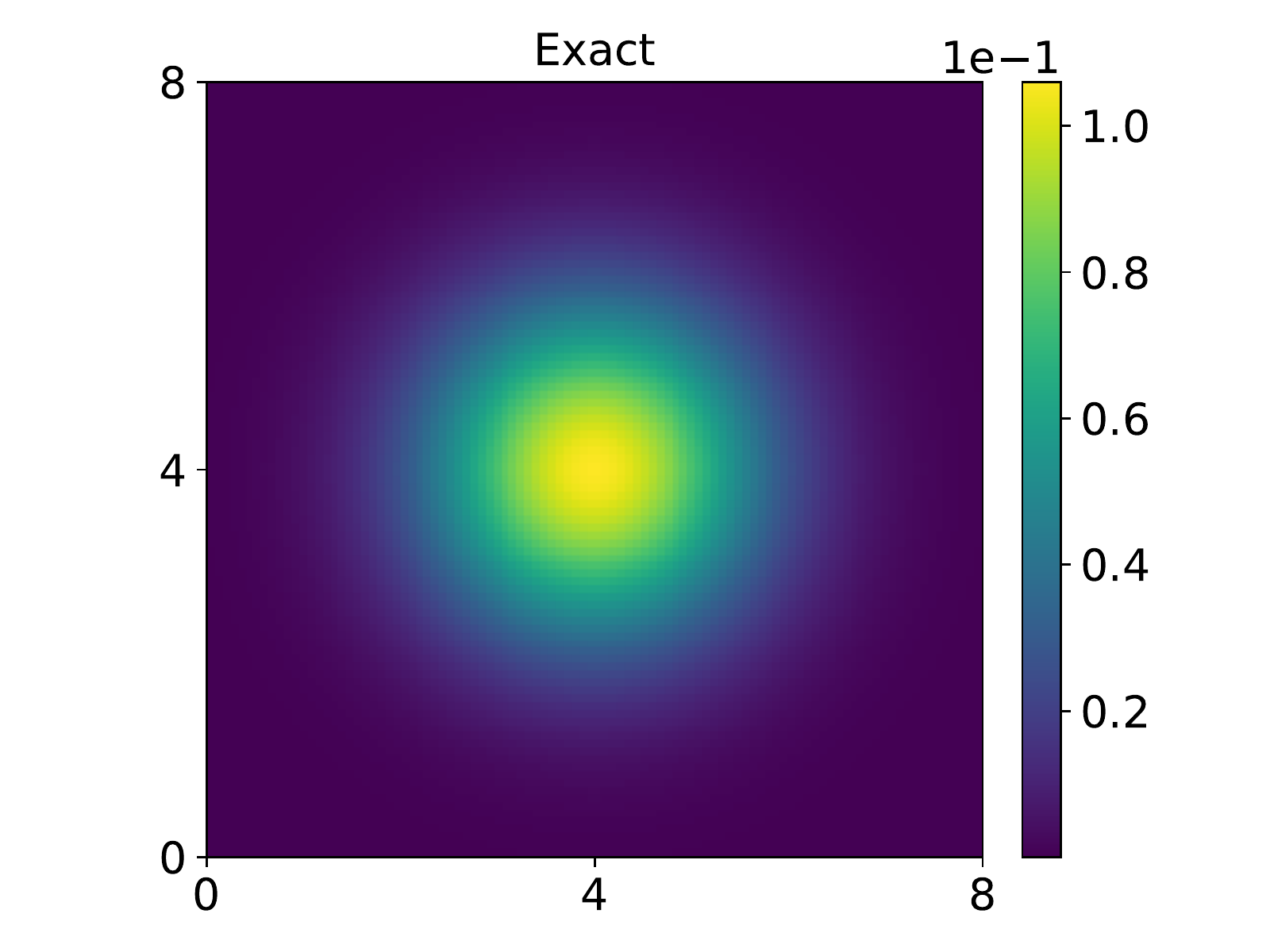}
		%		\subcaption{DGM-2-uncouple}
	\end{minipage}
	\begin{minipage}[t]{0.3\linewidth}
		\includegraphics[scale=0.3]{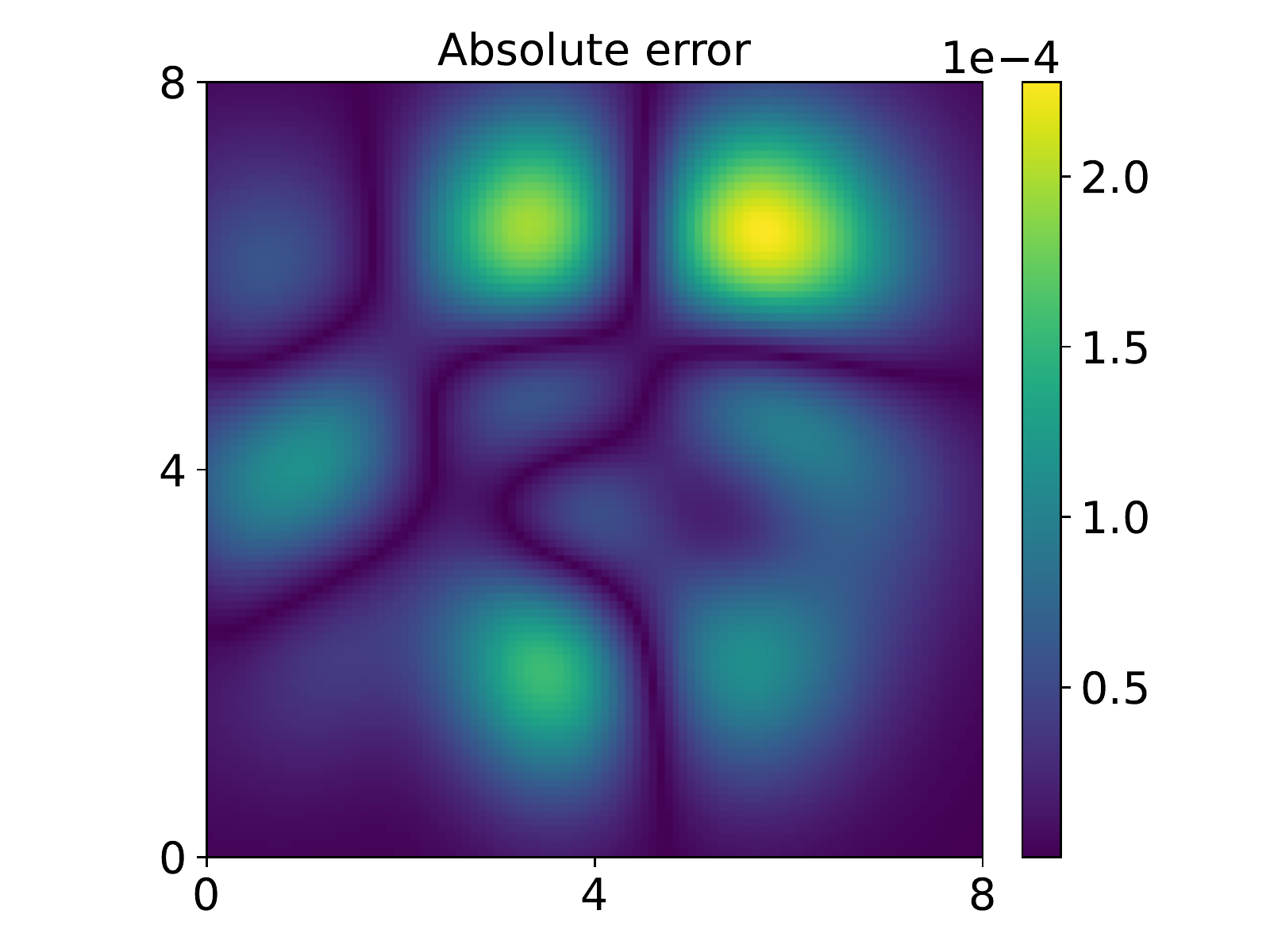}
		%		\subcaption{DGM-2-couple}
	\end{minipage}
	
	\begin{minipage}[t]{0.3\linewidth}
		\includegraphics[scale=0.3]{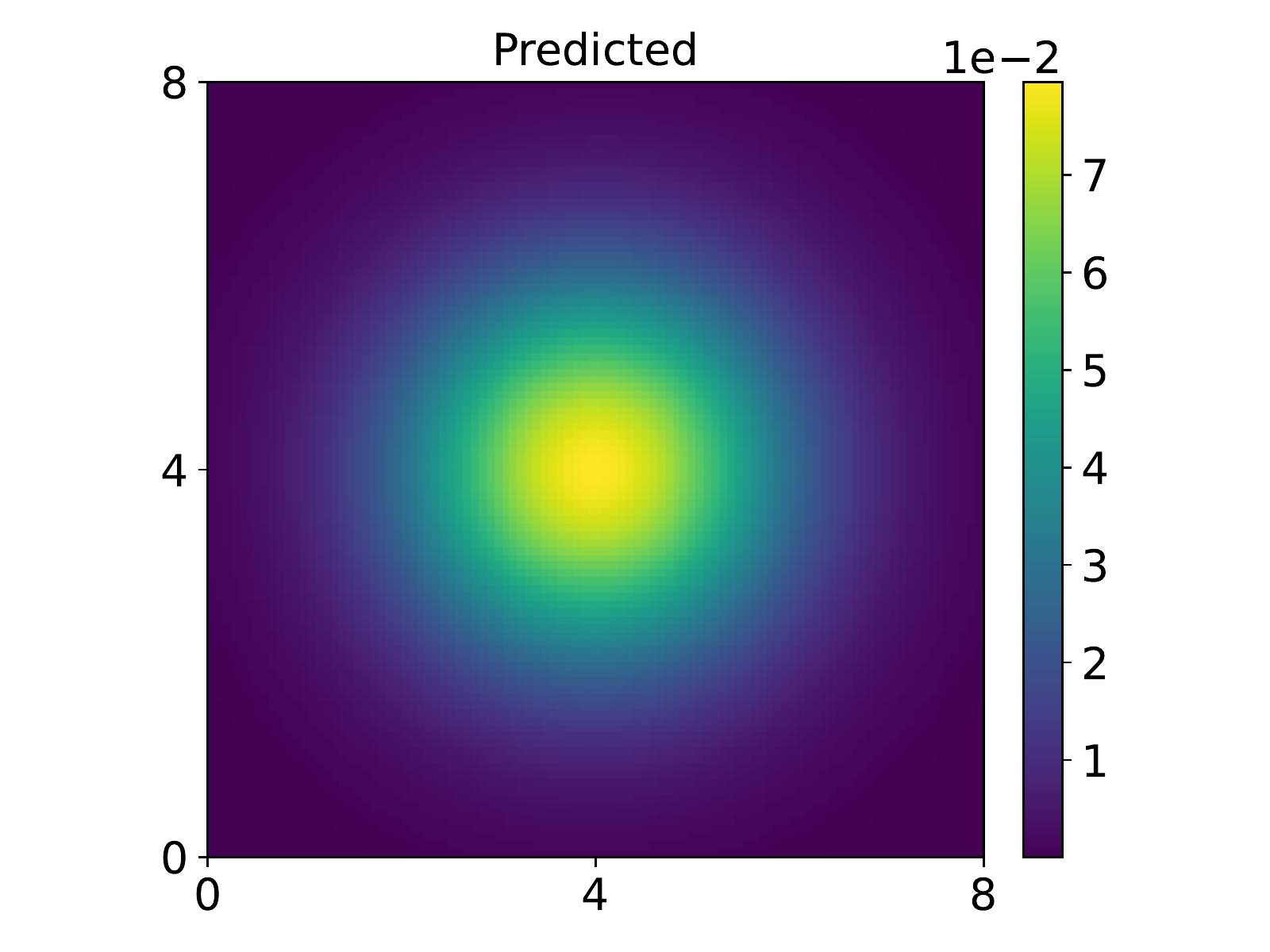}
		%		\subcaption{DGM-2}
	\end{minipage}
	\begin{minipage}[t]{0.3\linewidth}
		\includegraphics[scale=0.3]{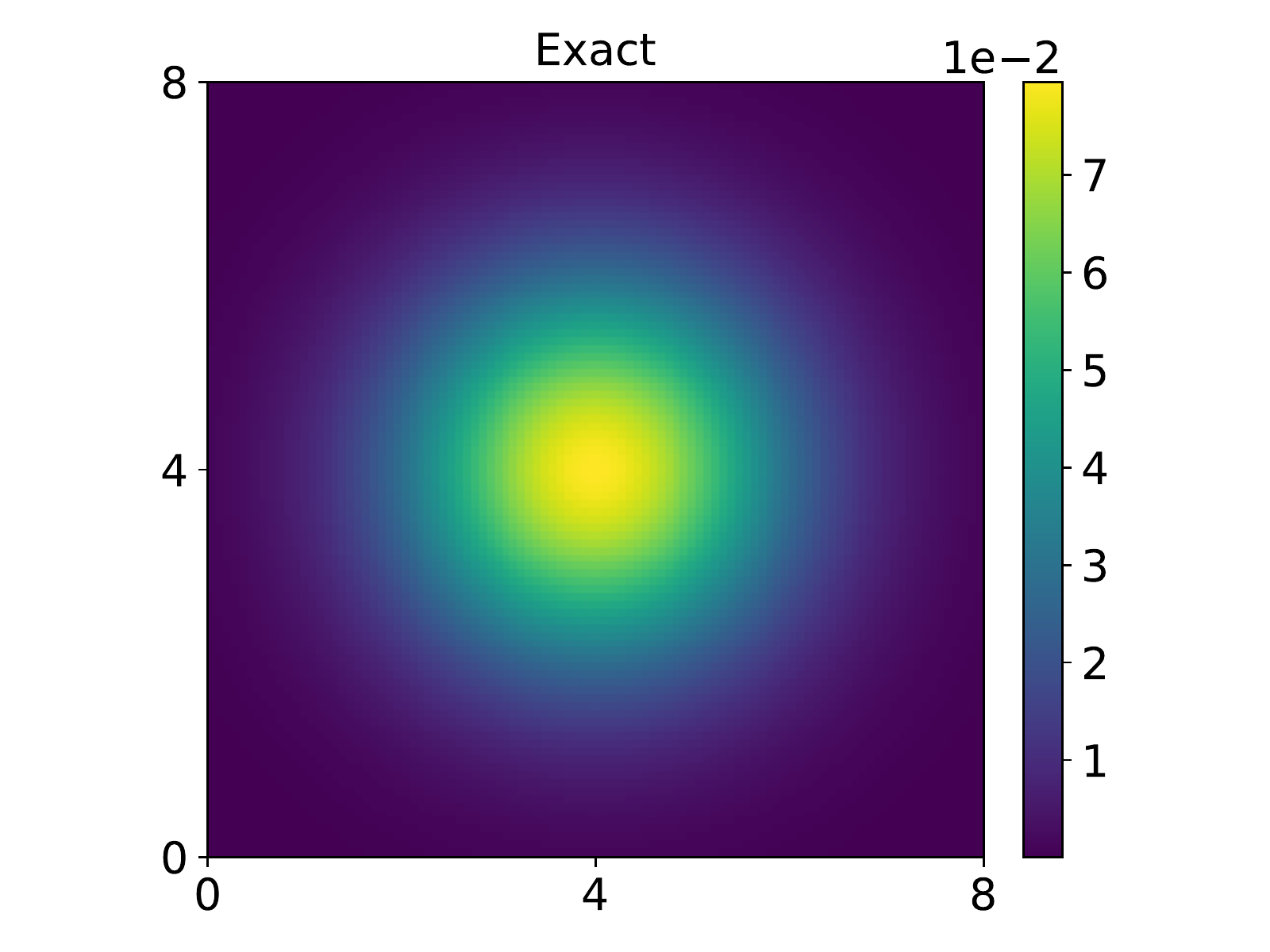}
		%		\subcaption{DGM-2-uncouple}
	\end{minipage}
	\begin{minipage}[t]{0.3\linewidth}
		\includegraphics[scale=0.3]{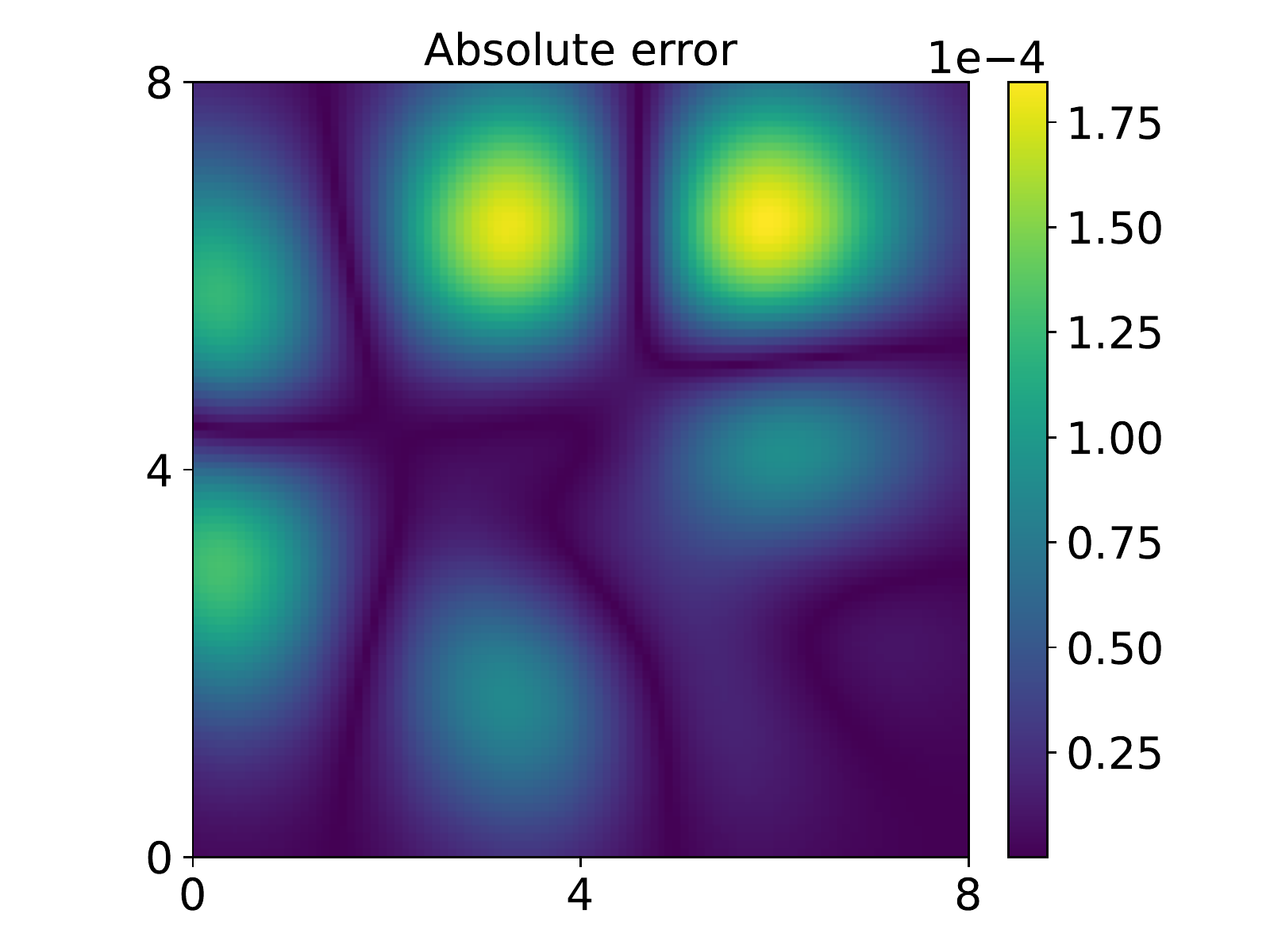}
		%		\subcaption{DGM-2-couple}
	\end{minipage}
	
	\caption{A toy example. Predicted solution versus the reference solution for different time $t$. Top row: $t=0$. Middle row: $t=0.5$. Bottom row: $t=1.$}
	\label{ex0_results}
\end{figure}

\begin{figure}
	\centering
	\includegraphics[scale=0.4]{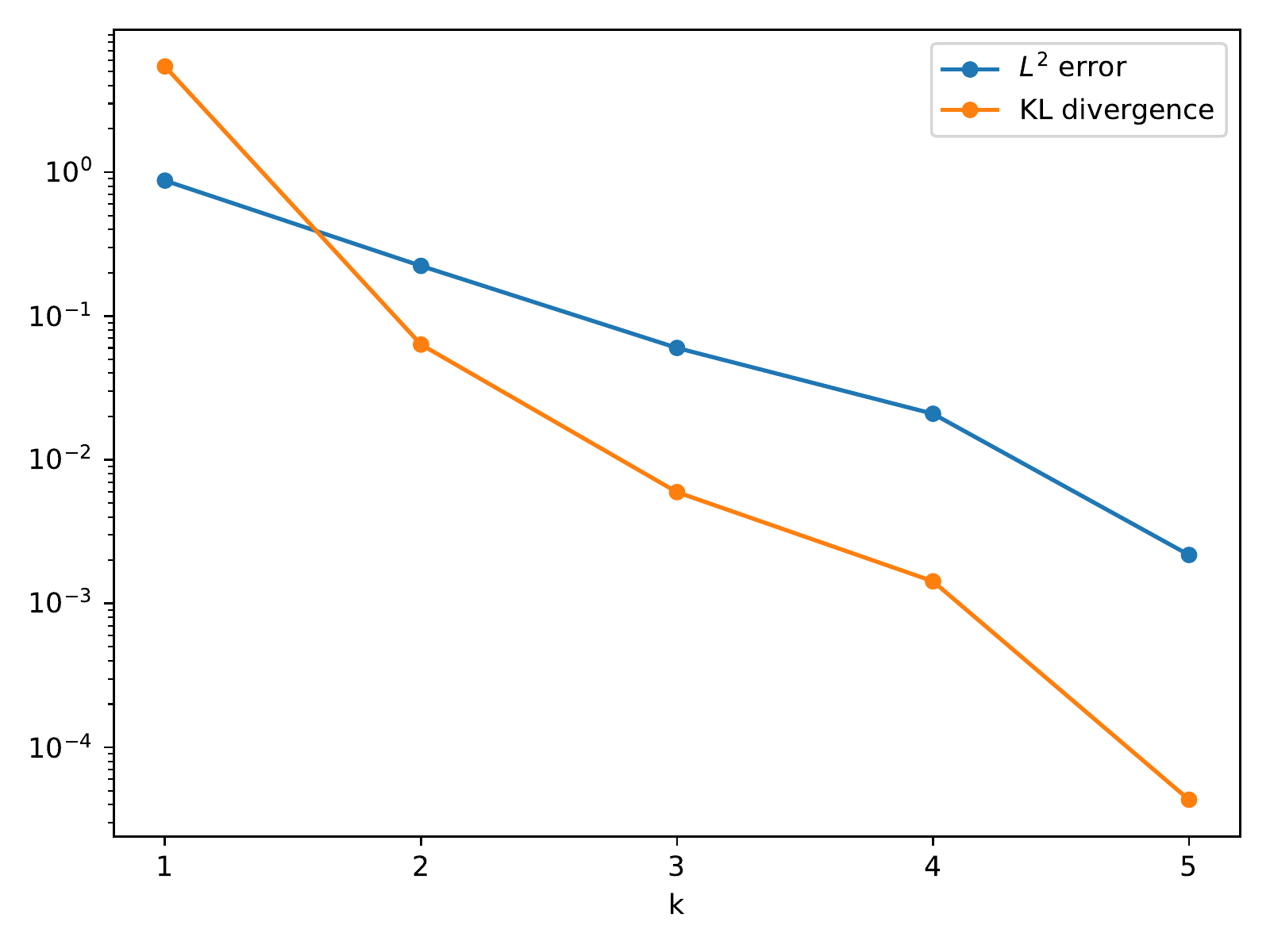}
	\caption{A toy example. Relative $L^2$ error and relative KL divergence at $t=1$ for different adaptive iterations $k$.}
	\label{fig:toy_err}
\end{figure}

The training points for different adaptive iterations at time $t=1$ are presented in Figure \ref{ex0_sample}. One can clearly observe that, the training points become increasingly closer to the ground truth as adaptive iterations increase, which shows that the adaptive sampling scheme is rather effective. The predicted solution and the exact solution are presented in Figure \ref{ex0_results} and these figures indicate that the predicted solution yields an excellent agreement with the exact solution. The relative $L^2$ error and relative KL divergence with different adaptive iterations are also provided in Figure \ref{fig:toy_err}.

%
%	\begin{figure}[!h]
%	\centering
%	\begin{minipage}[t]{0.3\linewidth}
%		\includegraphics[scale=0.3]{./fig/example0/l2_err.pdf}
%		%		\subcaption{DGM-2}
%	\end{minipage}
%	\begin{minipage}[t]{0.3\linewidth}
%		\includegraphics[scale=0.3]{./fig/example0/kl_err.pdf}
%		%		\subcaption{DGM-2-uncouple}
%	\end{minipage}
%	\caption{A toy example. Relative $L^2$ error versus relative KL divergence at different adaptive iterations $k$. Left panel: relative $L^2$ error. Right panel: relative KL divergence.}
%	\label{ex0_serr}
%\end{figure}

\subsection{Linear oscillator}
We next consider the TFP equation (\ref{time_depe_fp_eqn}) with
 \begin{equation}
	\bm{\mu} = (x_2,\; -0.2x_2-x_1), \quad \bm{D}=\mathrm{diag}(0,\;0.2).
\end{equation}
The initial condition is given by
\begin{equation}
	p(\bm{x},0) = \mathcal{N}\bigg((1,1),\; \frac{1}{9}\bm{I}_2\bigg),
\end{equation}
where $\mathcal{N}$ denotes Gaussian distribution. We consider that the time interval is $[0, 3]$ and the "exact" solution of this example is computed by the ADI scheme \cite{pichler2013numerical} in a truncated spatial domain $[-5, 5]^2$ with mesh size $\delta h = 0.01$ and $\delta t=0.01$. For our approach, we use $N_e=60$ and $\alpha=1$. For now, $1000$ batch size and four adaptivity iterations are conducted for
this problem, i.e.,  $N_{\mathrm{adaptive}} = 4$. We take $L=8$ affine coupling layers and actnorm layers, and turn off the polynomial spline layer. The initial spatial training set is generated through the uniform distribution with a range $[-5,5]^2$, and the corresponding initial temporal training set is uniformly sampled in the interval $[0,3]$, which results in total $2000\times 100= 200,000$ training points.

\begin{figure}[!h]
	\centering
	\begin{minipage}[t]{0.3\linewidth}
		\includegraphics[scale=0.3]{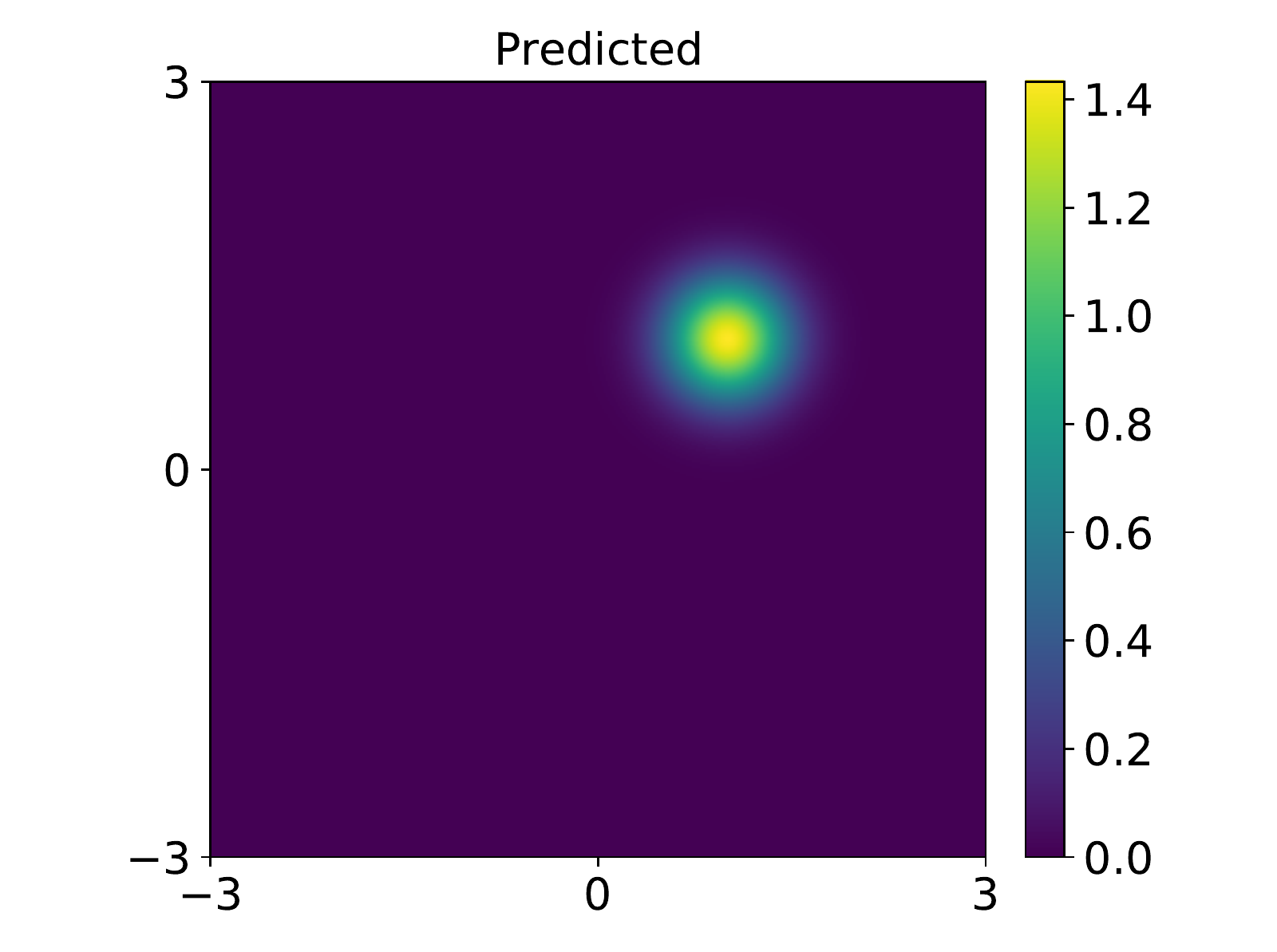}
		%		\subcaption{DGM-2}
	\end{minipage}
	\begin{minipage}[t]{0.3\linewidth}
		\includegraphics[scale=0.3]{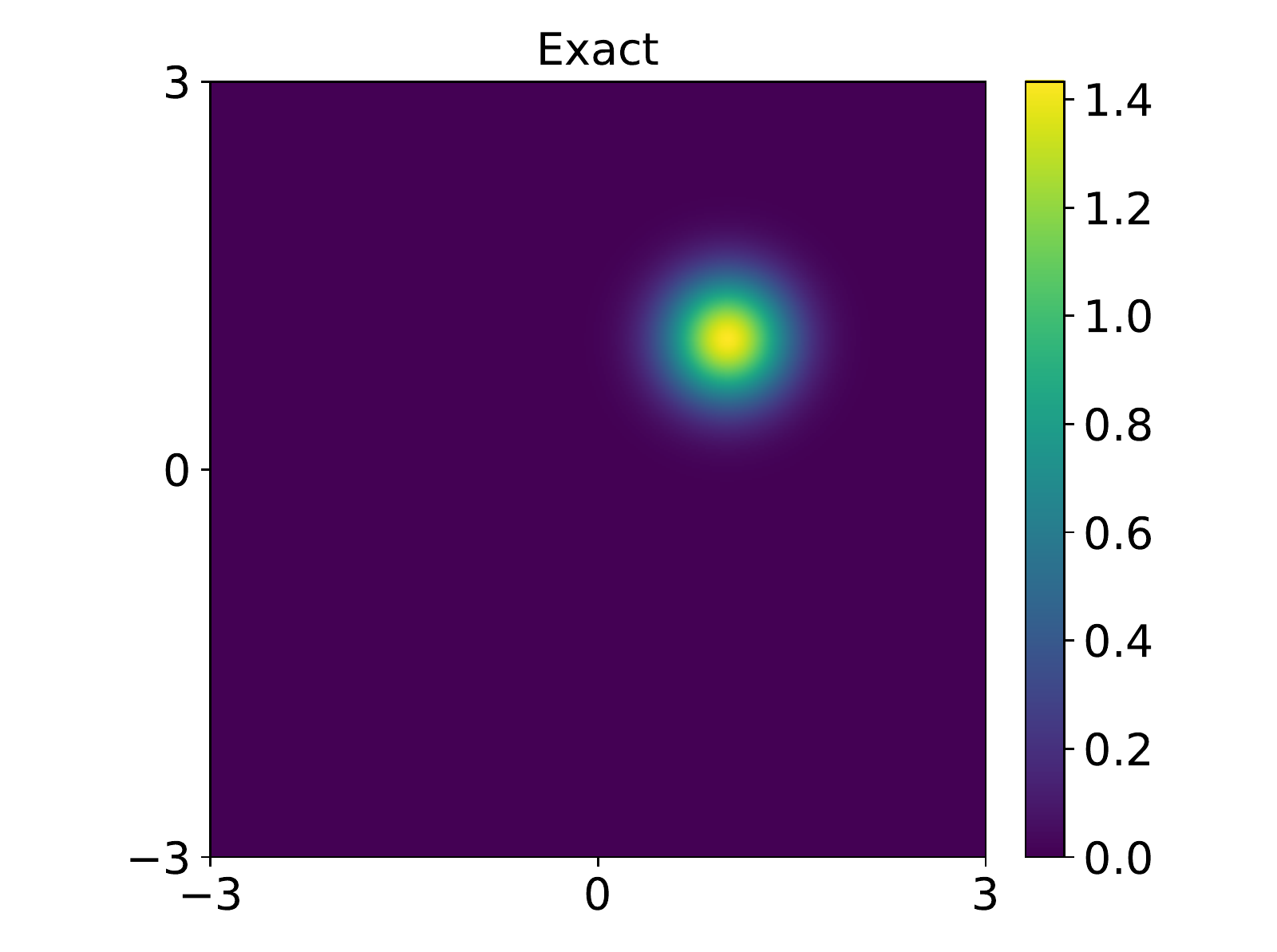}
		%		\subcaption{DGM-2-uncouple}
	\end{minipage}
	\begin{minipage}[t]{0.3\linewidth}
		\includegraphics[scale=0.3]{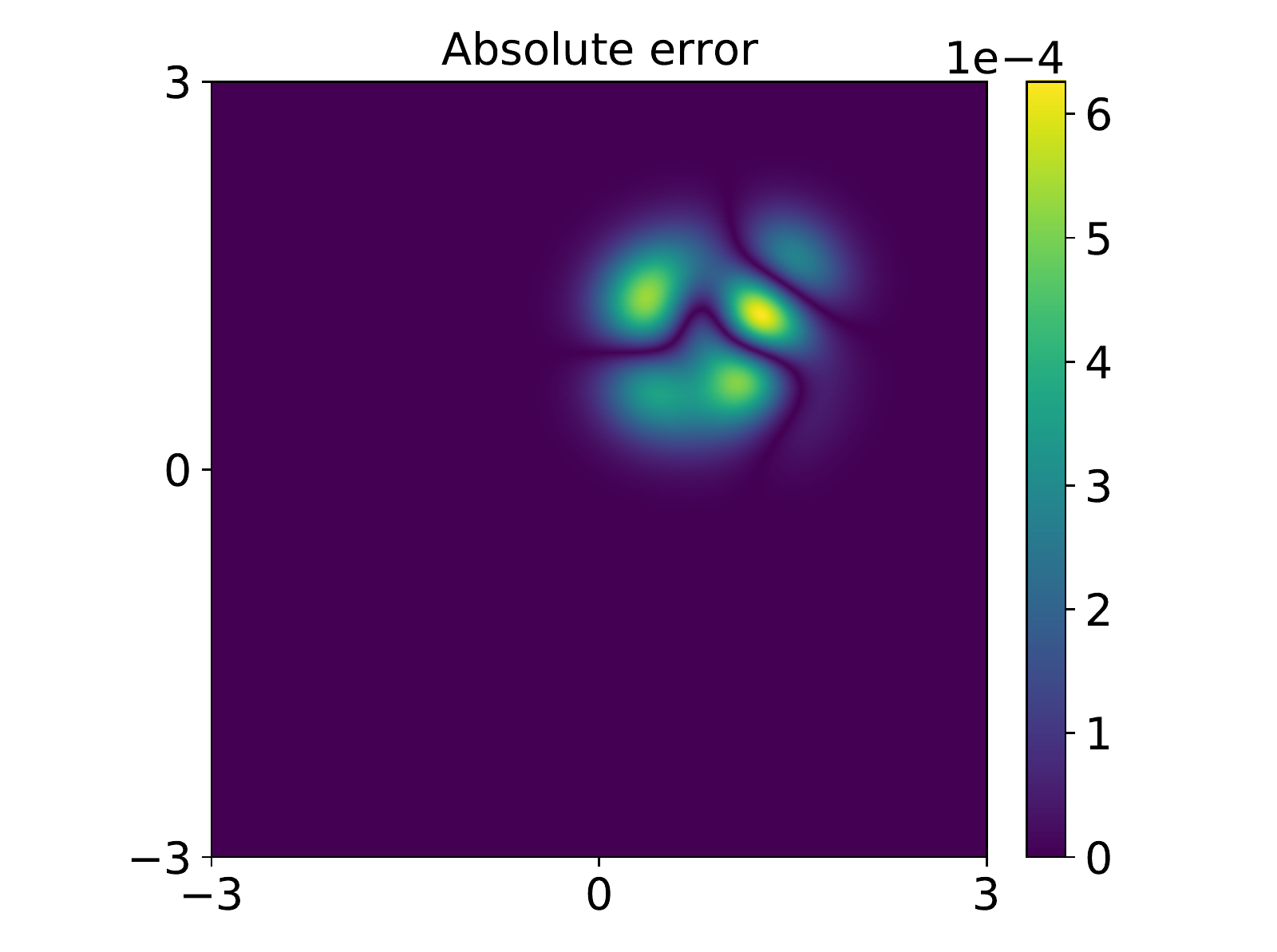}
		%		\subcaption{DGM-2-couple}
	\end{minipage}
	
	\begin{minipage}[t]{0.3\linewidth}
		\includegraphics[scale=0.3]{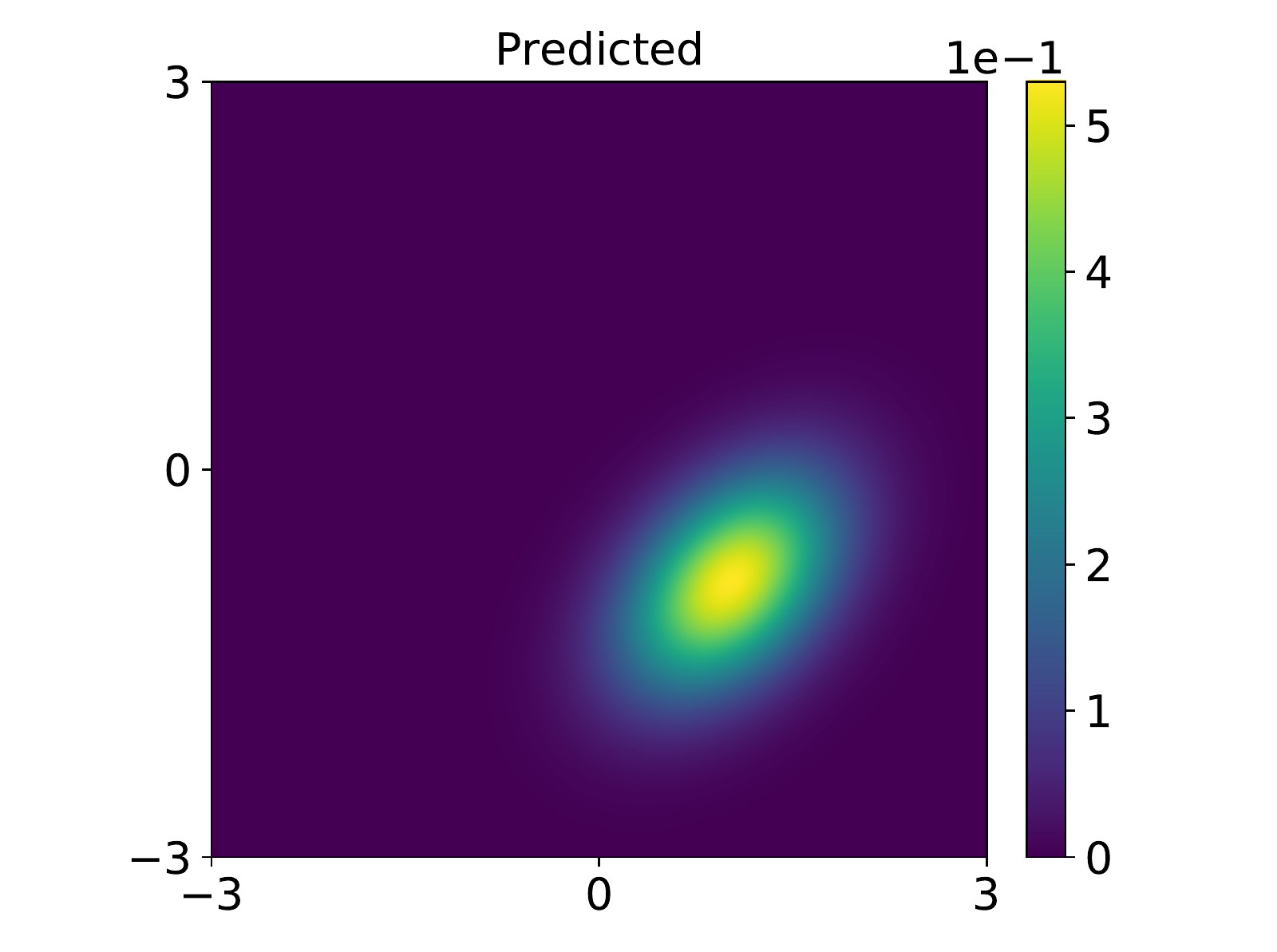}
		%		\subcaption{DGM-2}
	\end{minipage}
	\begin{minipage}[t]{0.3\linewidth}
		\includegraphics[scale=0.3]{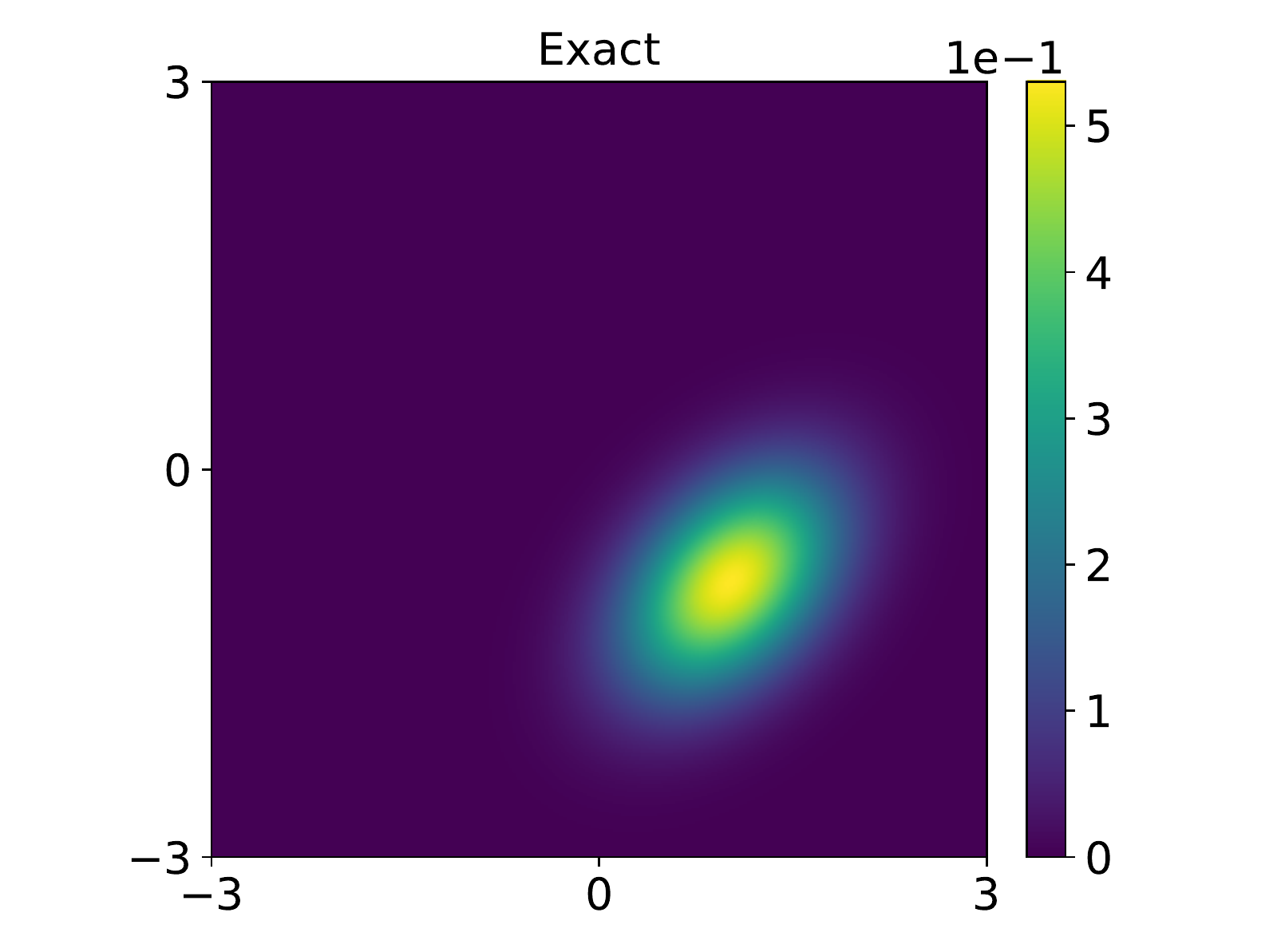}
		%		\subcaption{DGM-2-uncouple}
	\end{minipage}
	\begin{minipage}[t]{0.3\linewidth}
		\includegraphics[scale=0.3]{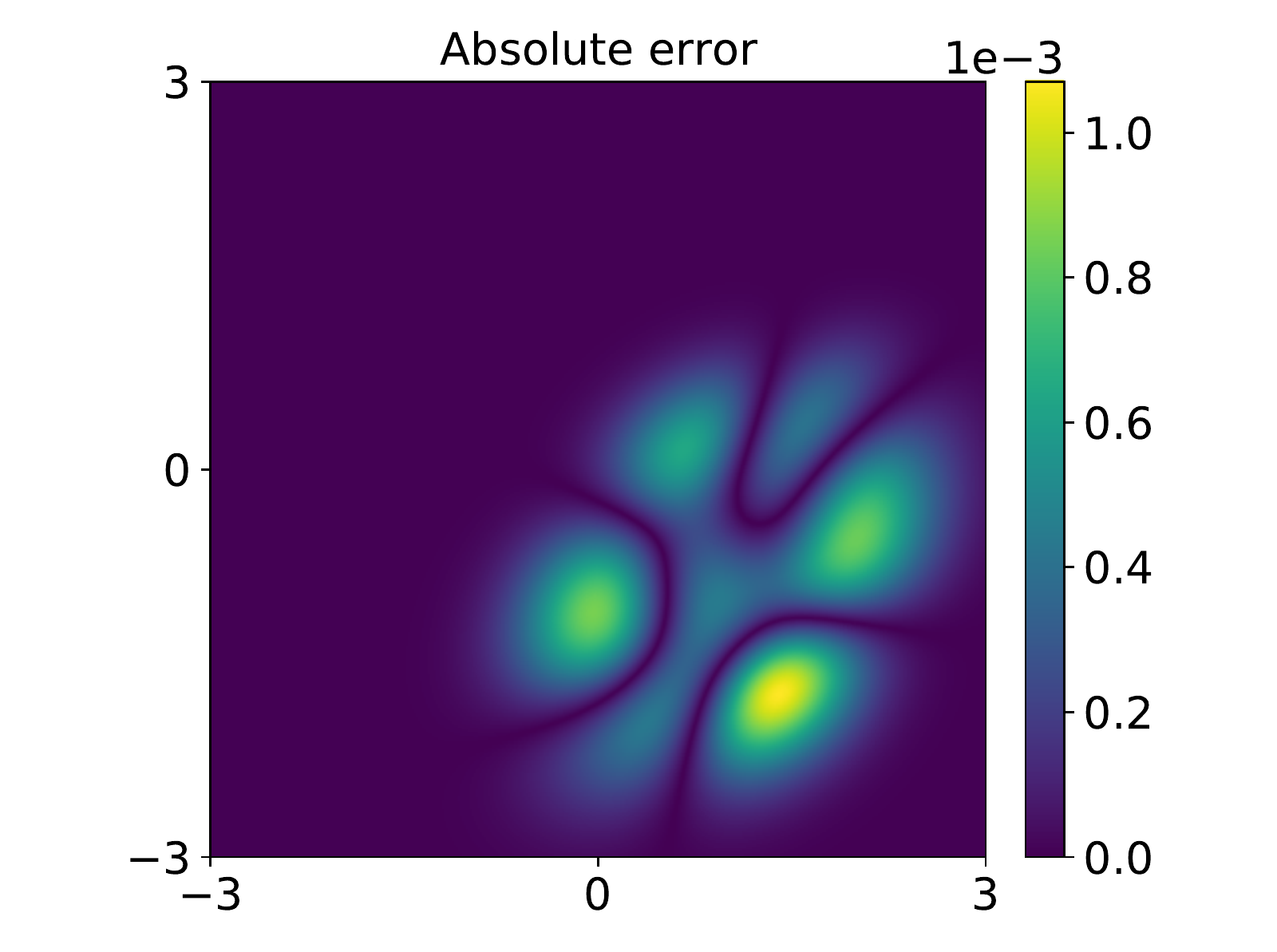}
		%		\subcaption{DGM-2-couple}
	\end{minipage}
	
	\begin{minipage}[t]{0.3\linewidth}
		\includegraphics[scale=0.3]{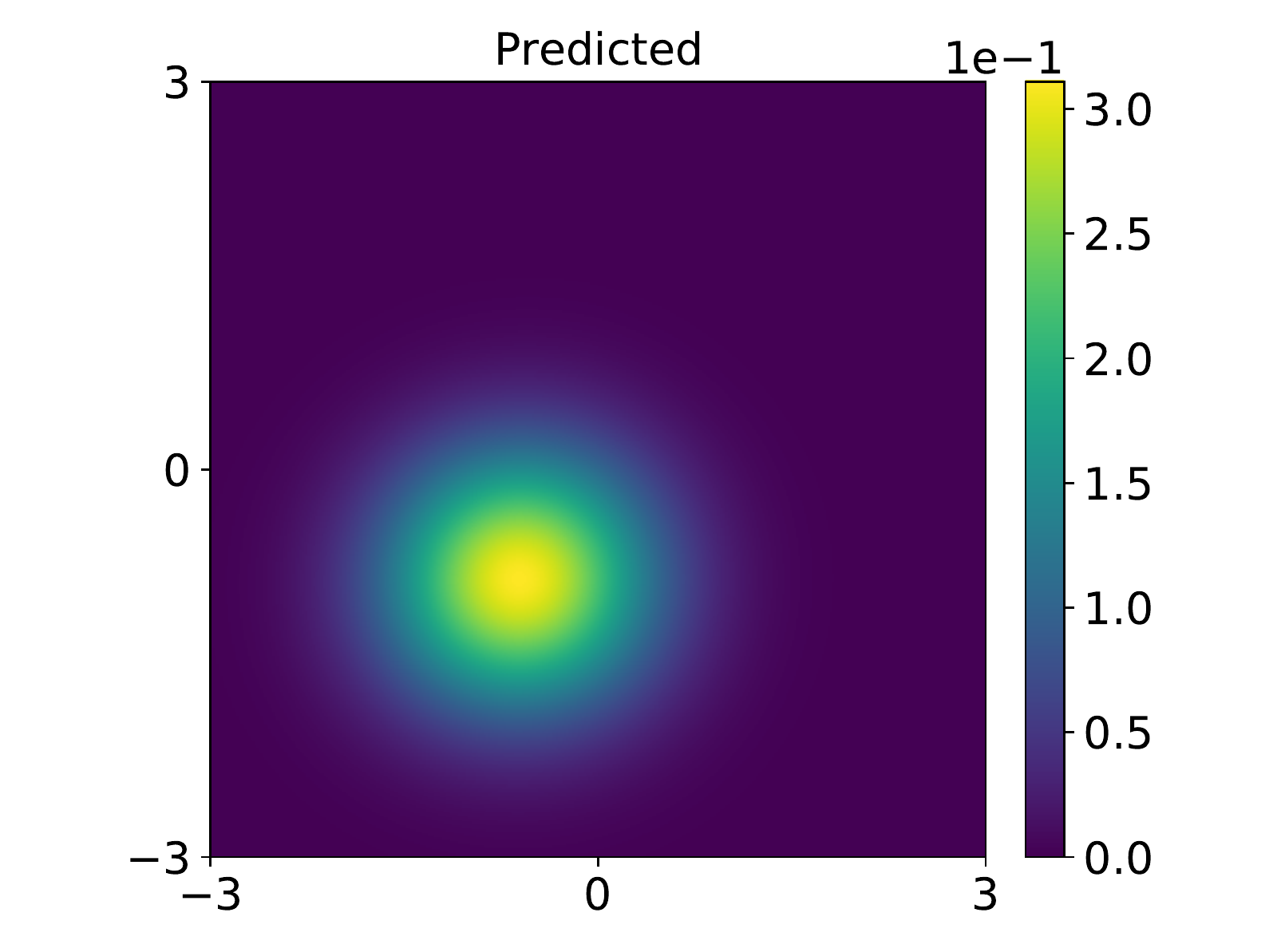}
		%		\subcaption{DGM-2}
	\end{minipage}
	\begin{minipage}[t]{0.3\linewidth}
		\includegraphics[scale=0.3]{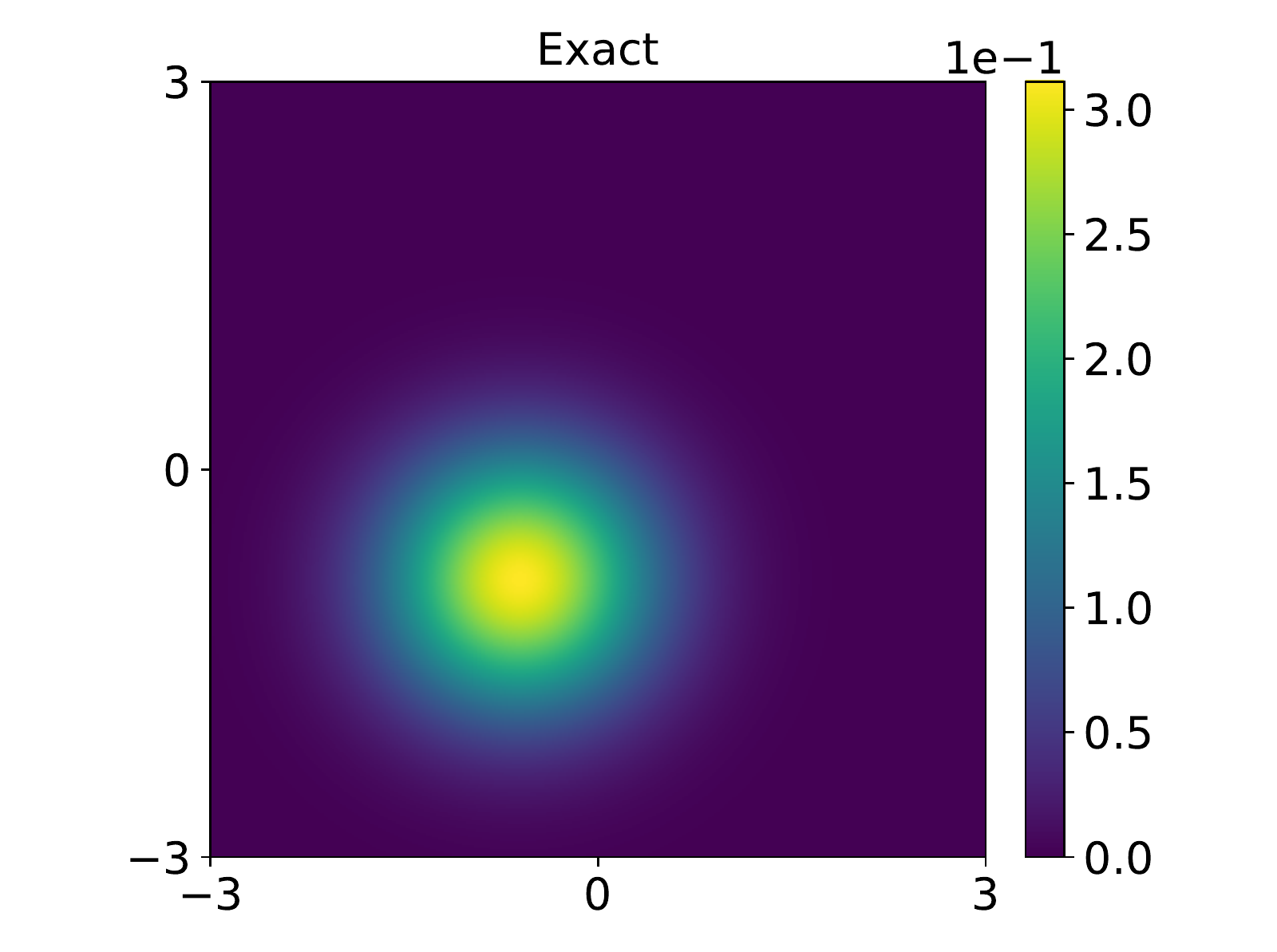}
		%		\subcaption{DGM-2-uncouple}
	\end{minipage}
	\begin{minipage}[t]{0.3\linewidth}
		\includegraphics[scale=0.3]{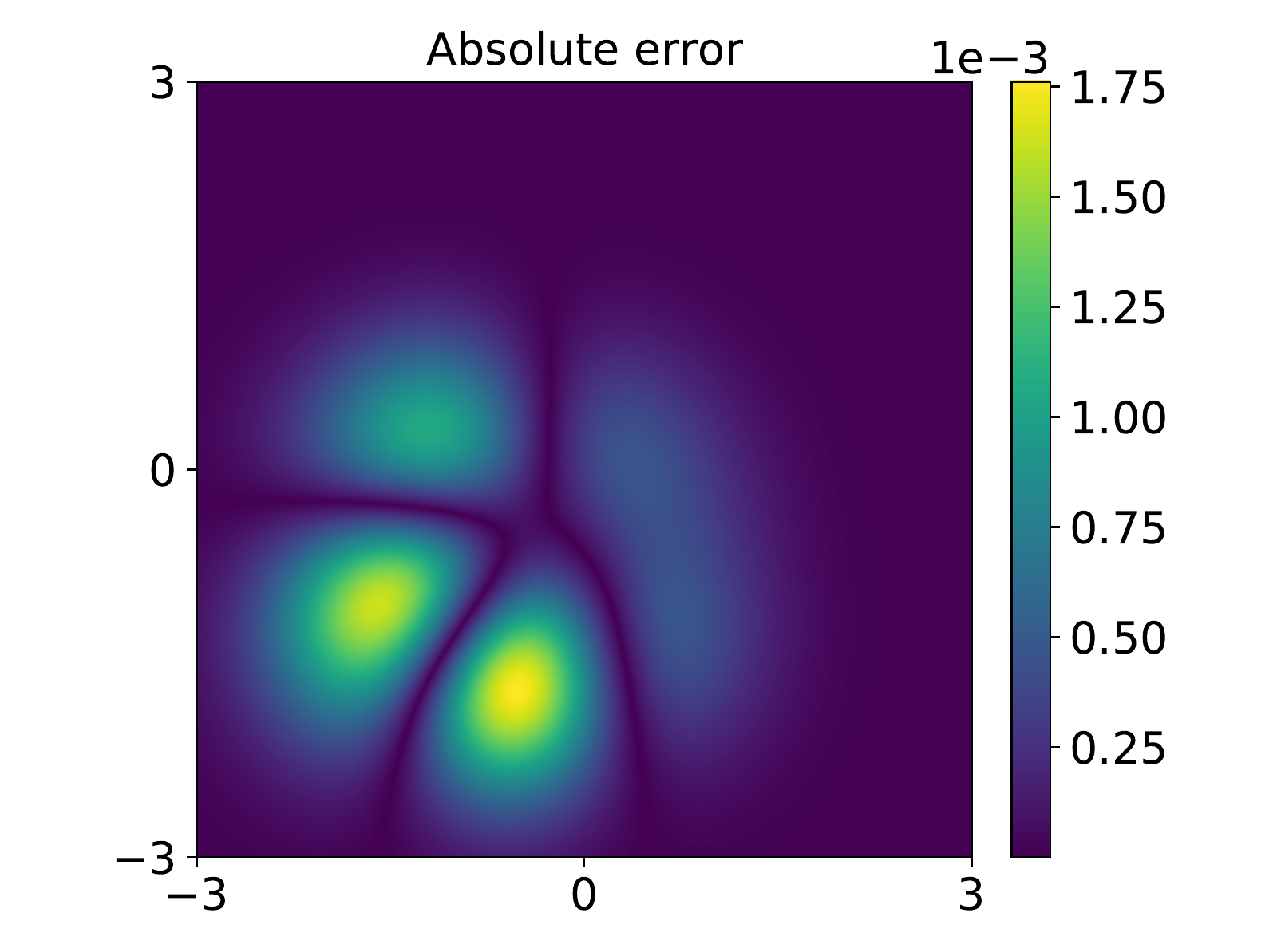}
		%		\subcaption{DGM-2-couple}
	\end{minipage}
	
	\caption{Linear oscillator. Predicted solution versus the reference solution for different time $t$. Top row: $t=0$. Middle row: $t=1.5$. Bottom row: $t=3.$}
	\label{linear_result}
\end{figure}

\begin{figure}
	\centering
	\begin{minipage}[t]{0.4\linewidth}
		\includegraphics[scale=0.4]{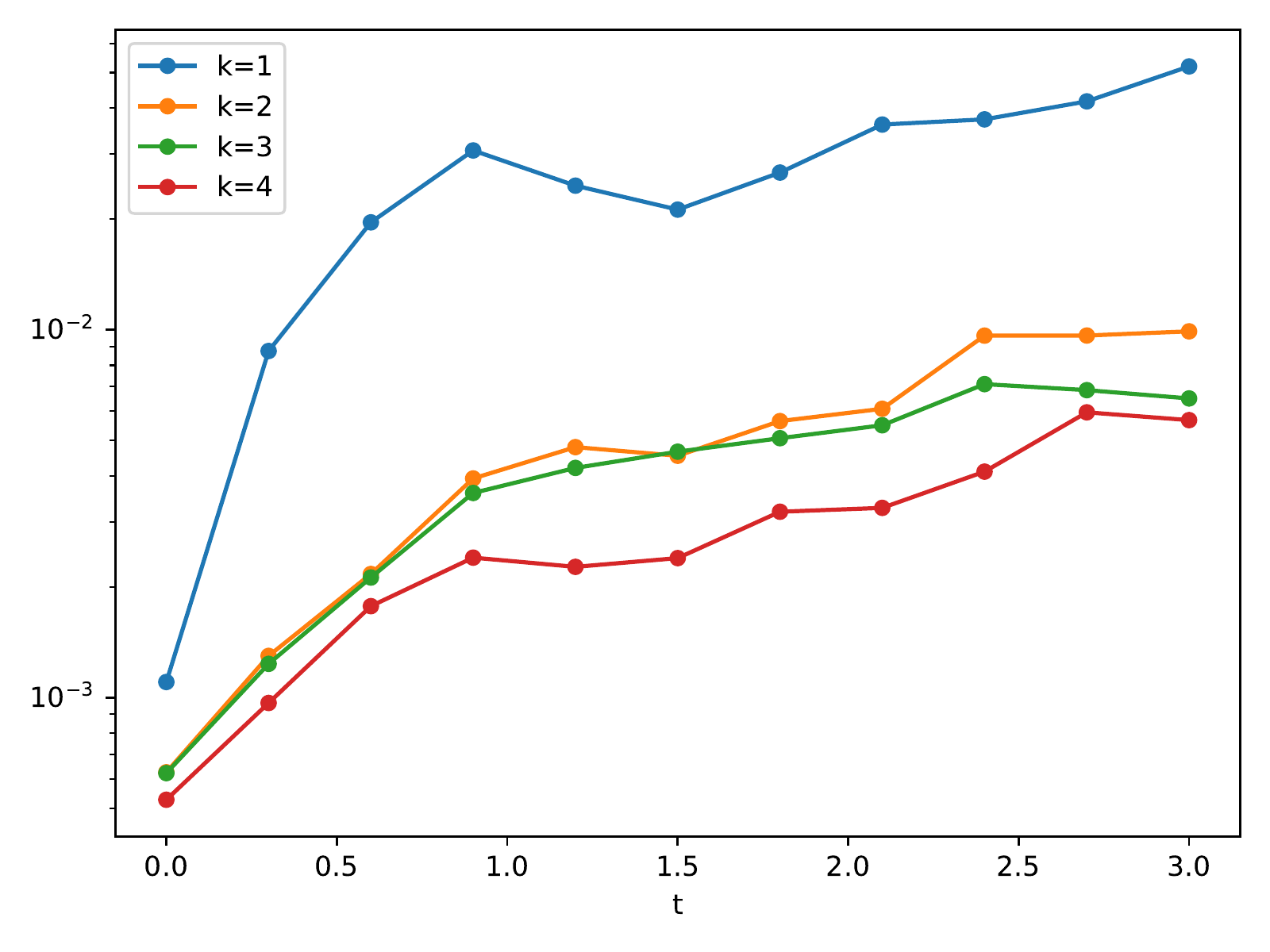}
		%		\subcaption{DGM-2-uncouple}
	\end{minipage}
	\begin{minipage}[t]{0.4\linewidth}
		\includegraphics[scale=0.4]{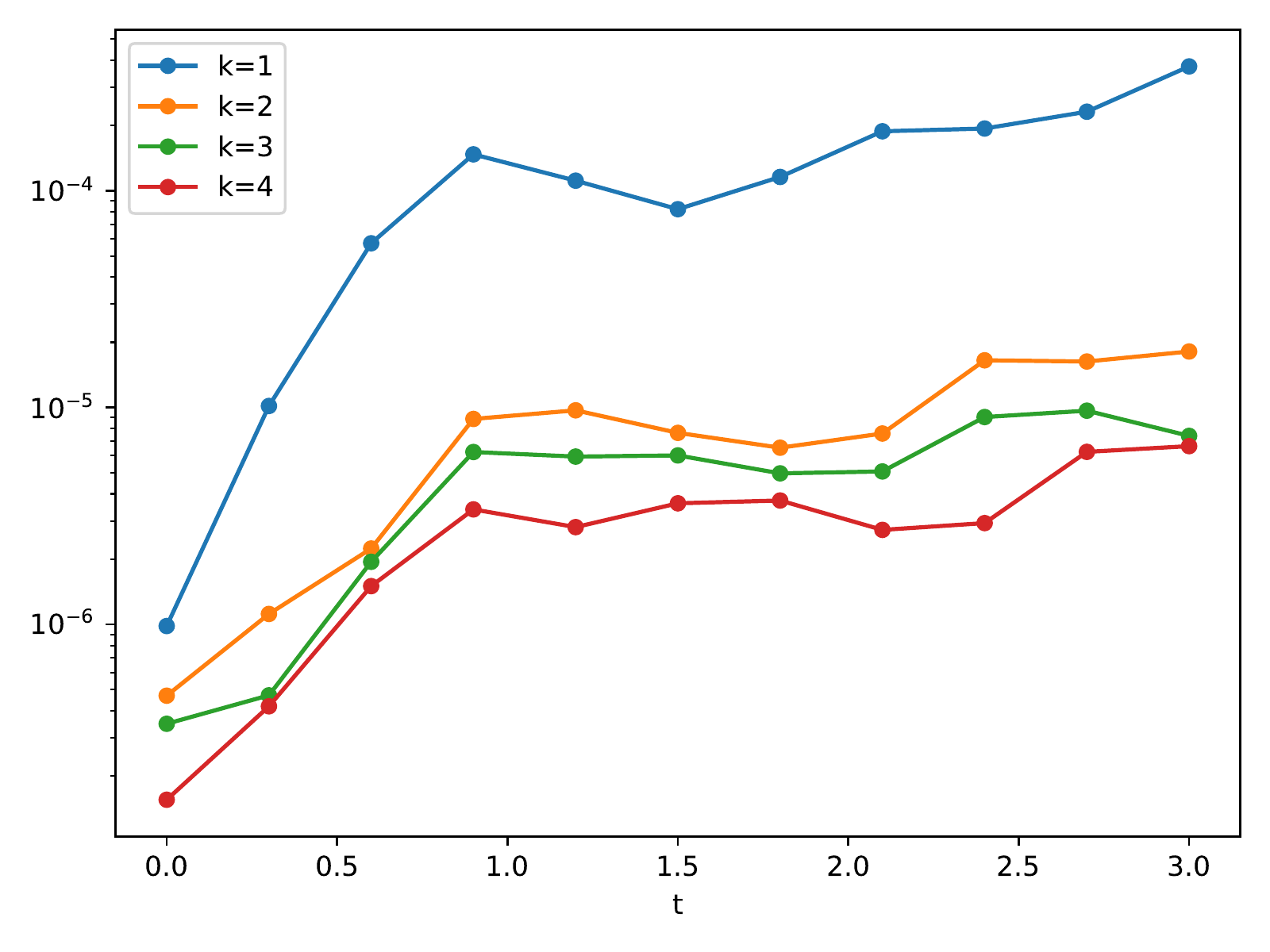}
		%		\subcaption{DGM-2-couple}
	\end{minipage}
	\caption{Linear oscillator. Relative $L^2$ error and relative KL divergence at different adaptive iterations $k$. Left panel: relative $L^2$ error. Right panel: relative KL divergence.}
	\label{linear_l2err}
\end{figure}

The comparison between the numerical solution and the ground truth is presented in Figure \ref{linear_result}. We again observe a good agreement between the predicted and the exact solutions. Moreover, Figure \ref{linear_l2err} shows the behaviors of the relative $L^2$ error for different adaptive iterations $N_{\mathrm{adaptive}}$, which suggests that a large number of iterations admit great help for improving the convergence.

\subsection{Nonlinear oscillator}
We now consider the TFP equation with a nonlinear drift term:
\begin{equation}
	\bm{\mu}= (x_2,\; x_1-0.4x_2 -0.1x_1^3), \quad \bm{D}=\mathrm{diag}(0,\;0.4).
\end{equation}
The initial distribution is given by
\begin{equation}
	p(\bm{x},0) = \mathcal{N}\big((0,5),\; \bm{I}_2\big).
\end{equation}

\begin{figure}[!h]
	\centering
		\begin{minipage}[t]{0.3\linewidth}
		\includegraphics[scale=0.3]{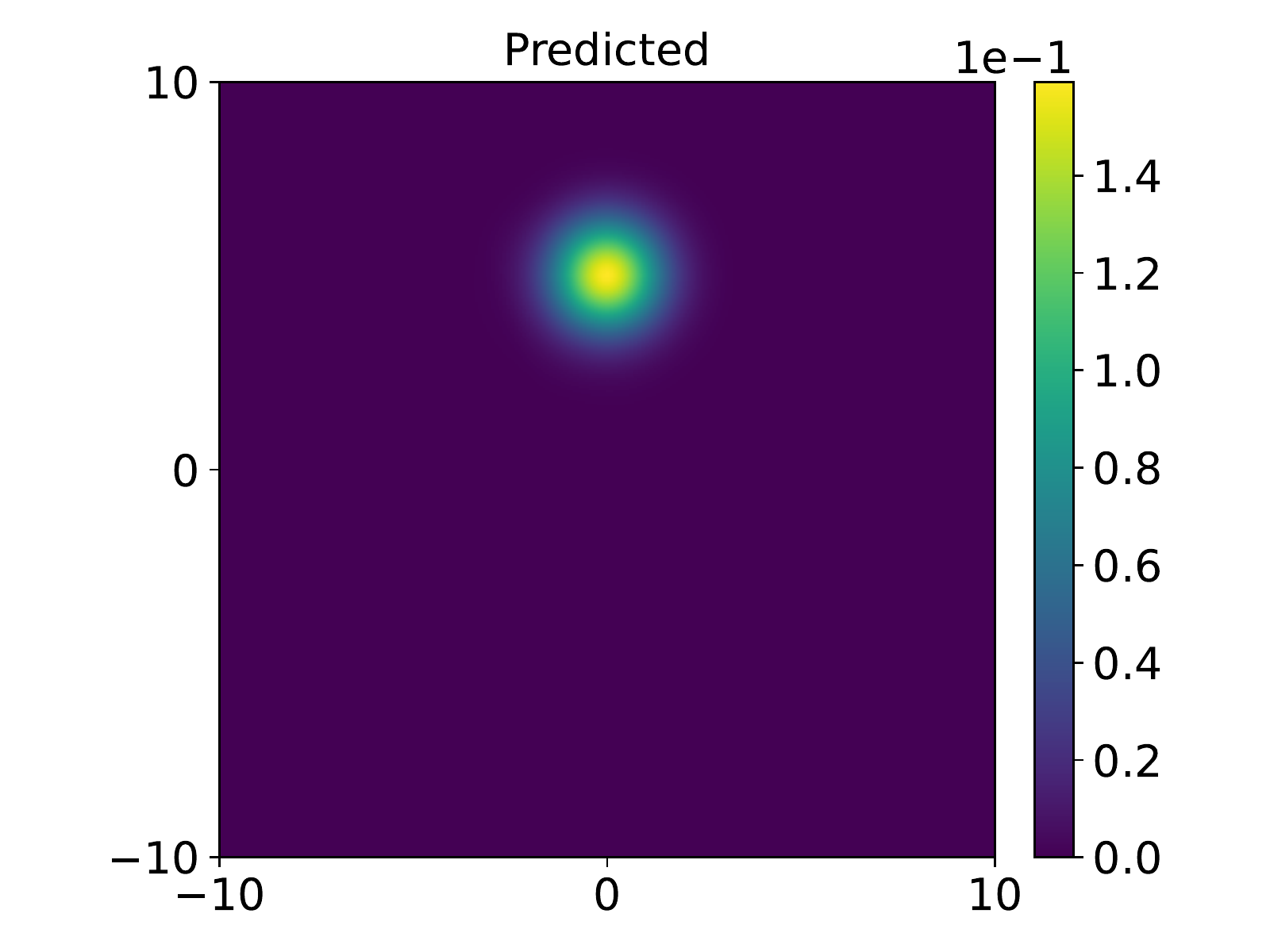}
		%		\subcaption{DGM-2}
	\end{minipage}
	\begin{minipage}[t]{0.3\linewidth}
		\includegraphics[scale=0.3]{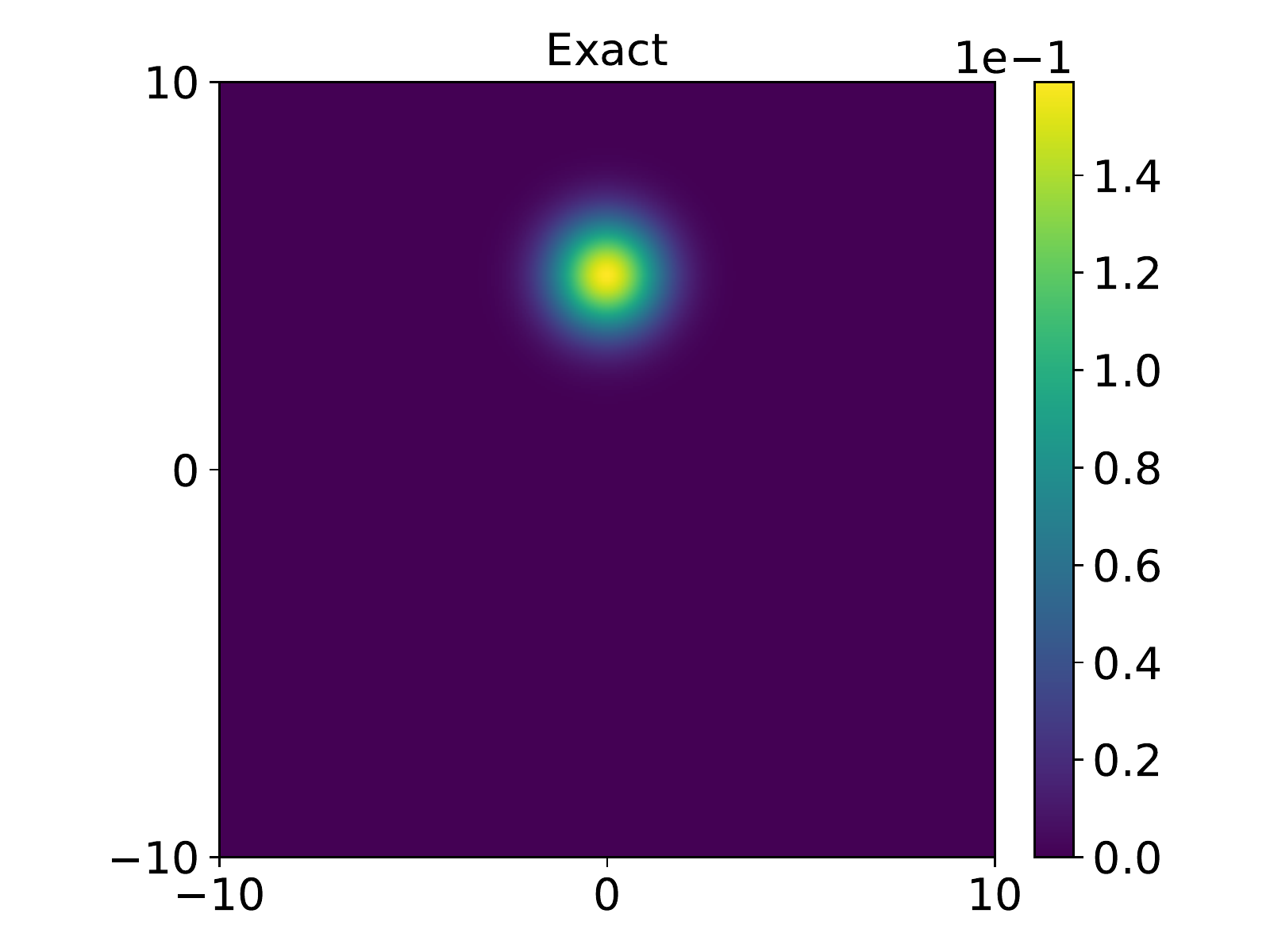}
		%		\subcaption{DGM-2-uncouple}
	\end{minipage}
	\begin{minipage}[t]{0.3\linewidth}
		\includegraphics[scale=0.3]{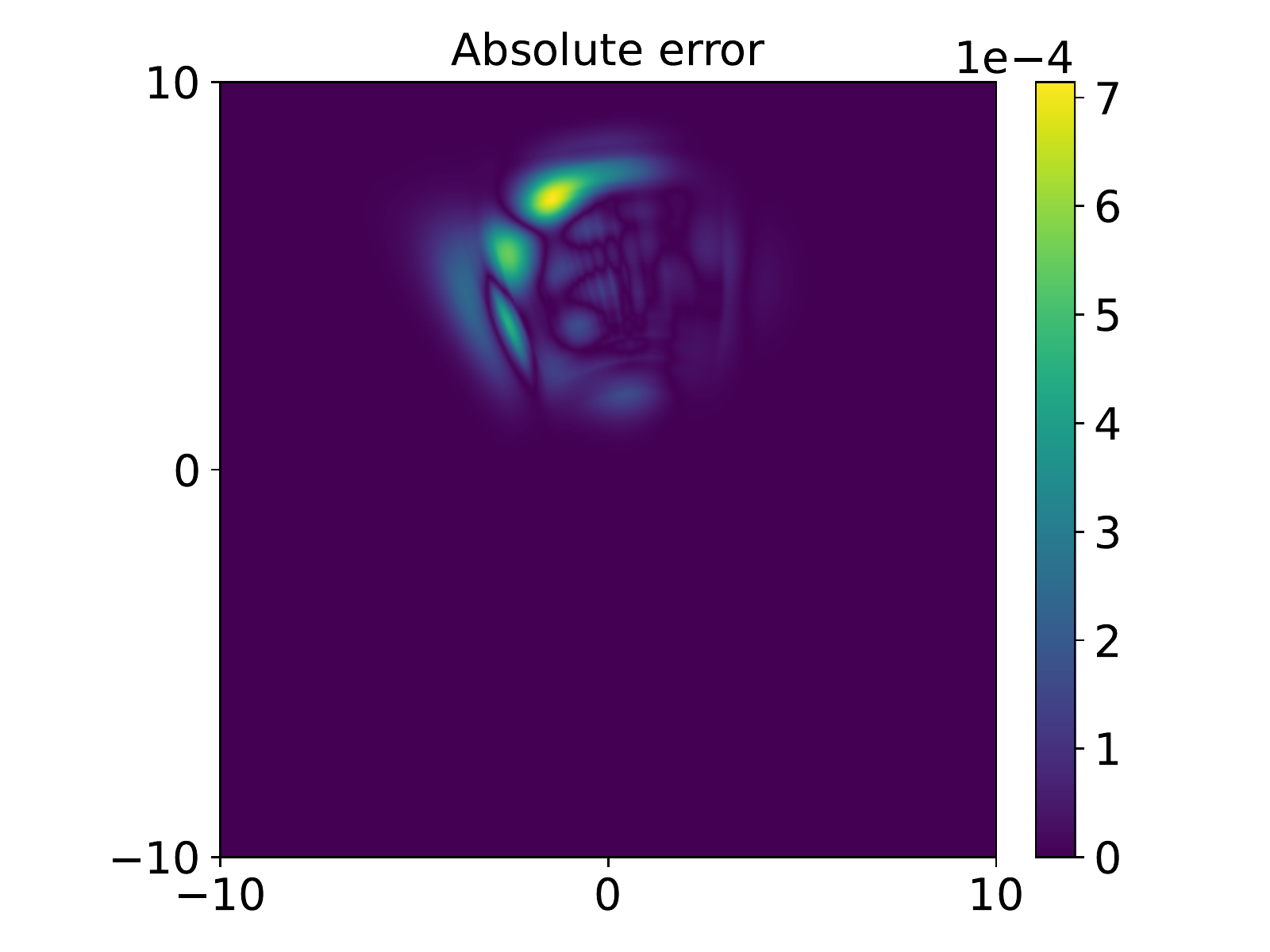}
		%		\subcaption{DGM-2-couple}
	\end{minipage}
%	\subcaption{$t=0.$}
	
	\begin{minipage}[t]{0.3\linewidth}
		\includegraphics[scale=0.3]{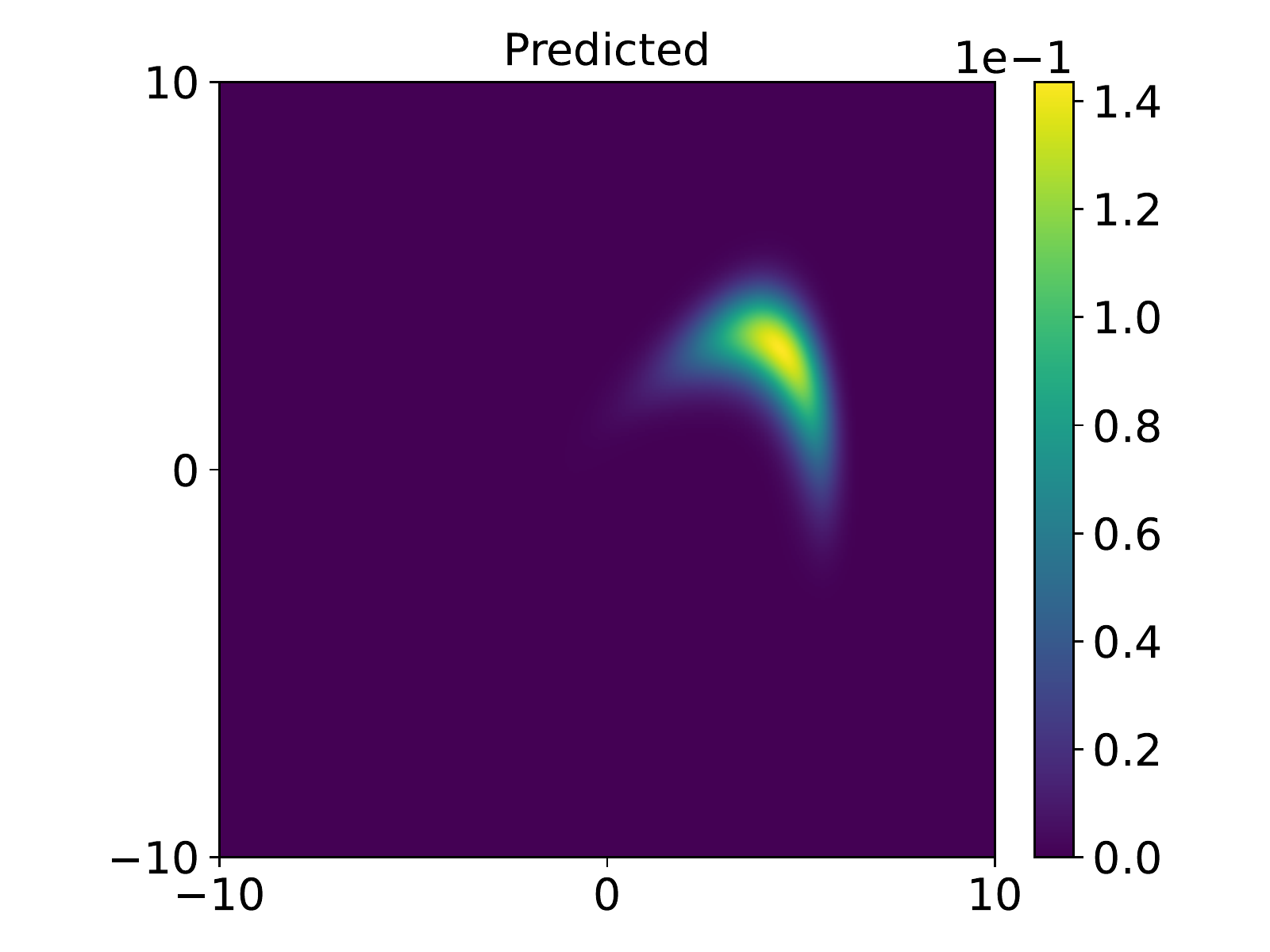}
		%		\subcaption{DGM-2}
	\end{minipage}
	\begin{minipage}[t]{0.3\linewidth}
		\includegraphics[scale=0.3]{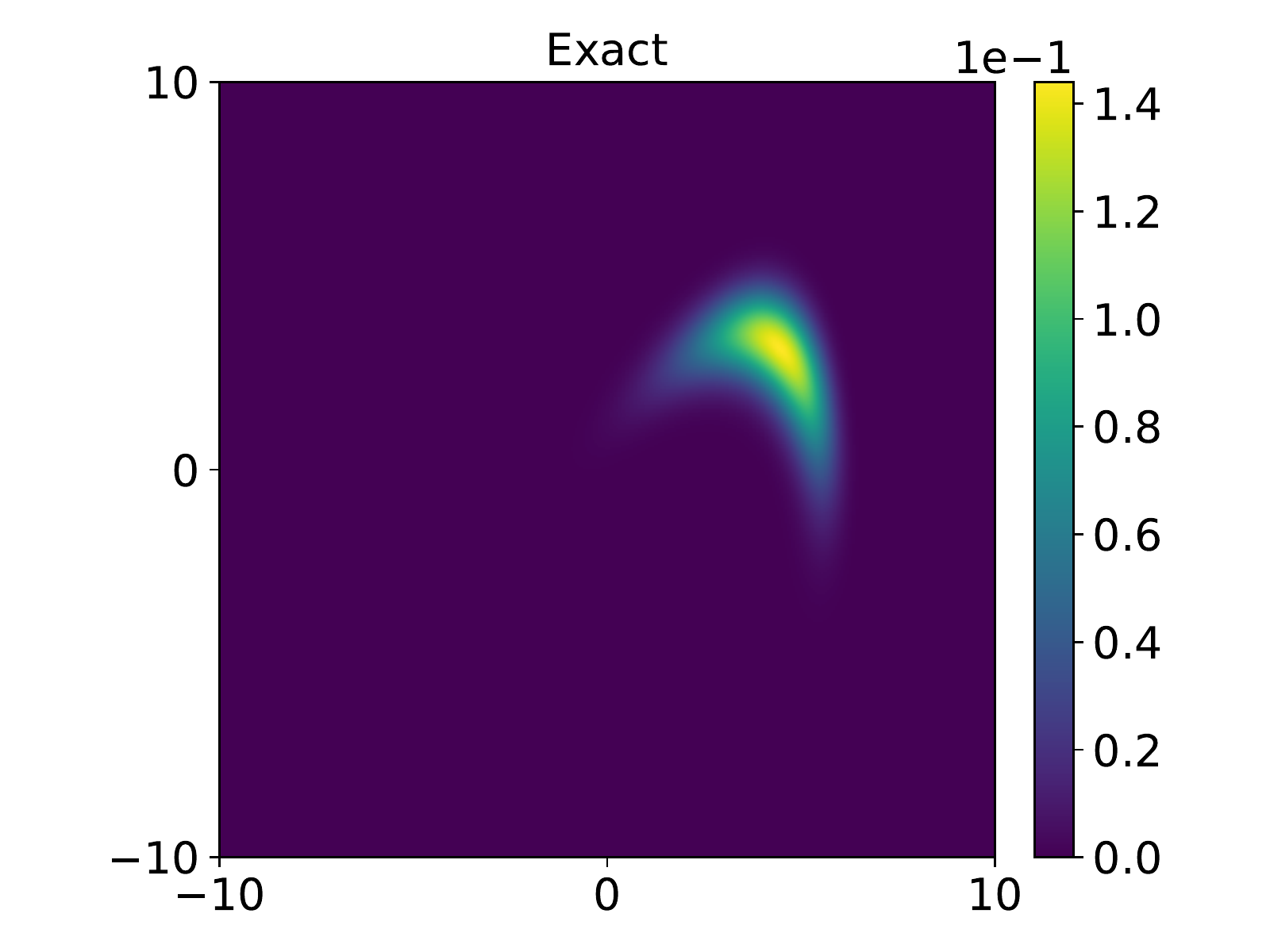}
		%		\subcaption{DGM-2-uncouple}
	\end{minipage}
	\begin{minipage}[t]{0.3\linewidth}
		\includegraphics[scale=0.3]{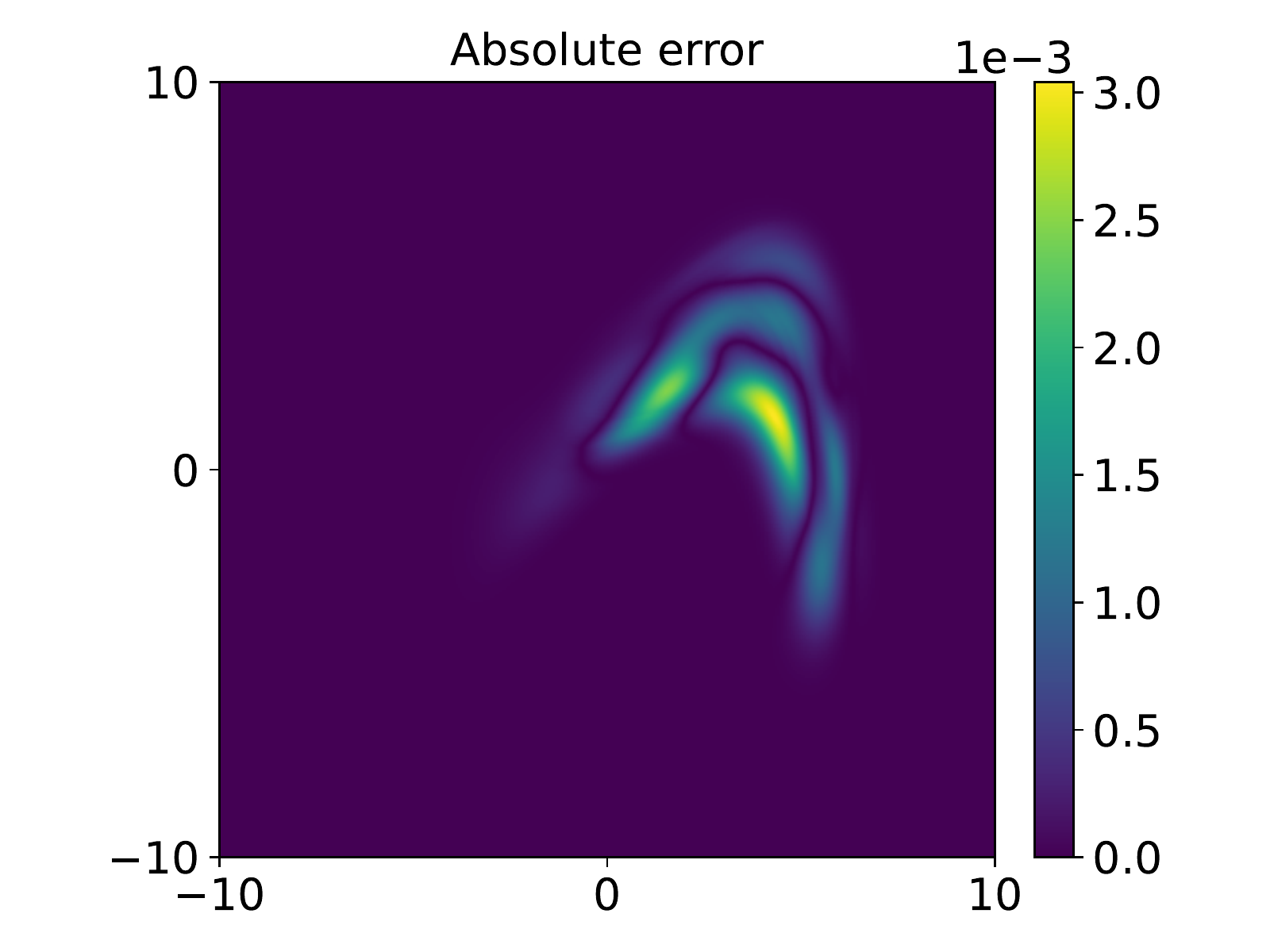}
		%		\subcaption{DGM-2-couple}
	\end{minipage}
%		\subcaption{$t=1.$}
		
	\begin{minipage}[t]{0.3\linewidth}
	\includegraphics[scale=0.3]{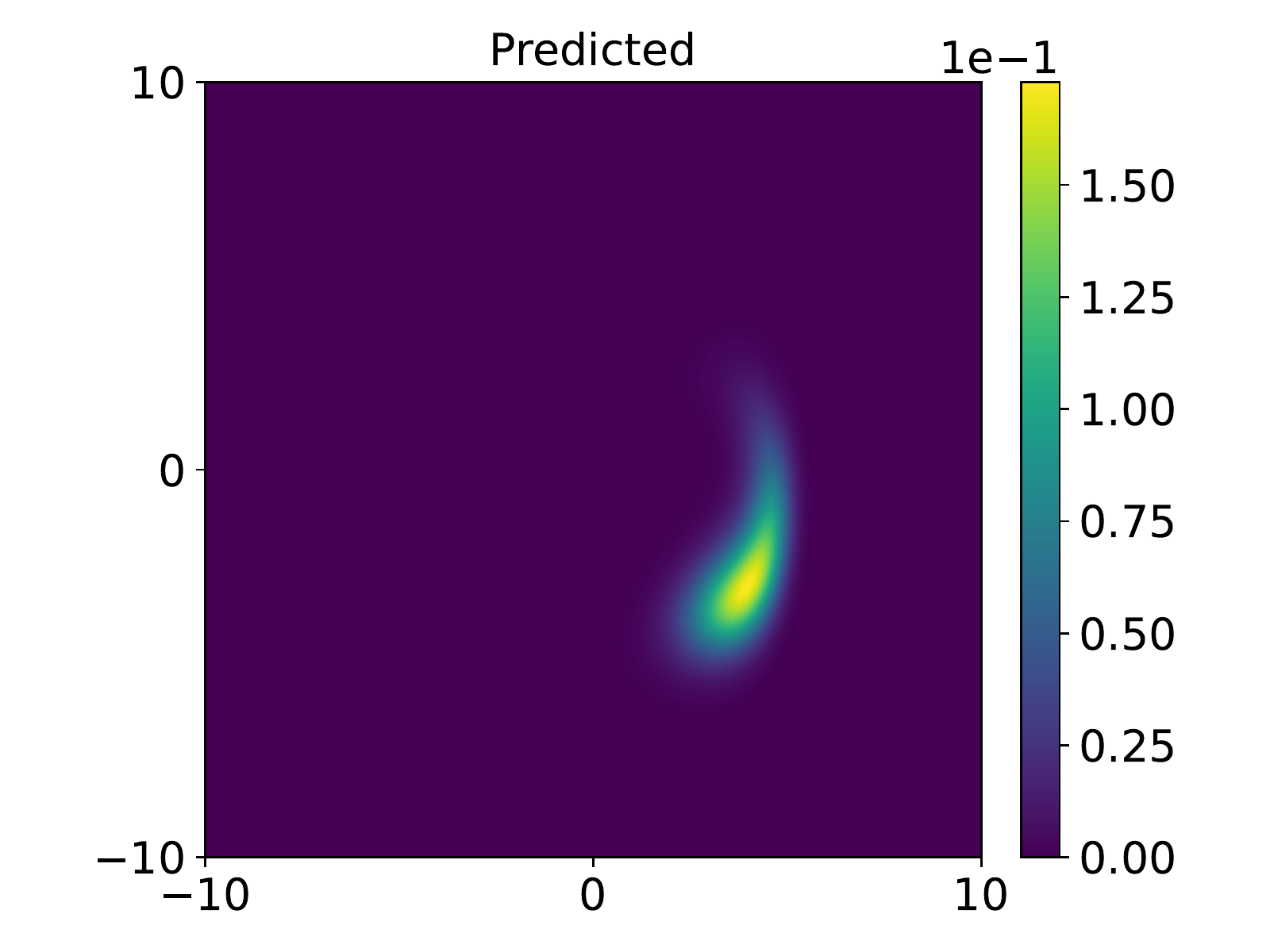}
	%		\subcaption{DGM-2}
\end{minipage}
\begin{minipage}[t]{0.3\linewidth}
	\includegraphics[scale=0.3]{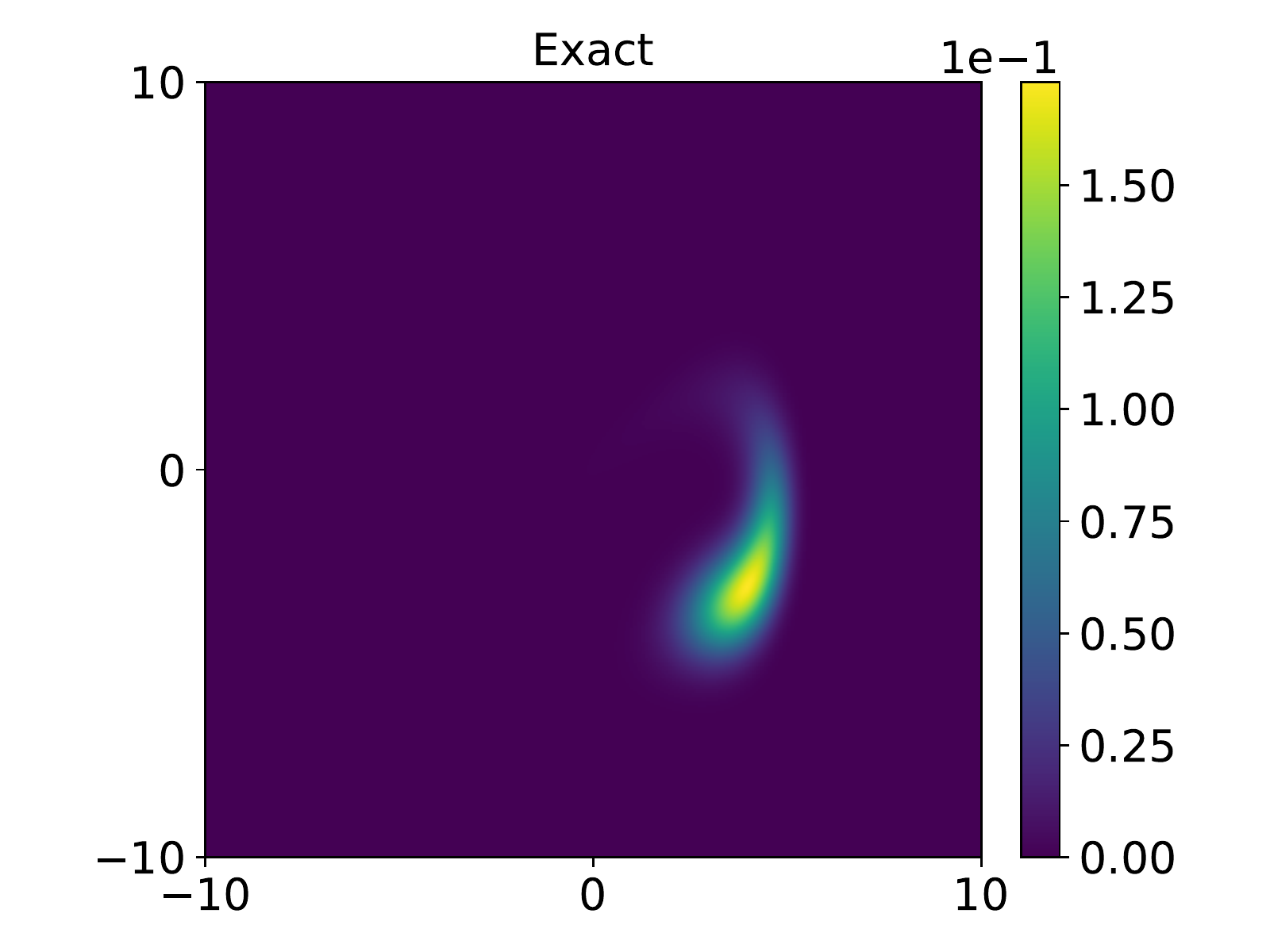}
	%		\subcaption{DGM-2-uncouple}
\end{minipage}
\begin{minipage}[t]{0.3\linewidth}
	\includegraphics[scale=0.3]{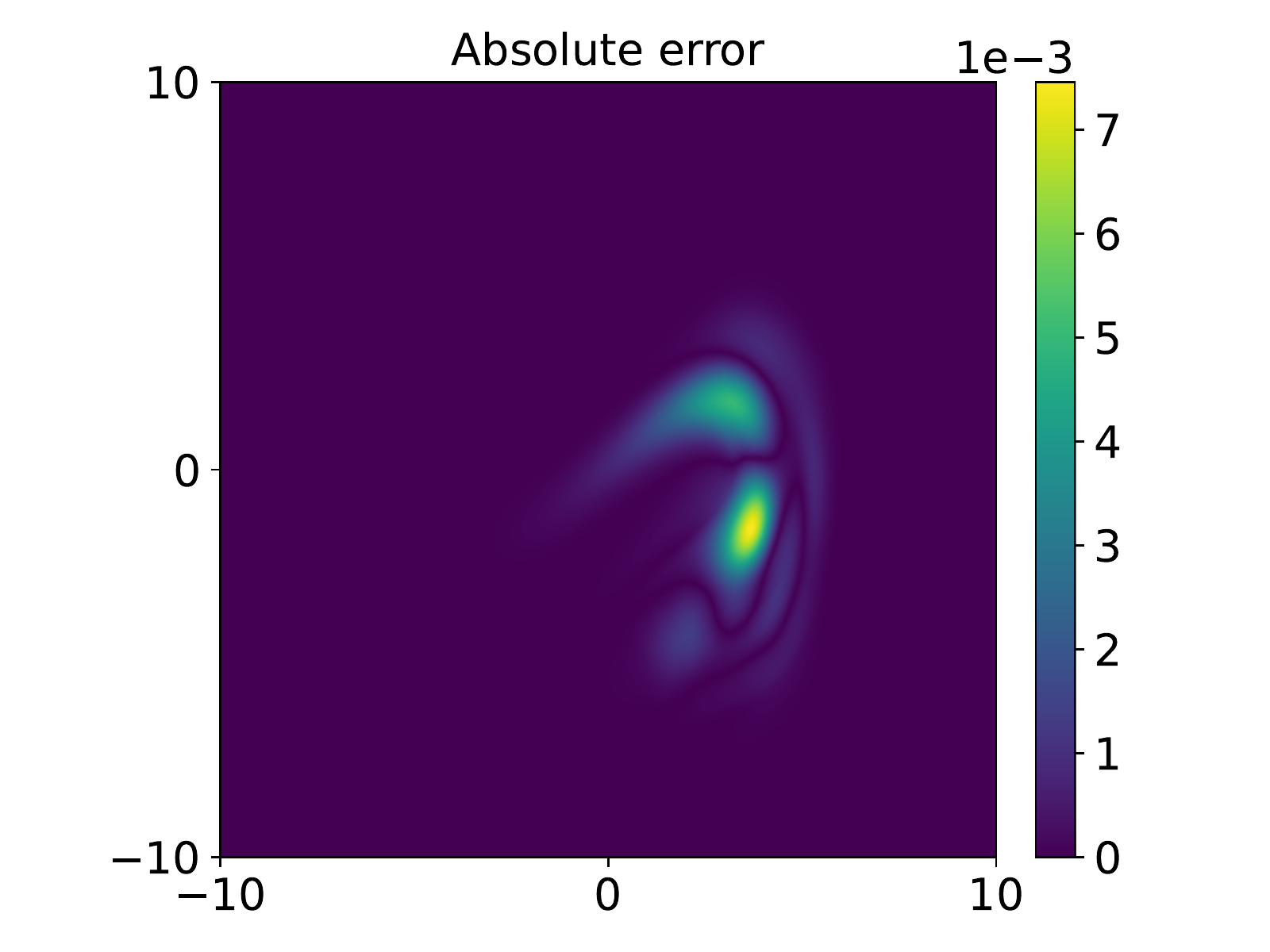}
	%		\subcaption{DGM-2-couple}
\end{minipage}
%		\subcaption{$t=2.$}

	\begin{minipage}[t]{0.3\linewidth}
	\includegraphics[scale=0.3]{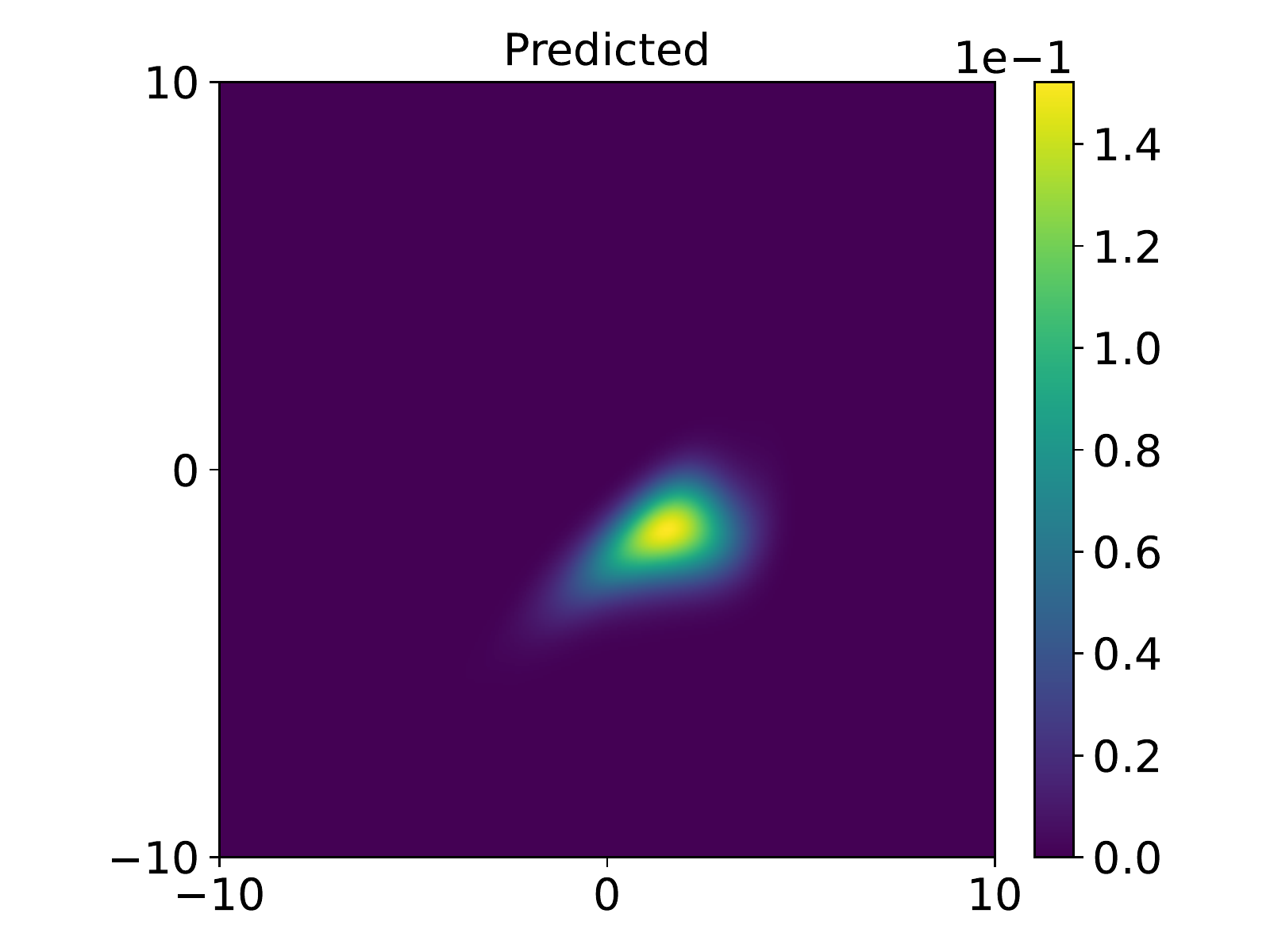}
	%		\subcaption{DGM-2}
\end{minipage}
\begin{minipage}[t]{0.3\linewidth}
	\includegraphics[scale=0.3]{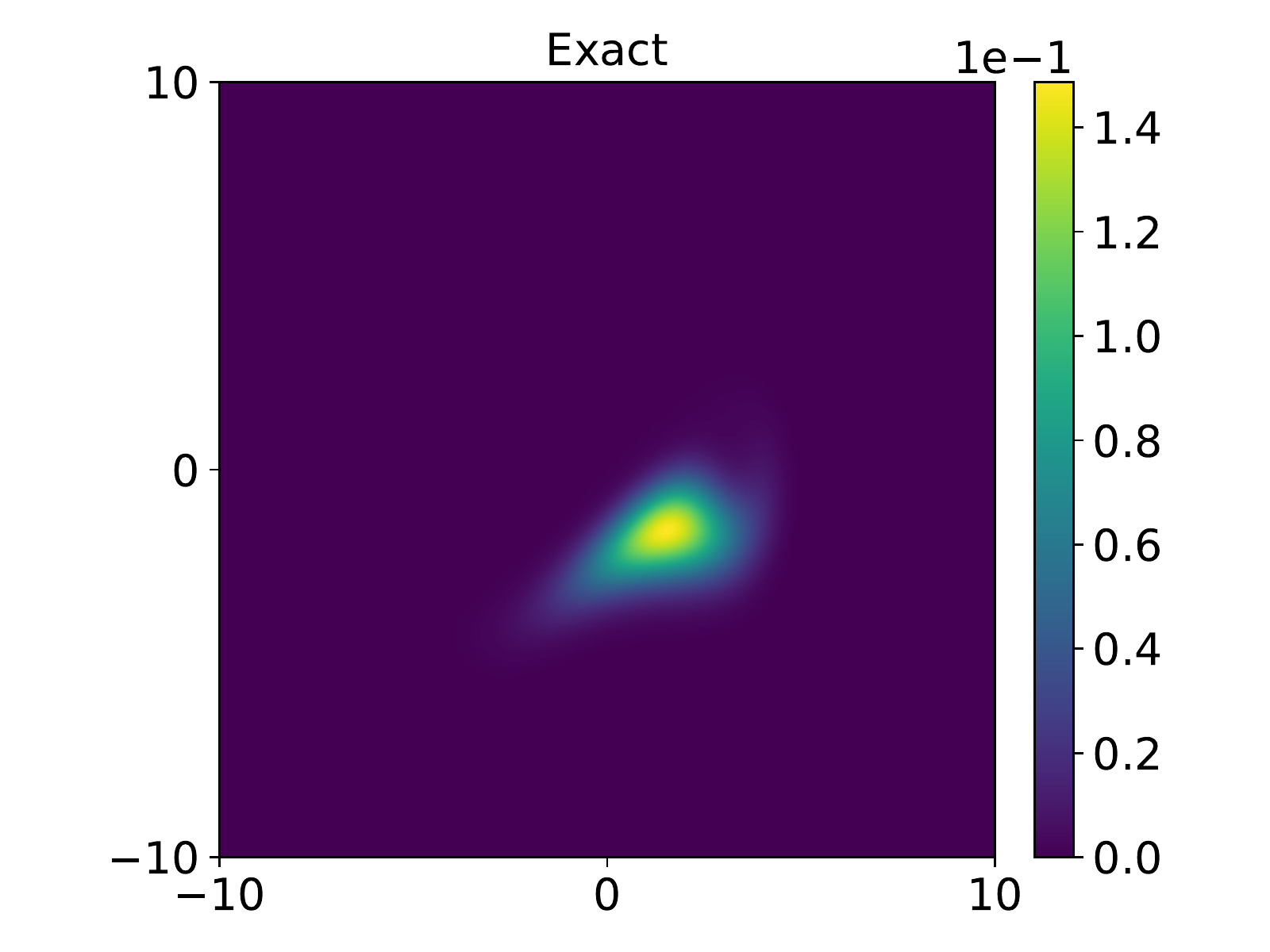}
	%		\subcaption{DGM-2-uncouple}
\end{minipage}
\begin{minipage}[t]{0.3\linewidth}
	\includegraphics[scale=0.3]{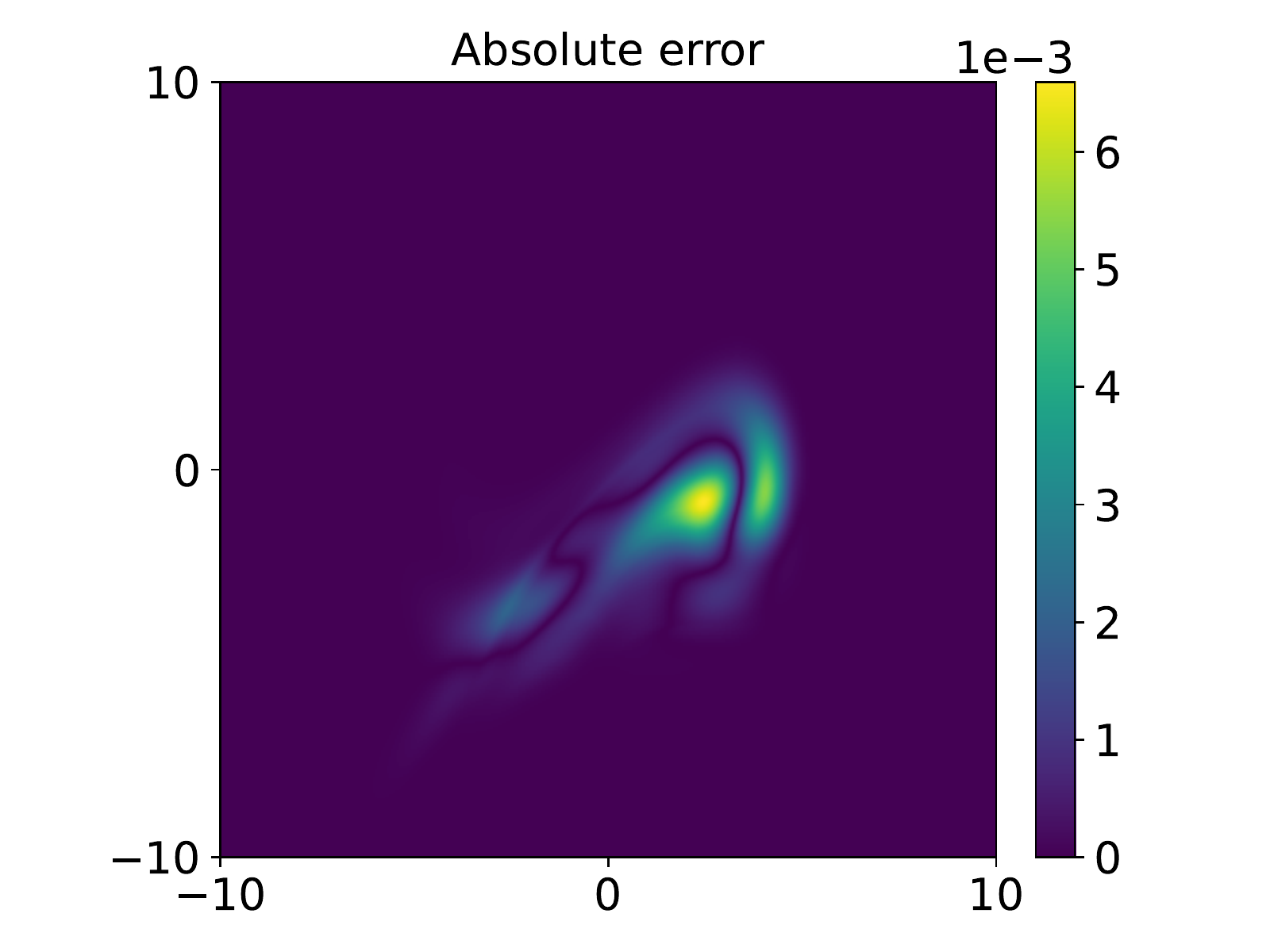}
	%		\subcaption{DGM-2-couple}
\end{minipage}
%	\subcaption{$t=3.$}
	
	\caption{Nonlinear oscillator. Predicted solution versus the reference solution for different time $t$. Top row: $t=0$. Second row: $t=1$. Third row: $t=2$. Bottom row: $t=3.$}
	\label{nonlinear_results}
\end{figure}

We solve this problem in time interval $[0,3],$ and again we compute the reference solution by the ADI scheme in a truncated spatial domain $[-10,10]^2$, with mesh size $\delta h = 0.01$ and $\delta t=0.005$. For our learning scheme, the parameters are chosen as $N_e=50,$ $\alpha=1.5$. We use five adaptivity iterations, i.e.,  $N_{\mathrm{adaptive}} = 5$, and set batch size to 10000. We use $L=4$ affine coupling layers and actnorm layers, and turn on the polynomial spline layer with $50$ nodes. The initial spatial training set is generated through a uniform distribution in a range $[-10,10]^2$ with sample size 5000, and the corresponding initial temporal training set is chosen from a nonuniform partition(with 100 equidistant nodes in the $[0,1.5]$ and 200 equidistant nodes in the $[1.5,3]$), which results in total $5000\times 300= 1,500,000$ training points. As shown in Figure \ref{nonlinear_results}, a good agreement can be achieved between the predictions and the exact solutions.

\begin{figure}
	\centering
	\begin{minipage}[t]{0.4\linewidth}
		\includegraphics[scale=0.4]{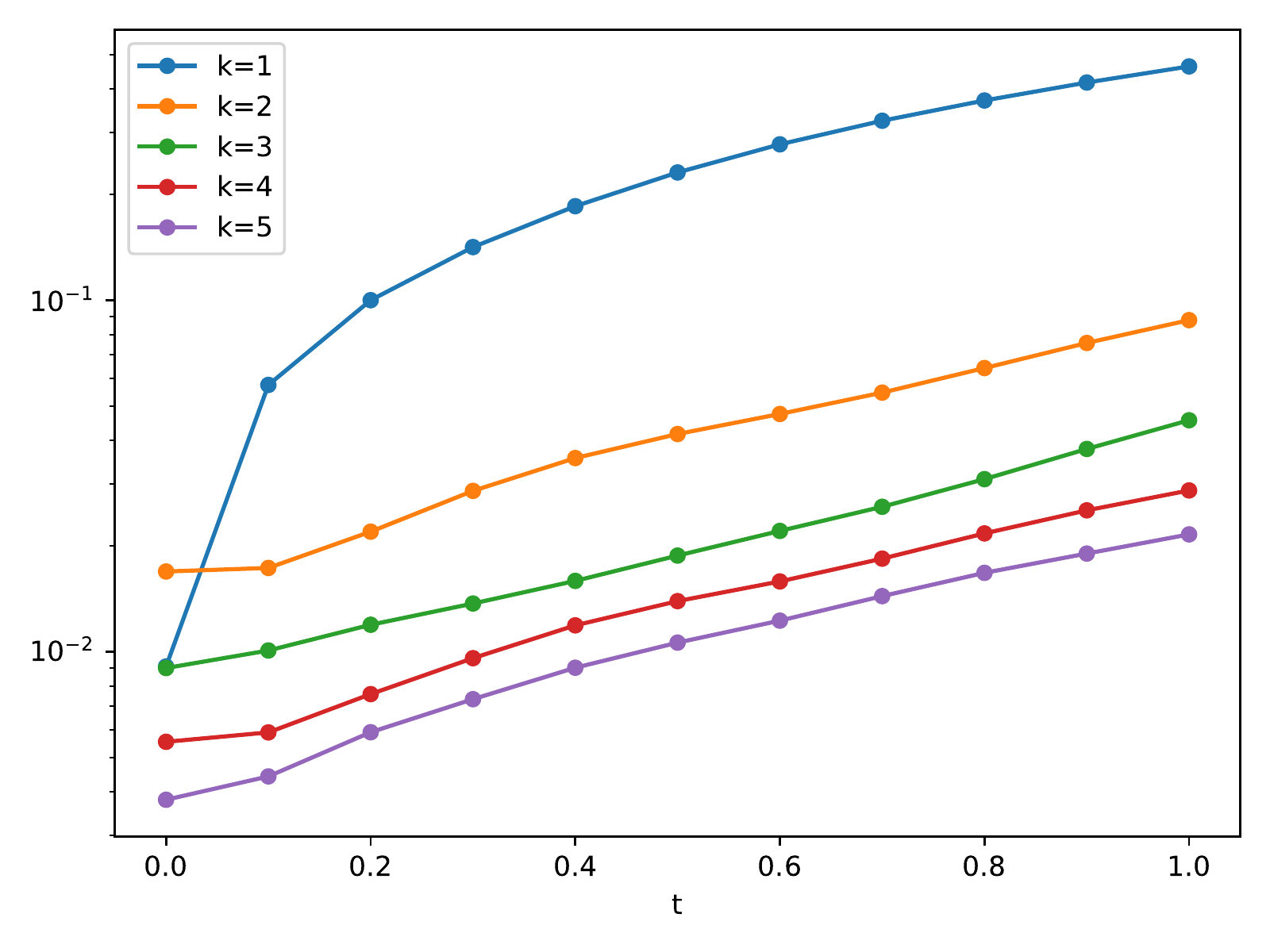}
		%		\subcaption{DGM-2-uncouple}
	\end{minipage}
	\begin{minipage}[t]{0.4\linewidth}
		\includegraphics[scale=0.4]{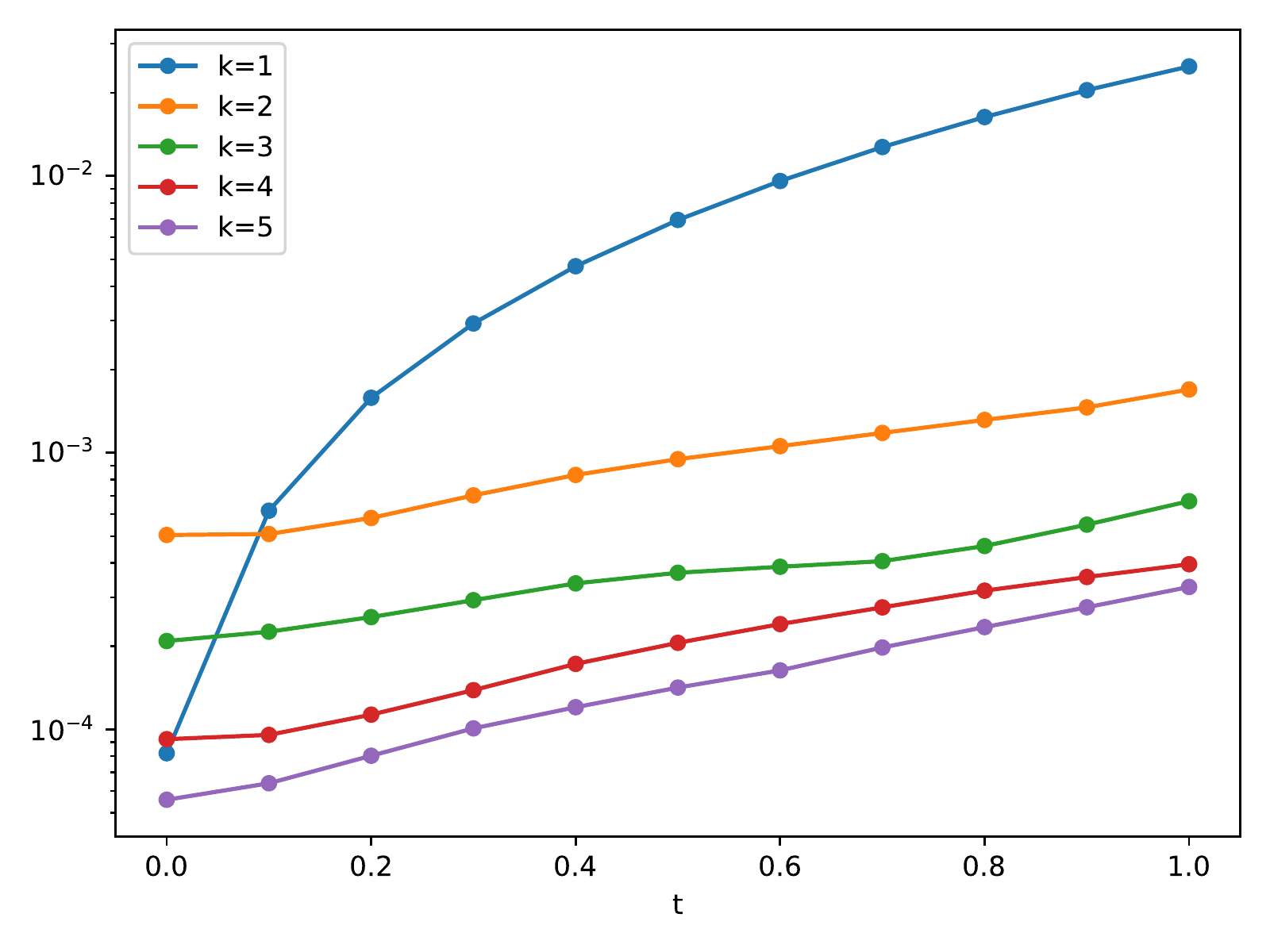}
		%		\subcaption{DGM-2-couple}
	\end{minipage}
	\caption{Nonlinear oscillator. Relative $L^2$ error and relative KL divergence at different adaptive iterations $k$. Left panel: relative $L^2$ error. Right panel: relative KL divergence.}
	\label{nonlinear_l2err}
\end{figure}

We also present the relative $L^2$ errors and relative KL divergences with different adaptive iterations in Figure \ref{nonlinear_l2err}. Again, we can see that more  iterations provide great help for improving the convergence. However, similar  to the last example, we also observe that the numerical error seems to increase as time evolves. Nevertheless, we shall show in the next examples that this can be improved by increasing the training points and iterations as time evolves.

\subsection{High dimensional TFP equations}
\begin{figure}[!h]
	\centering
	\begin{minipage}[t]{0.3\linewidth}
		\includegraphics[scale=0.3]{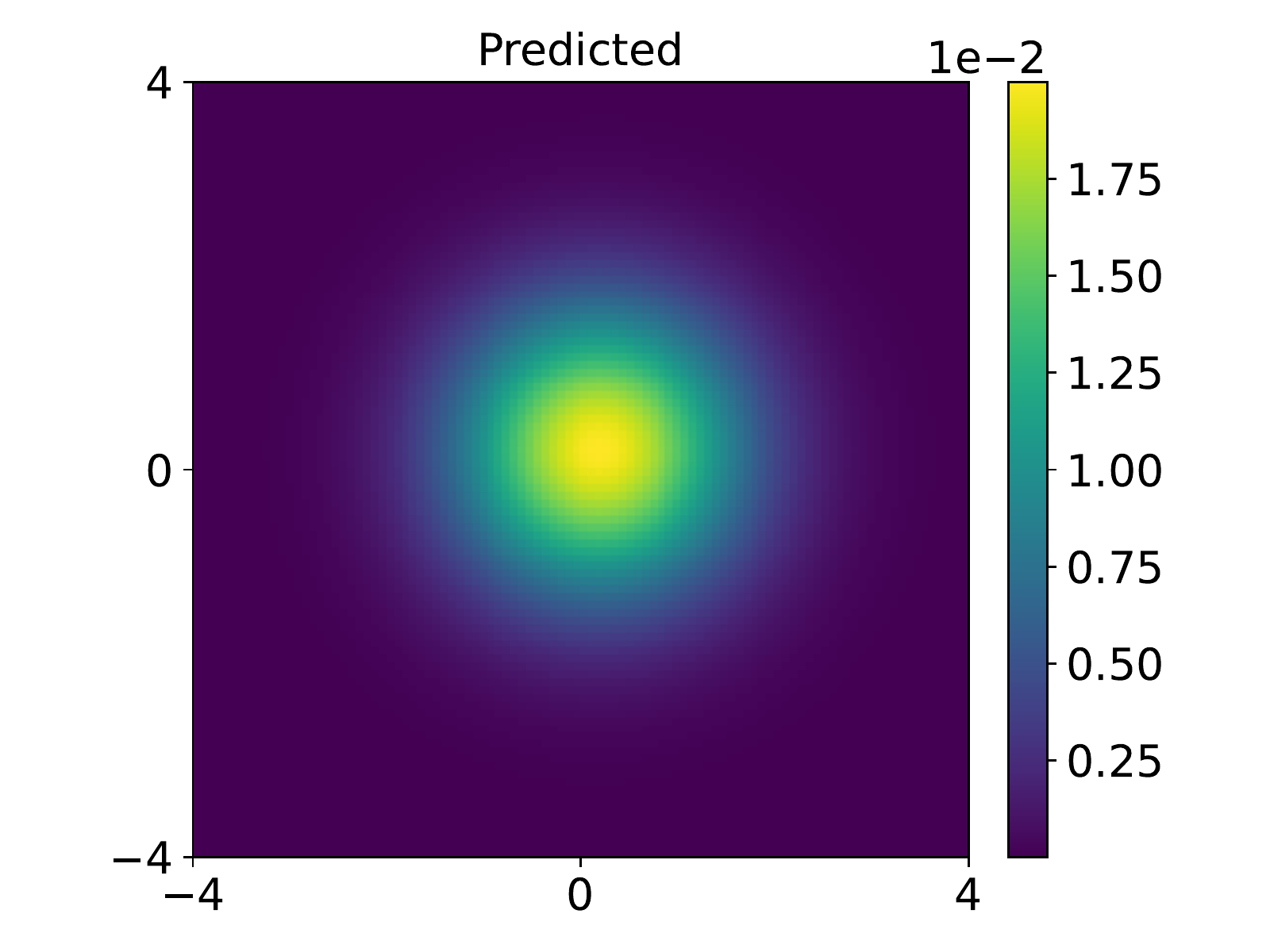}
		%		\subcaption{DGM-2}
	\end{minipage}
	\begin{minipage}[t]{0.3\linewidth}
		\includegraphics[scale=0.3]{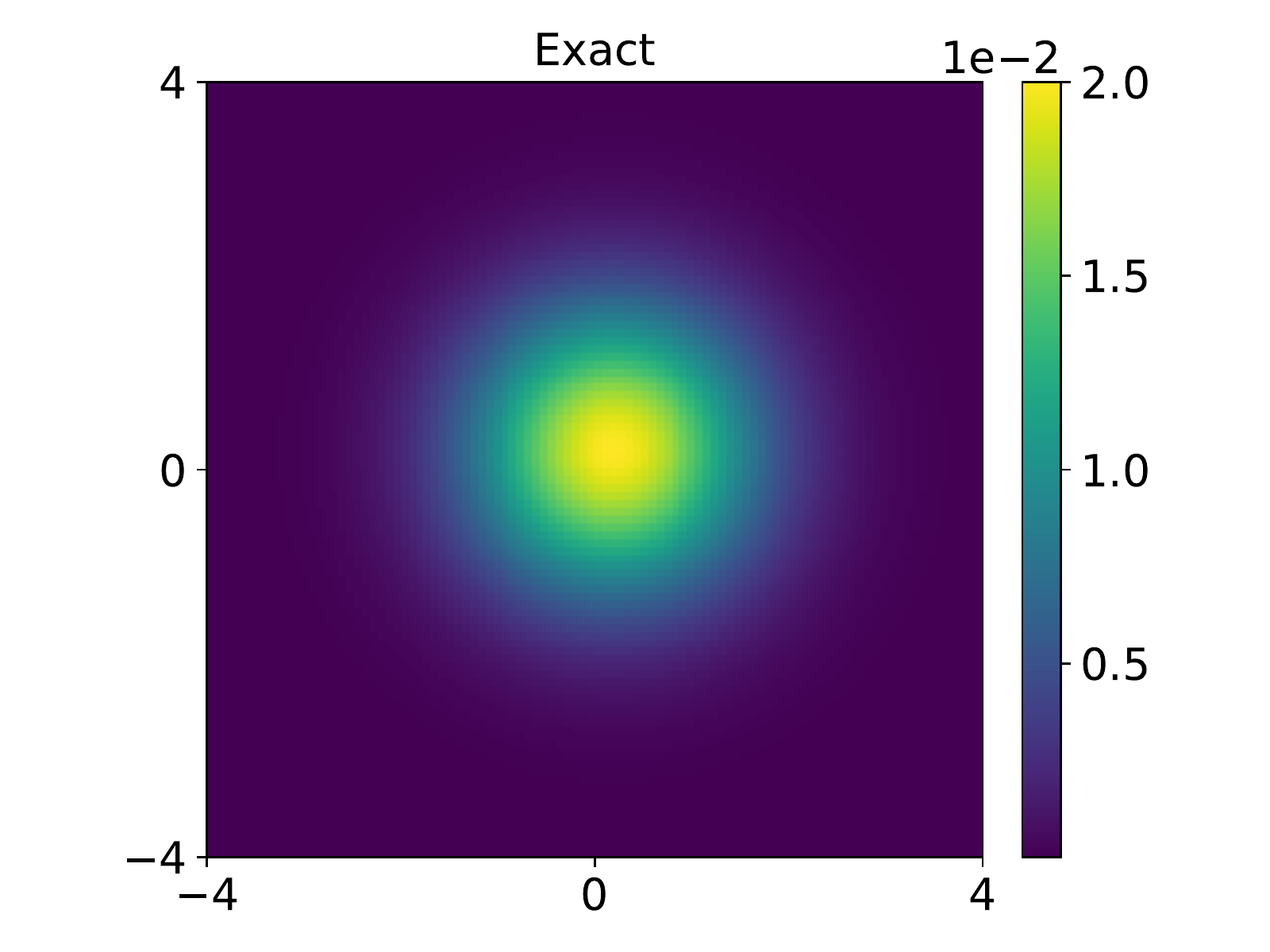}
		%		\subcaption{DGM-2-uncouple}
	\end{minipage}
	\begin{minipage}[t]{0.3\linewidth}
		\includegraphics[scale=0.3]{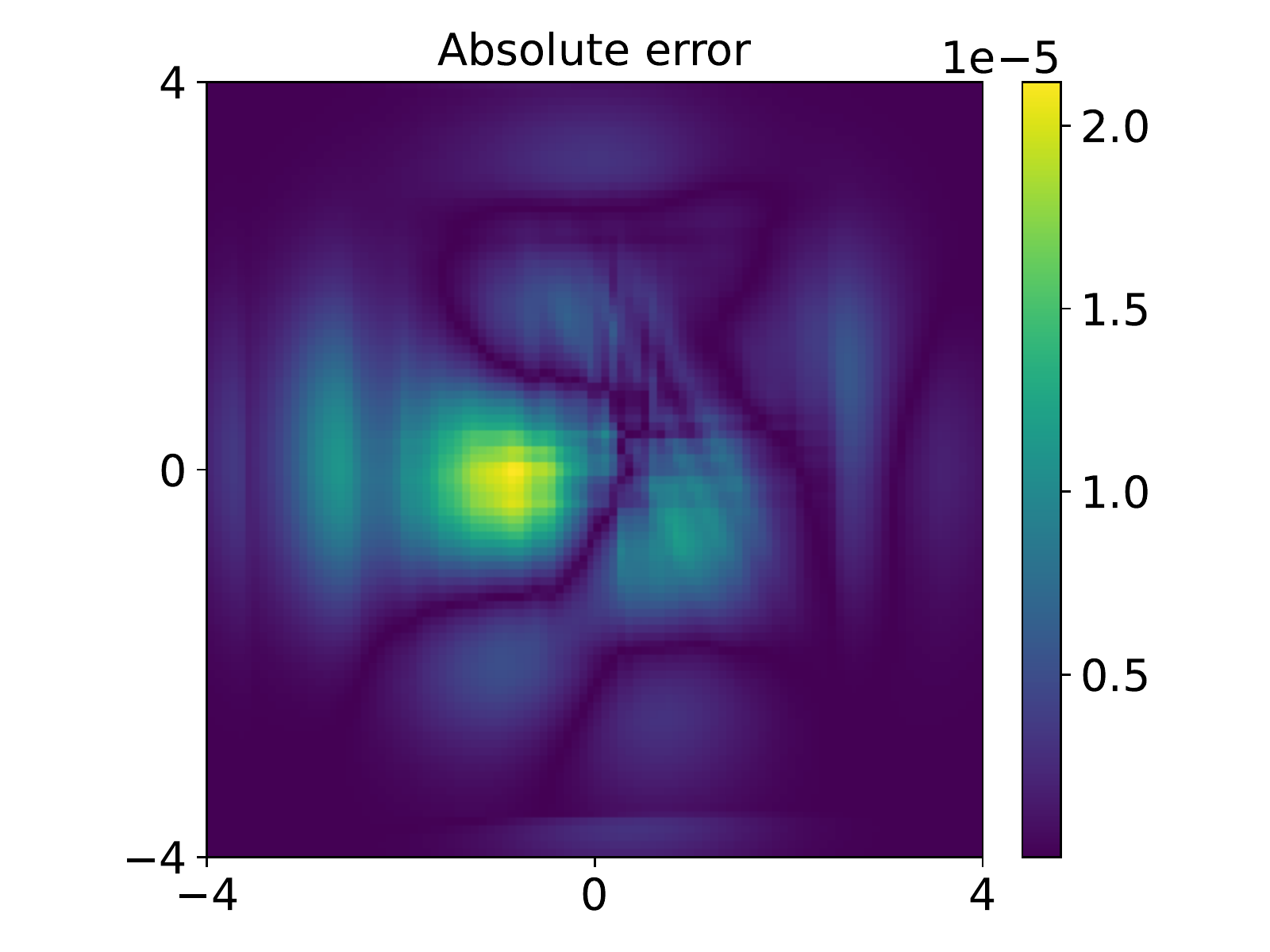}
		%		\subcaption{DGM-2-couple}
	\end{minipage}

	\begin{minipage}[t]{0.3\linewidth}
	\includegraphics[scale=0.3]{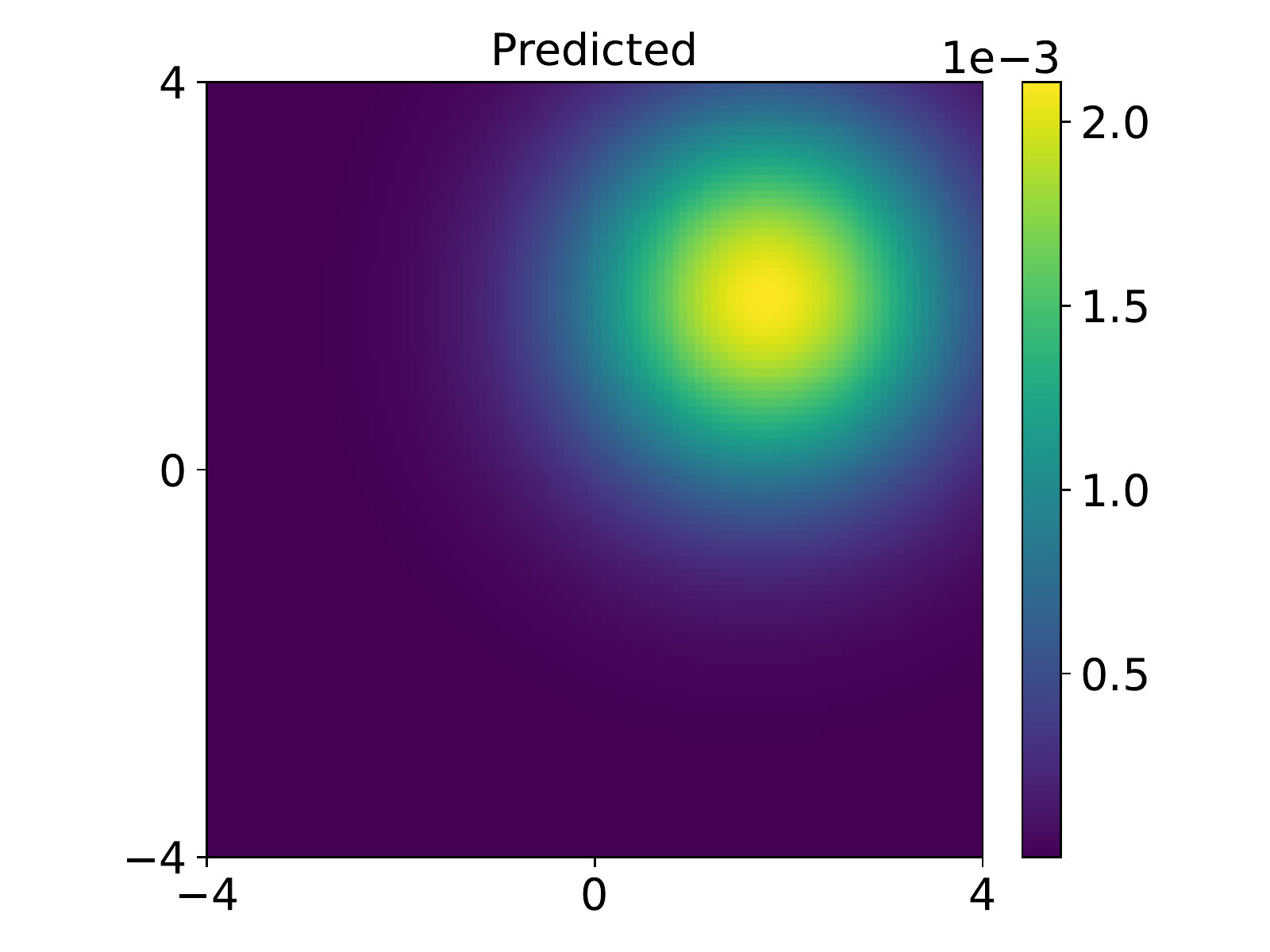}
	%		\subcaption{DGM-2}
\end{minipage}
\begin{minipage}[t]{0.3\linewidth}
	\includegraphics[scale=0.3]{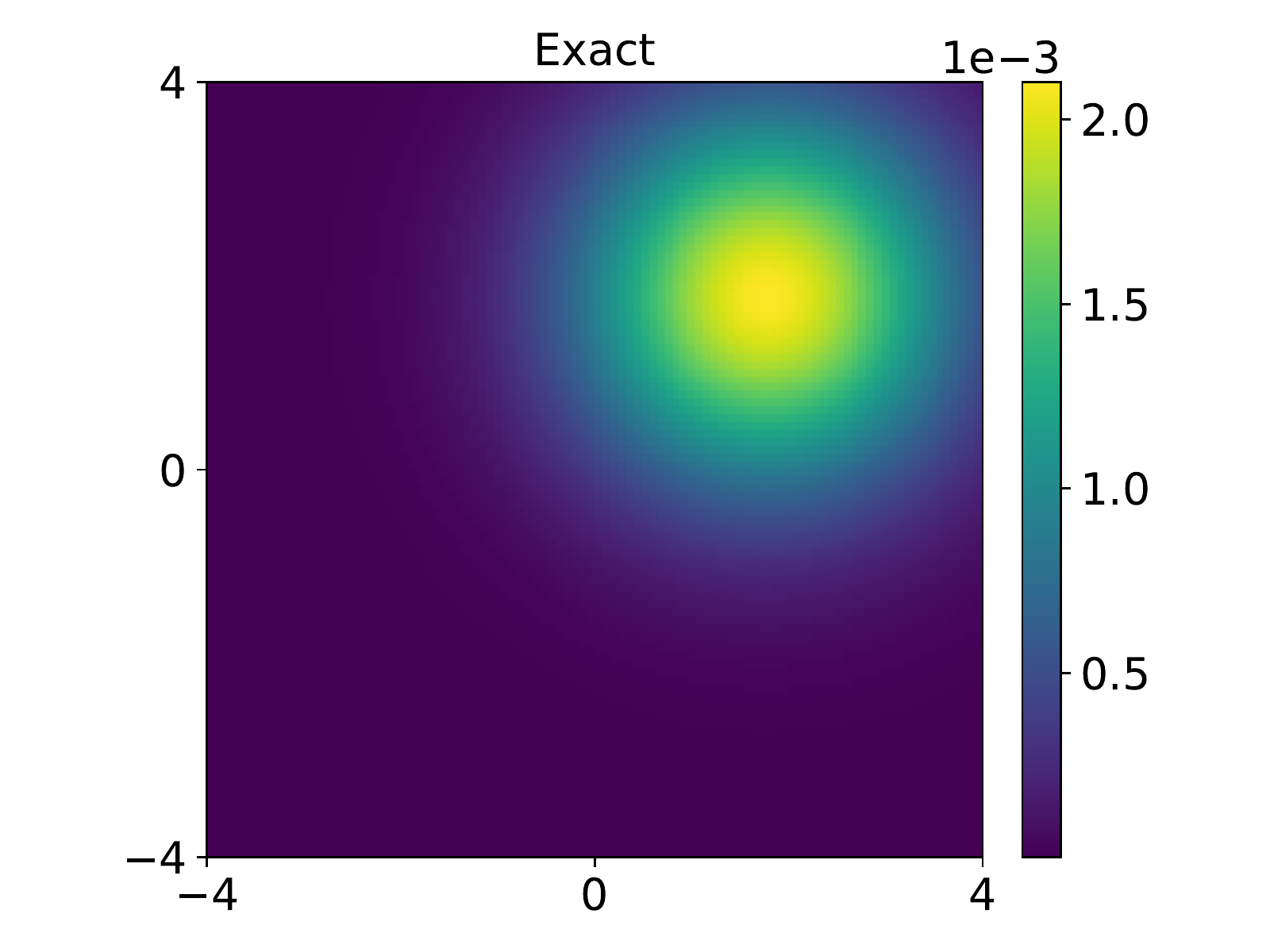}
	%		\subcaption{DGM-2-uncouple}
\end{minipage}
\begin{minipage}[t]{0.3\linewidth}
	\includegraphics[scale=0.3]{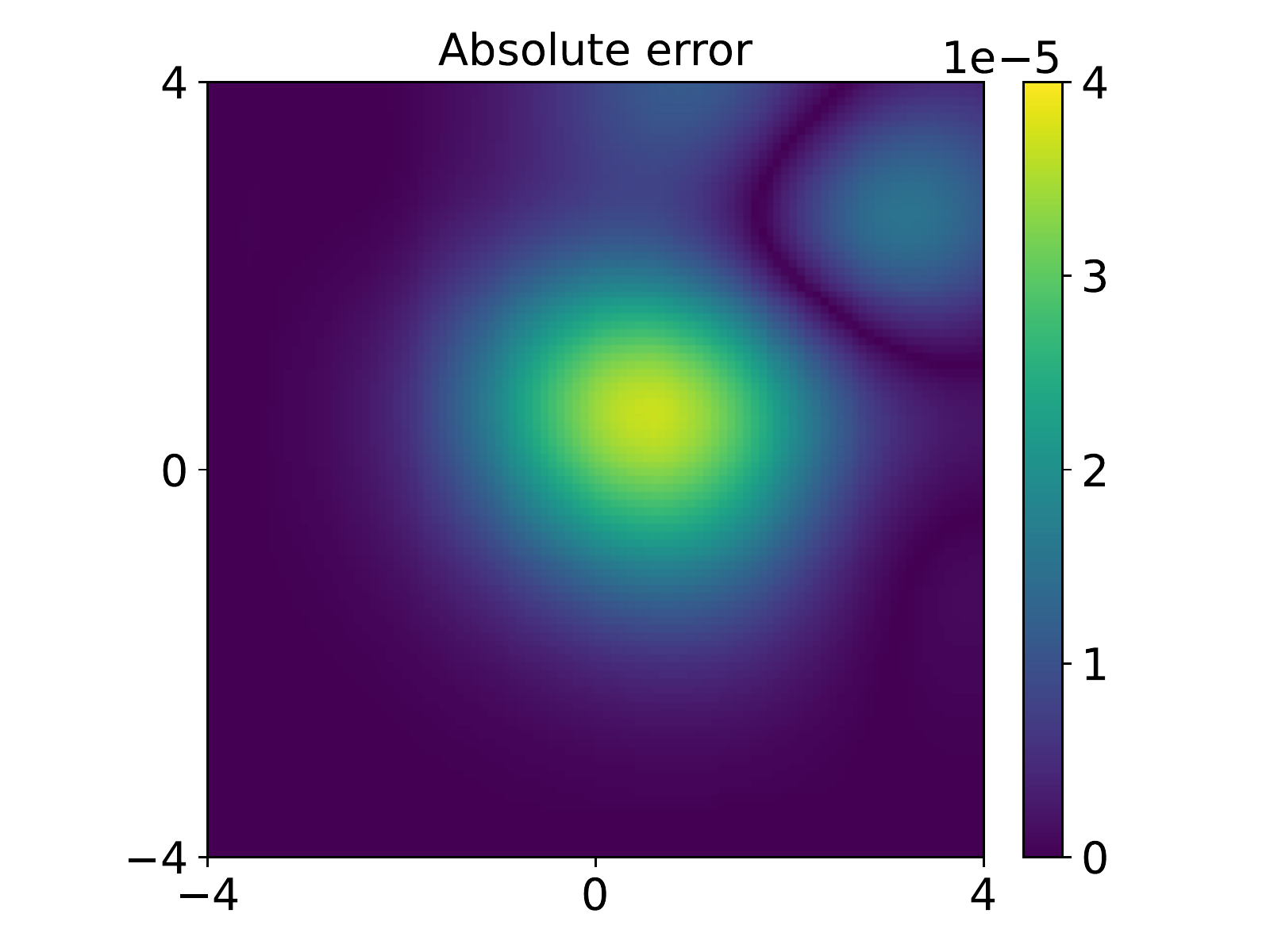}
	%		\subcaption{DGM-2-couple}
\end{minipage}
	\caption{4D example. Predicted solution versus the reference solution for different time $t$. Top row: $x_1=0.1, x_4=0.5,t=0.1$. Bottom row: $x_1=0.1, x_4=0.5,t=0.9.$}
	\label{4d_example_result}
\end{figure}

Finally, we consider a relatively high dimensional TFP equation:
\begin{equation}
\begin{aligned}
	&	\frac{\partial p}{\partial t} - \frac{1}{2}\Delta p + 2\nabla \cdot p=0,\\
	&p(\bm{x},0) = (2\pi)^{-d/2}\exp(-\|\bm{x}\|^2/2).
\end{aligned}
\end{equation}
Here $\bm{x}\in \mathbb{R}^d$ and $t\in [0,1]$. The corresponding exact solution is given by
\begin{equation}
	p(\bm{x},t)=\frac{1}{(2\pi (t+1))^{d/2}}\exp\bigg(-\frac{\|\bm{x}-2t\cdot\bm{1}_d\|^2}{2(t+1)}\bigg).
\end{equation}
We shall test this problem for the cases of $d=4$ and $d=8$. For $d=4,$ we set $N_e=100, \alpha=2, N_{\mathrm{adaptive}}=2$, batch size is 10000. We take $L=8$ affine coupling layers and actnorm layers, and we turn off the polynomial spline layer. The initial spatial training set is generated through a uniform distribution with a range $[-3,3]^4$, and the corresponding initial temporal training set is uniformly sampled in the unit interval $[0,1]$, which results in total $10000\times 50= 500,000$ training points. The comparison between the prediction and the ground truth is shown in Figure \ref{4d_example_result}. We can observe a good agreement between the predicted and exact solutions.

\begin{figure}[!h]
	\centering
		\begin{minipage}[t]{0.45\linewidth}
		\includegraphics[scale=0.4]{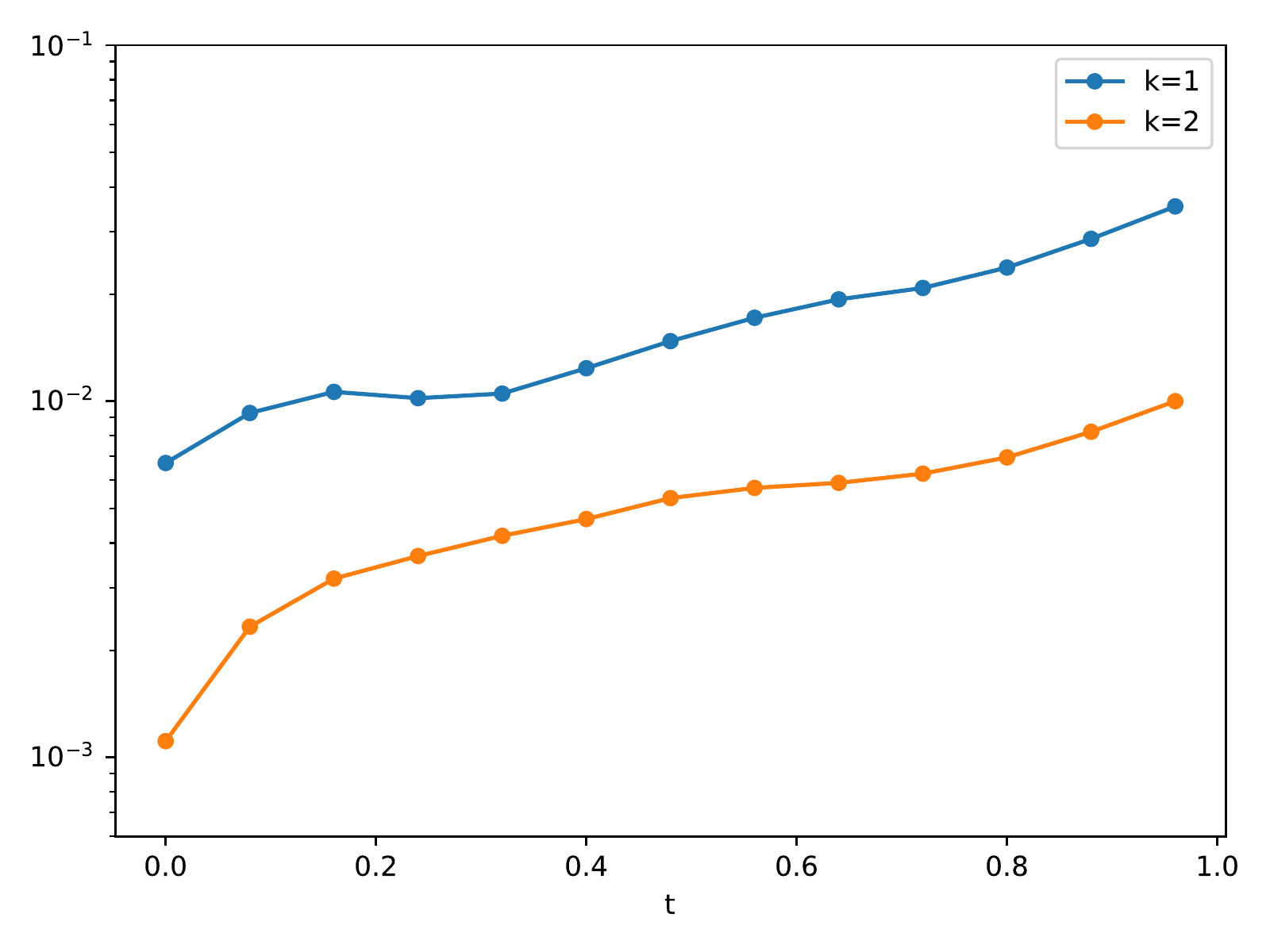}
		%		\subcaption{DGM-2}
	\end{minipage}
	\begin{minipage}[t]{0.45\linewidth}
		\includegraphics[scale=0.4]{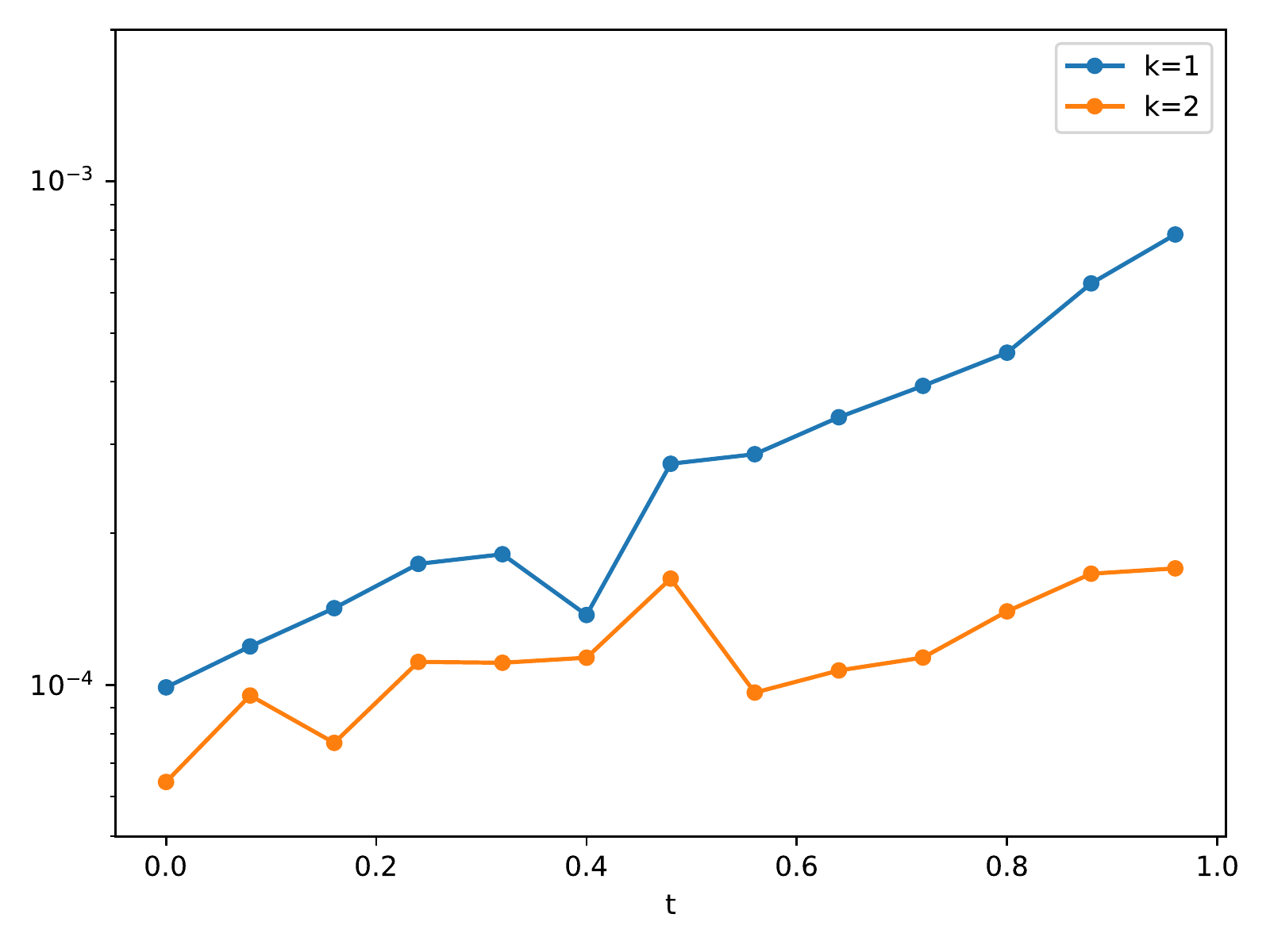}
		%		\subcaption{DGM-2-couple}
	\end{minipage}
	
	\caption{4D example. Relative $L^2$ error and relative KL divergence for different adaptive iterations $k$. Left panel: Relative $L^2$ error. Right panel: relative KL divergence.}
	\label{dim4_result}
\end{figure}

\begin{figure}[!h]
	\centering
	\begin{minipage}[t]{0.45\linewidth}
		\includegraphics[scale=0.4]{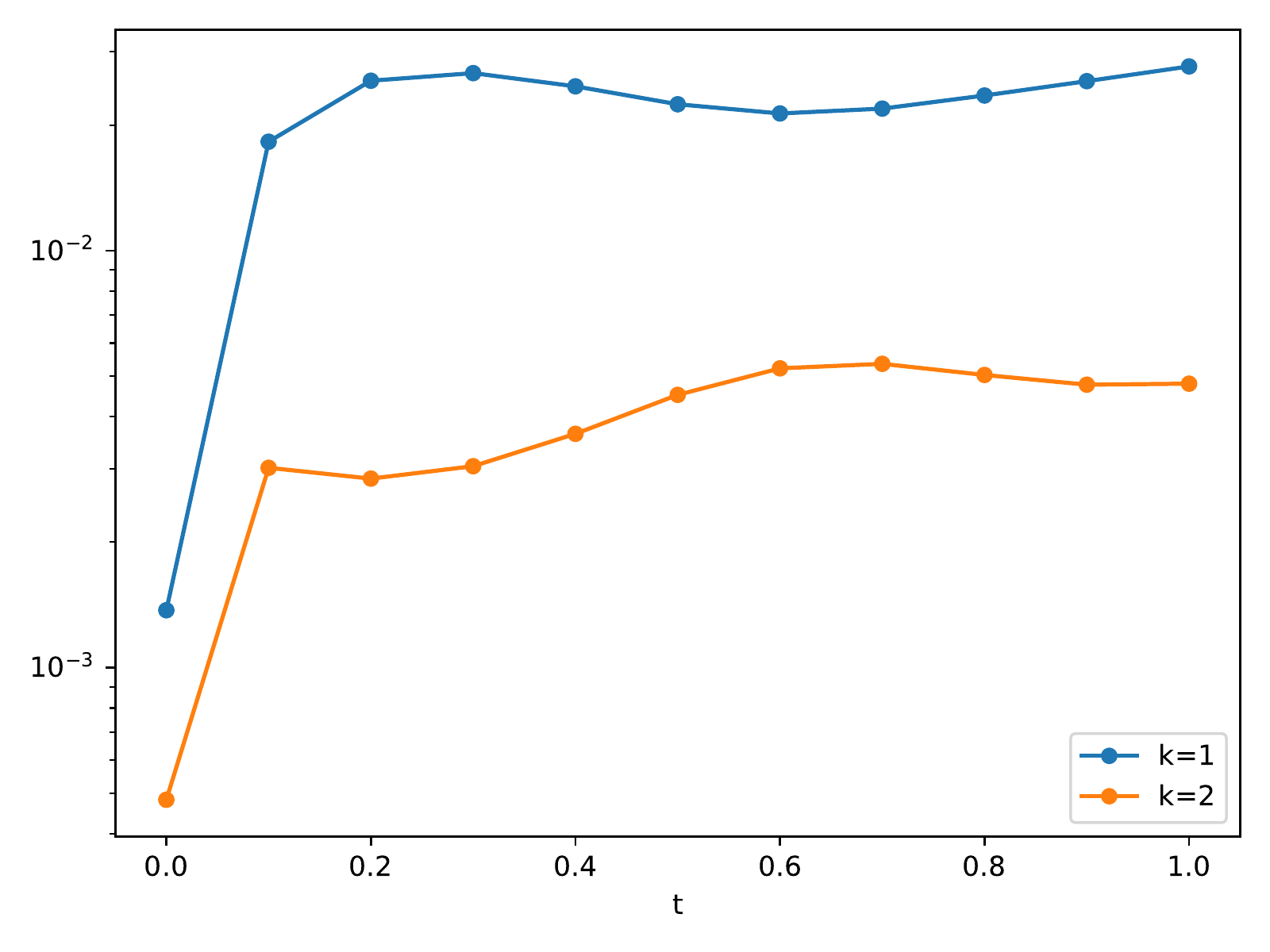}
		%		\subcaption{DGM-2}
	\end{minipage}
	\begin{minipage}[t]{0.45\linewidth}
		\includegraphics[scale=0.4]{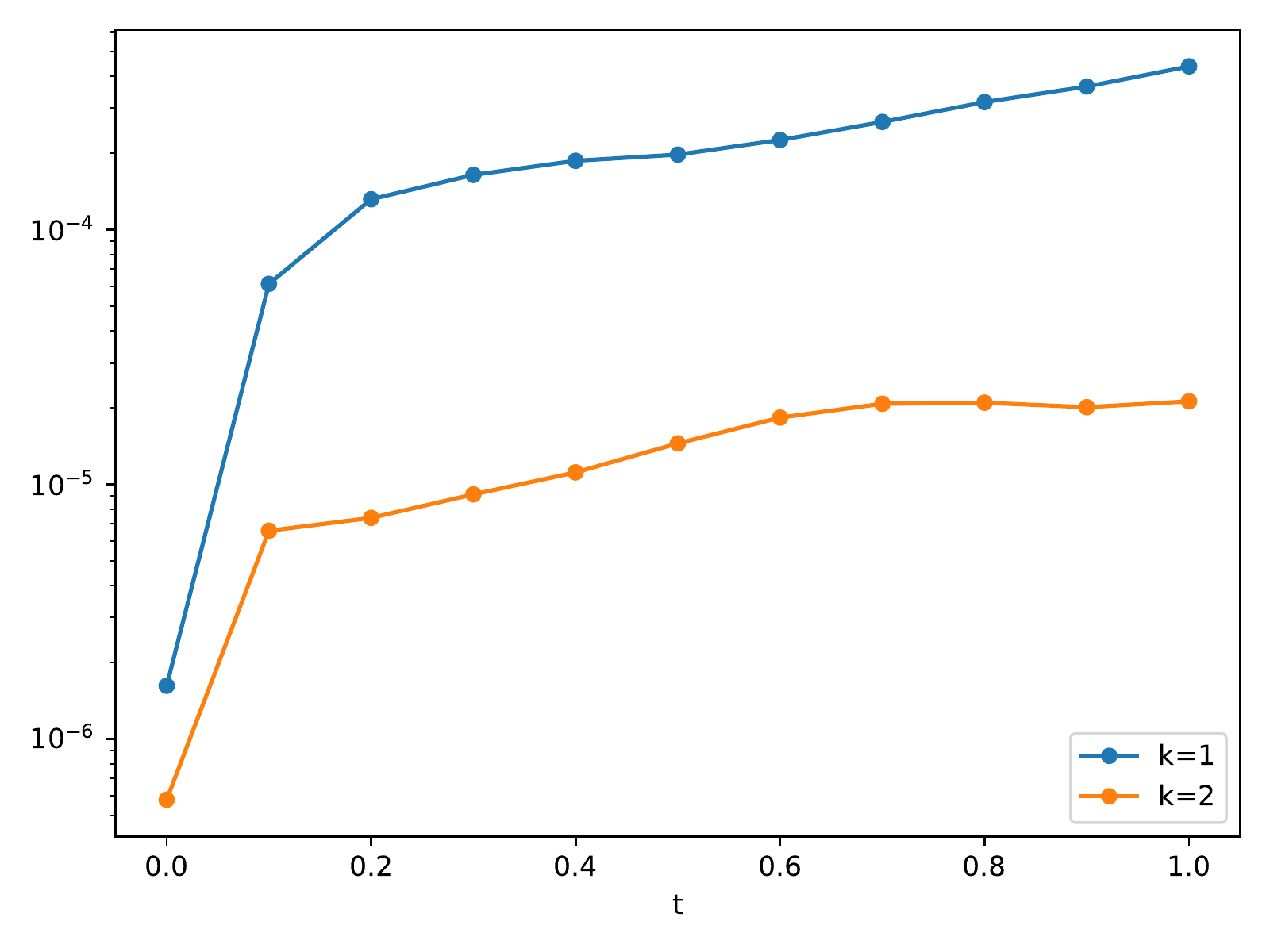}
		%		\subcaption{DGM-2-couple}
	\end{minipage}
	
	\caption{4D example. Relative $L^2$ error and relative KL divergence with nonuniform time partition. Left panel: Relative $L^2$ error. Right panel: relative KL divergence.}
	\label{dim4_result_nonuniform}
\end{figure}

To test the effectiveness of the adaptivity procedure, we present the relative $L^2$ errors and relative KL divergences with $k=1$ and $k=2$ in Figure \ref{dim4_result}. Notice that a clear (linear) increase of the numerical error over time is observed. To alleviate this, we next consider a nonuniform sampling procedure for both the time domain and physical domain. More precisely, the time interval $[0,1]$ is divided into $n$ parts $\{t_1, \dots, t_n\}$:
\begin{equation}
	t_i=1-\frac{r^{n-i}+1}{r^n+1}, \quad i=1,\cdots,n.
\end{equation}
where $r=1.05$ and $n=100$. For fixed $t$, the number of spatial sample points $N_{\bm{x}}$ is decided by
\begin{equation}
	N_{\bm{x}}(t_i)=N_0(1+\lfloor{(i-1)/20}\rfloor)
\end{equation}
where $N_0=5000$. Such a sampling procedure results in total $\sum\limits^n_iN_{\bm{x}}(t_i)=1.5\times 10^6$ training points. Then, we repeat the computation and the numerical result is presented in Figure \ref{dim4_result_nonuniform}, it is clearly shown that the numerical errors are well controlled over time (compared to Figure \ref{dim4_result}).

\begin{figure}[!h]
	\centering
	\begin{minipage}[t]{0.3\linewidth}
		\includegraphics[scale=0.3]{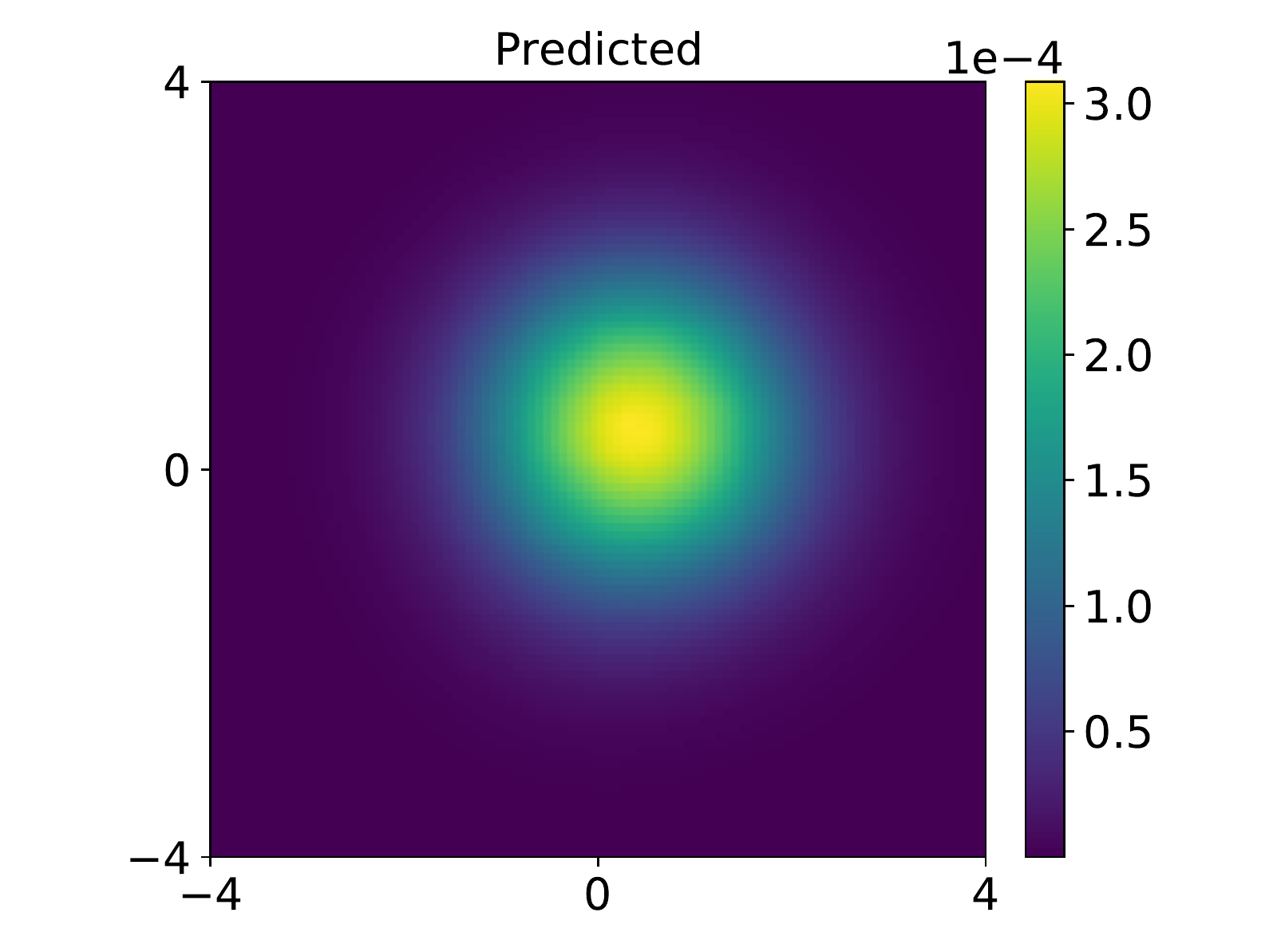}
		%		\subcaption{DGM-2}
	\end{minipage}
	\begin{minipage}[t]{0.3\linewidth}
		\includegraphics[scale=0.3]{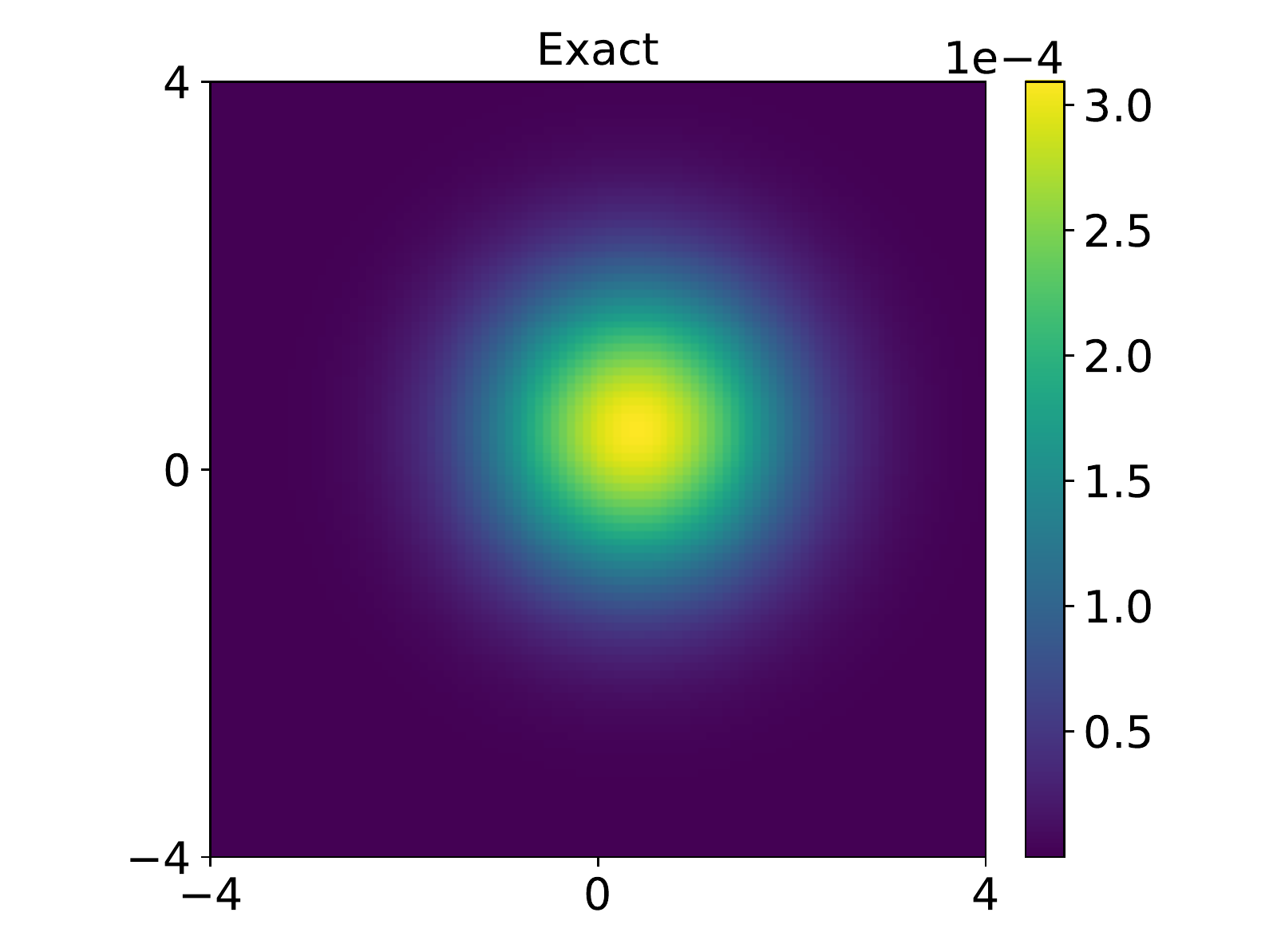}
		%		\subcaption{DGM-2-uncouple}
	\end{minipage}
	\begin{minipage}[t]{0.3\linewidth}
		\includegraphics[scale=0.3]{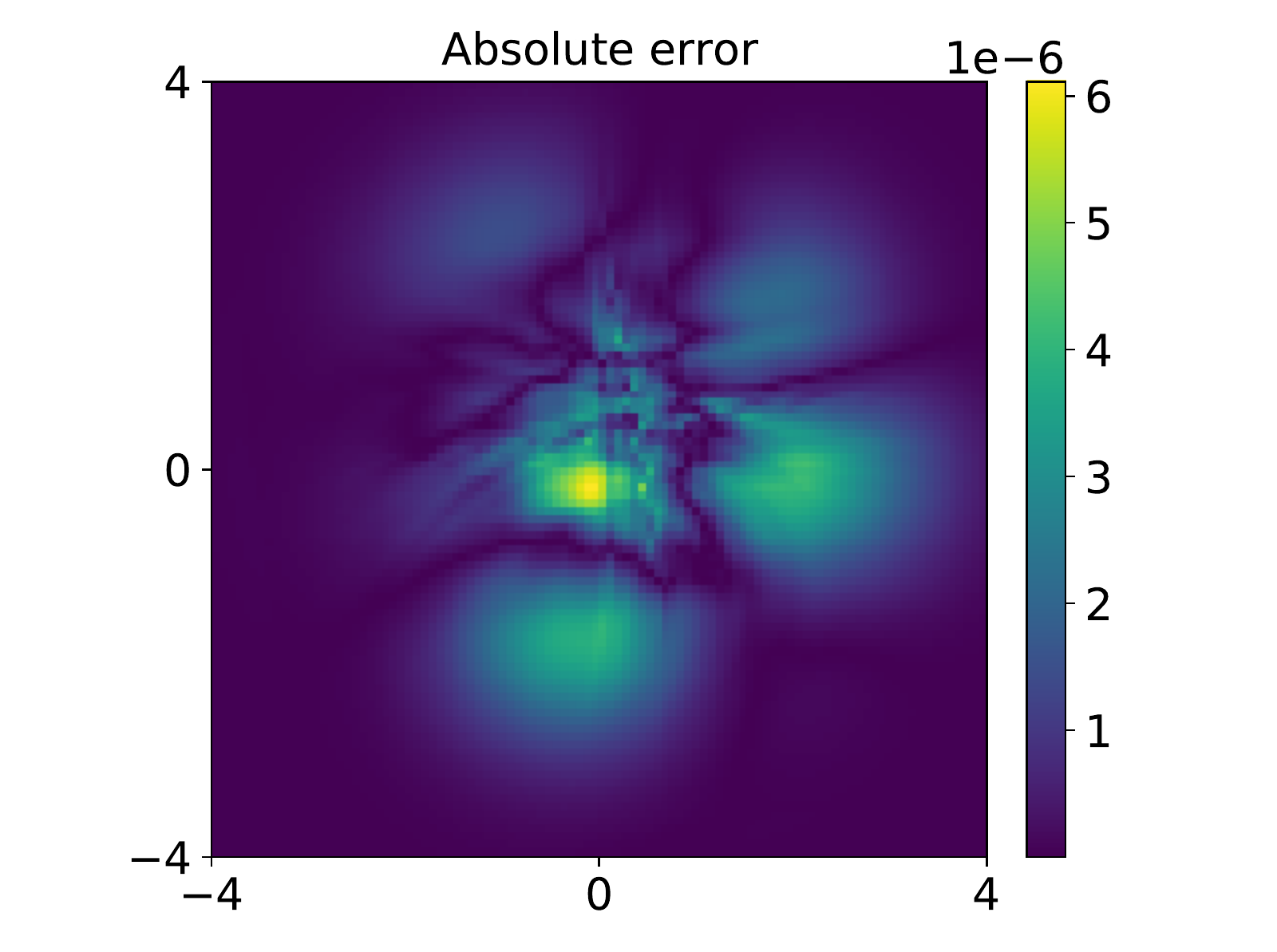}
		%		\subcaption{DGM-2-couple}
	\end{minipage}
	
	\begin{minipage}[t]{0.3\linewidth}
		\includegraphics[scale=0.3]{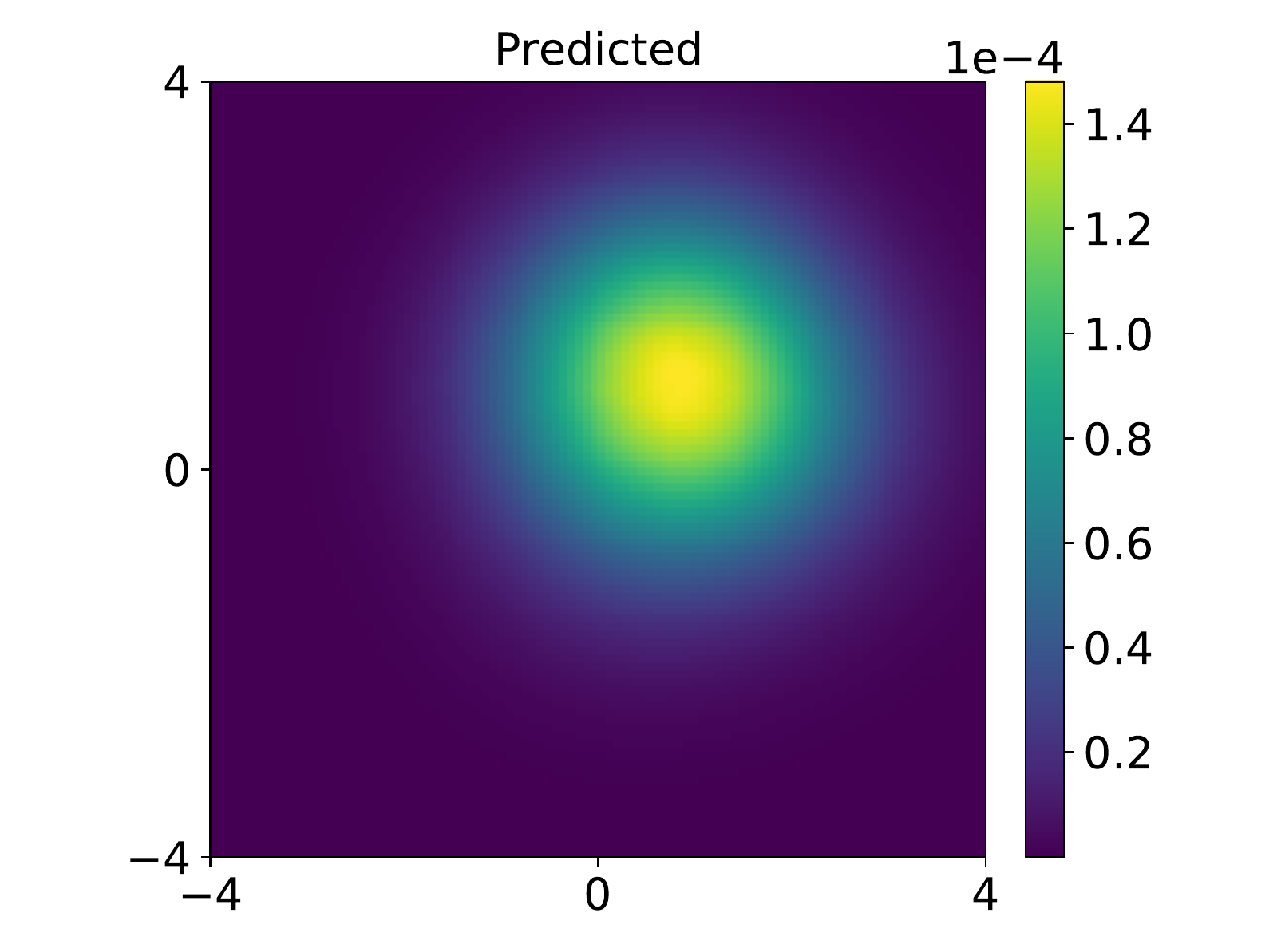}
		%		\subcaption{DGM-2}
	\end{minipage}
	\begin{minipage}[t]{0.3\linewidth}
		\includegraphics[scale=0.3]{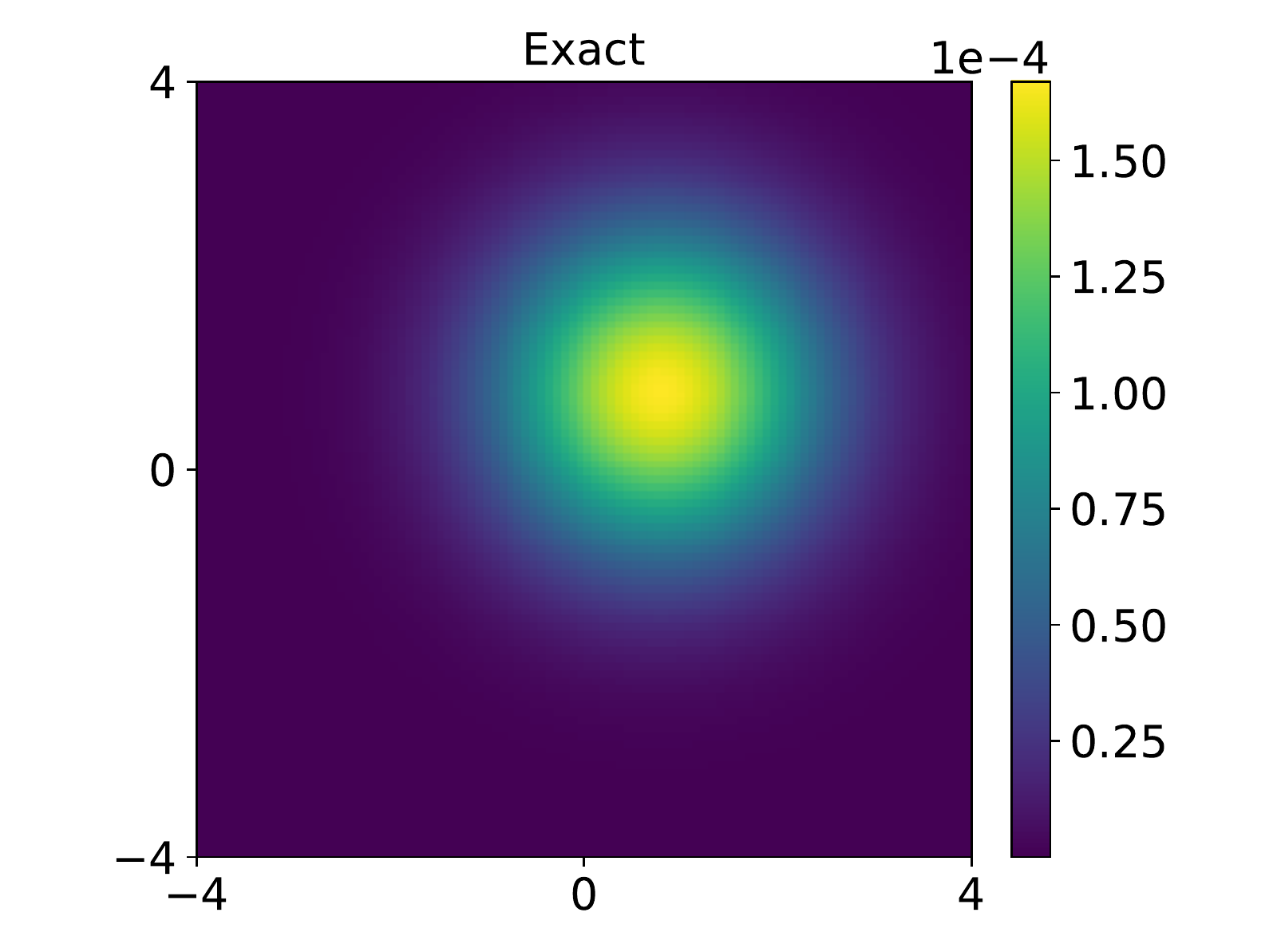}
		%		\subcaption{DGM-2-uncouple}
	\end{minipage}
	\begin{minipage}[t]{0.3\linewidth}
		\includegraphics[scale=0.3]{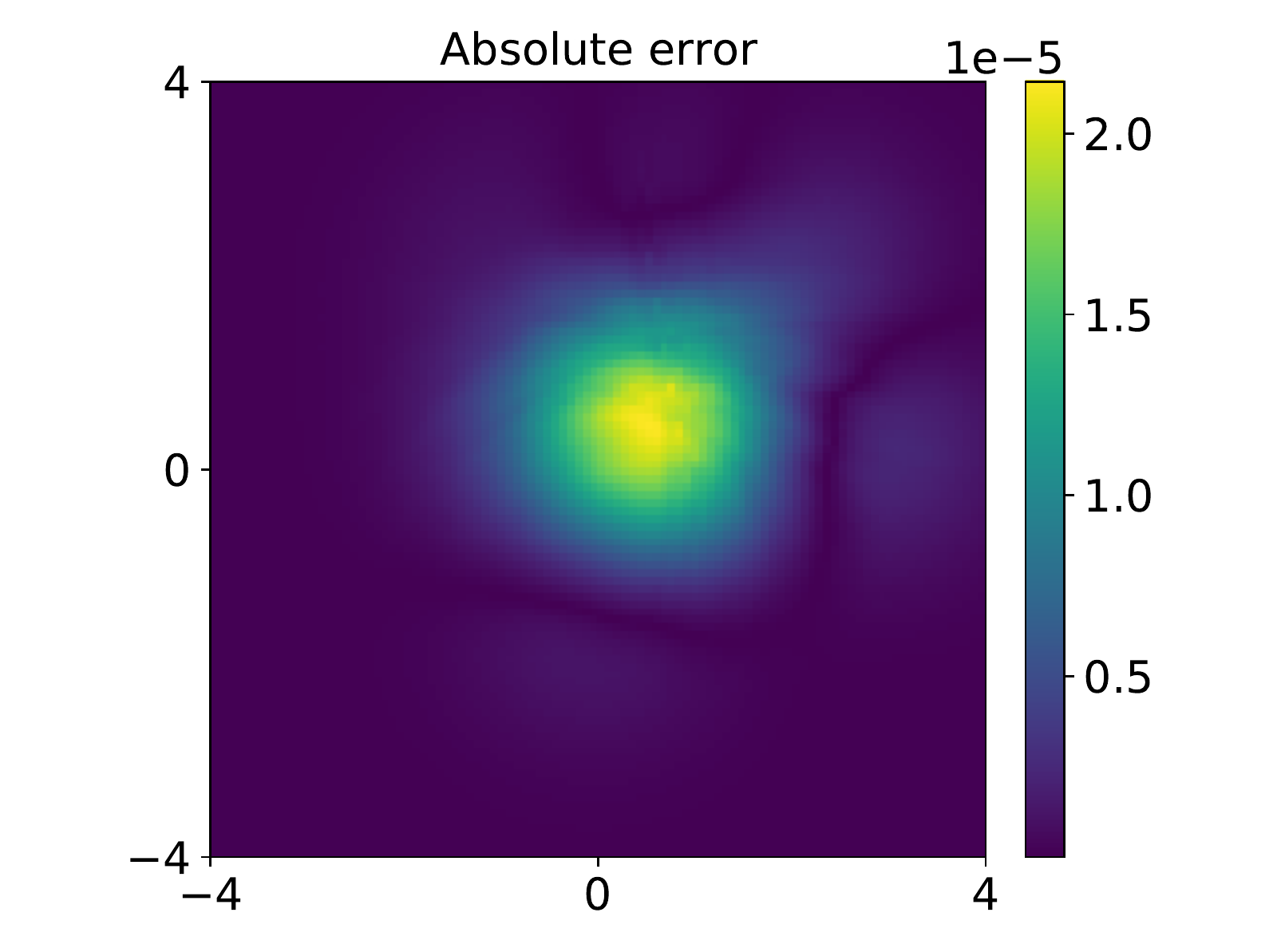}
		%		\subcaption{DGM-2-couple}
	\end{minipage}
	\caption{8D example. Predicted solution versus the reference solution projected for different time $t$. Top row: $x_2=\cdots=x_7=0.4, t=0.2$. Bottom row: $x_2=\cdots=x_7=0.8, t=0.4.$}
	\label{8d_example_result}
\end{figure}

\begin{figure}[!h]
	\centering
	\begin{minipage}[t]{0.45\linewidth}
		\includegraphics[scale=0.4]{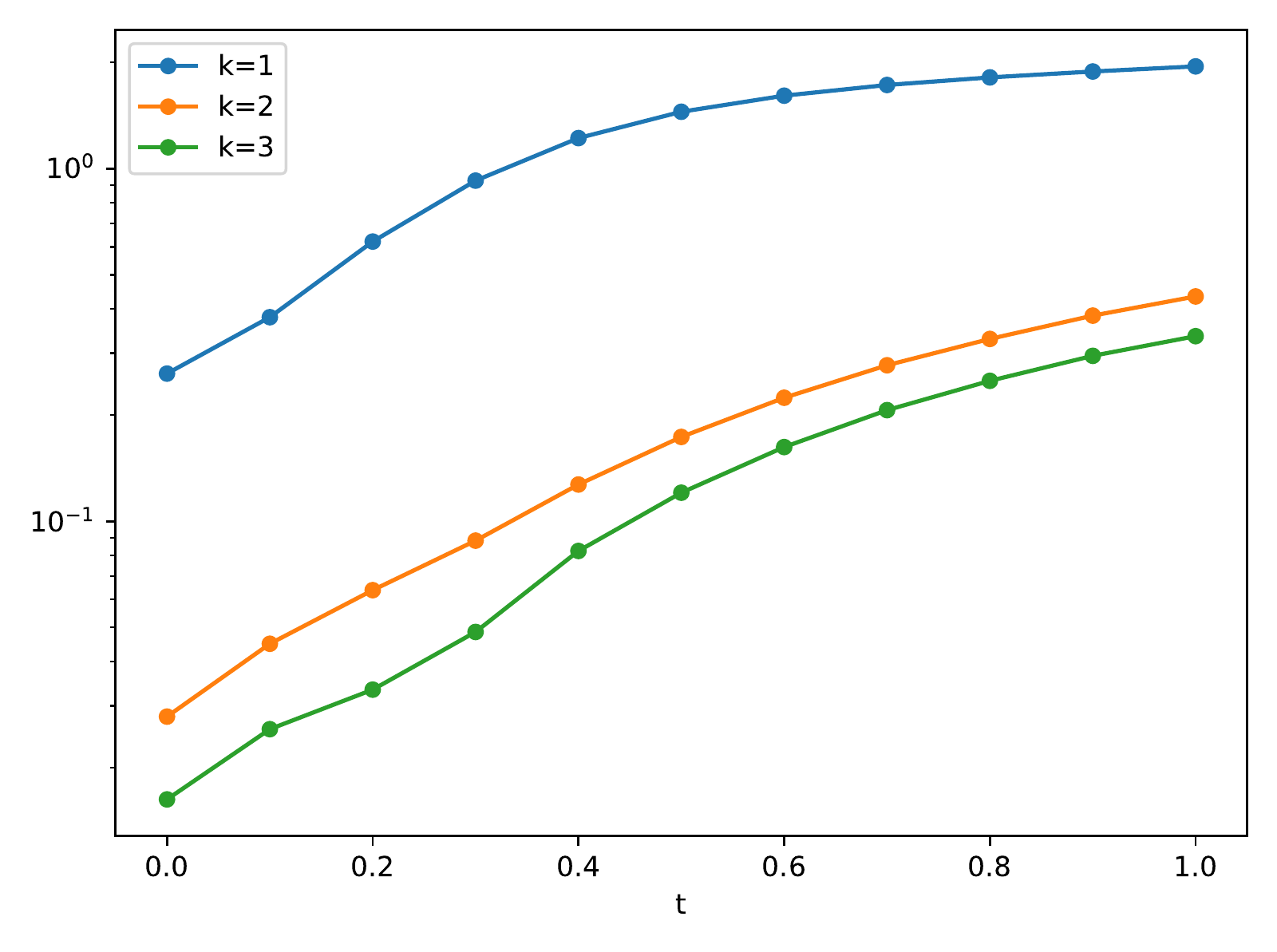}
		%		\subcaption{DGM-2}
	\end{minipage}
	\begin{minipage}[t]{0.45\linewidth}
		\includegraphics[scale=0.4]{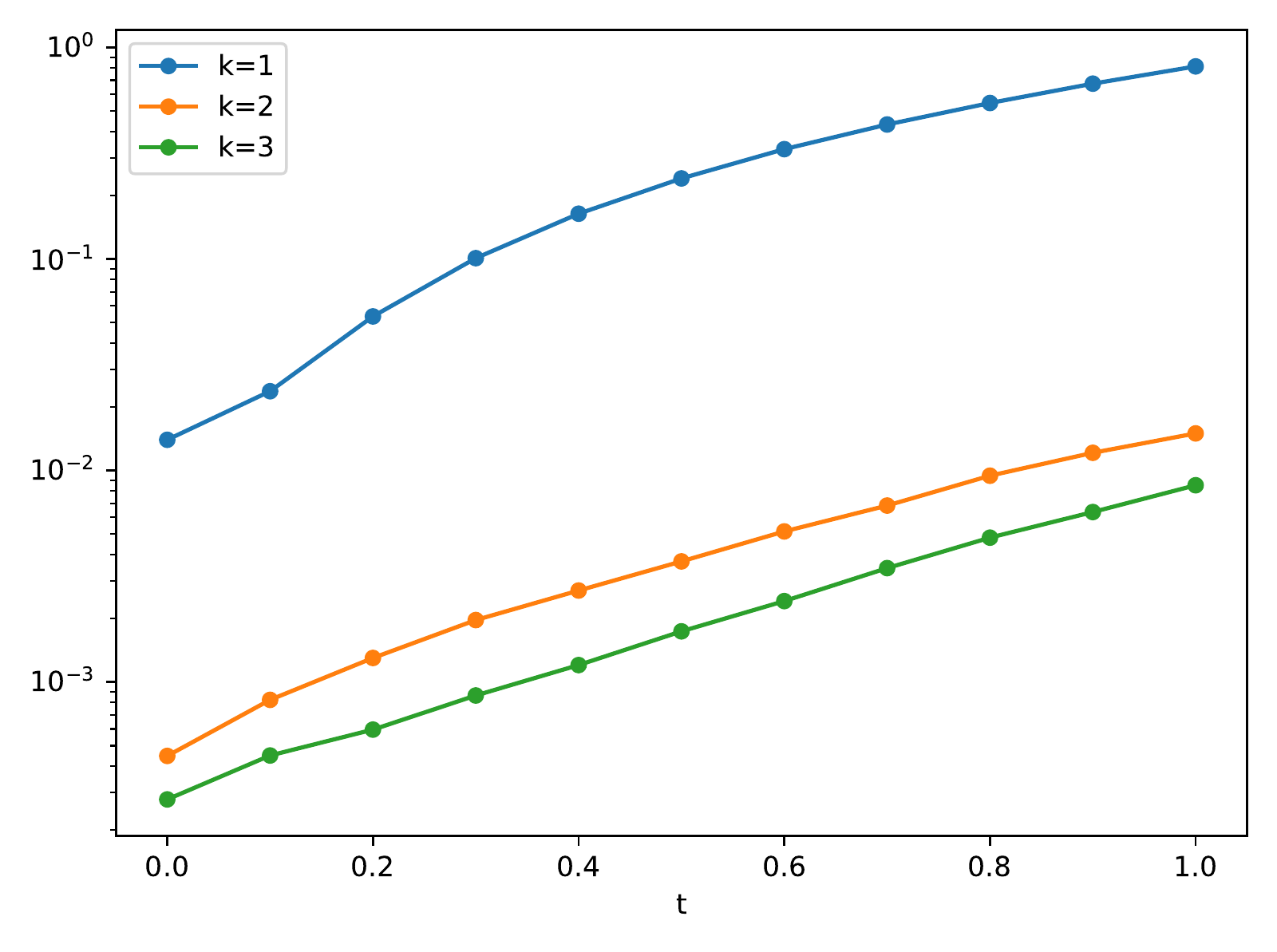}
		%		\subcaption{DGM-2-couple}
	\end{minipage}
	
	\caption{8D example. Relative $L^2$ error versus relative KL divergence for different adaptive iterations $k$. Left panel: Relative $L^2$ error. Right panel: relative KL divergence.}
	\label{dim8_err}
\end{figure}

For $d=8$, we set $N_e=100, \alpha=2$, $N_{\mathrm{adaptive}}=3$, and batch size is $10000$. We take $L=10$ affine coupling layers and actnorm layers, and we turn off the polynomial spline layer. The initial spatial training set is generated through a uniform distribution with a range $[-5,5]^8$, and the corresponding initial temporal training set is uniformly sampled in the unit interval $[0,1]$, and thus the total sample size is equal to $20000\times 25= 500,000$.
The predicted solution and the exact solution for $t=0.2$ and $t=0.4$ are presented in Figure \ref{8d_example_result}. Furthermore, the relative errors with different adaptive iterations are shown in Figure \ref{dim8_err}. Again, we may improve the convergence by adding more training points as time evolves; however, for high-dimensional problems this will introduce a huge computational cost.

\section{Conclusions} \label{section:5}
We have proposed to use the temporal normalizing flows for solving time-dependent Fokker-Planck (TFP) equations. Our approach relies on modelling the target solution by temporal normalizing flows. The temporal normalizing flow is then trained based on the TFP loss function, without requiring any labeled data. Being a machine learning scheme, the proposed approach is mesh-free and can be easily applied to high dimensional problems. We present a variety of test problems to show the effectiveness of the learning approach. There are, however, many important issues need to be addressed. From a theoretical point of view, a rigorous stability and accuracy analysis is still missing. From a practical viewpoint, for high dimensional problems, one needs to carefully design the training sets to balance the computational cost and the approximation error (which seems to increase as time evolves). Another possible way to tackle this issue is to include more temporal information in the reference process or in the network architecture.

\bibliographystyle{plainnat}
\bibliography{references}

\end{document}